\documentclass[11pt]{article}

\usepackage[final]{acl}

\usepackage{times}
\usepackage{latexsym}

\usepackage[T1]{fontenc}

\usepackage[utf8]{inputenc}

\usepackage{microtype}

\usepackage{inconsolata}

\usepackage{graphicx}

\usepackage{booktabs}
\usepackage{makecell}
\usepackage[T2A,T1]{fontenc}
\usepackage[russian,english]{babel}
\usepackage{subcaption}
\usepackage{multirow}
\usepackage{xcolor}
\usepackage{enumitem}
\usepackage{amsmath}
\usepackage{fvextra}
\usepackage{tabularx}

\title{ConlangBench: Exploring Language Knowledge and Learning in LLMs through Diverse Constructed Languages}

\author{
 \textbf{Jinhong Jeong\textsuperscript{1}} \quad
 \textbf{Seungyeop Yi\textsuperscript{2}} \quad
 \textbf{Sangah Lee\textsuperscript{2}} \quad
 \textbf{Youngjae Yu\textsuperscript{2}}
\\
 \textsuperscript{1}Yonsei University \quad
 \textsuperscript{2}Seoul National University
\\
\\
\texttt{jjhsnail0822@yonsei.ac.kr}
}

\begin{document}
\maketitle

\begin{abstract}
  Constructed languages (conlangs) are intentionally created human languages with a rich tradition of linguistic creativity. Despite their potential for studying language learning in large language models (LLMs), existing conlangs remain largely underexplored in LLM research. We present \textbf{ConlangBench}, the first large-scale benchmark for evaluating and training LLMs on 21 existing conlangs. We collect over 21M conlang--English parallel sentence pairs (including 430K pairs across the 20 non-Esperanto conlangs) and 321K vocabulary entries. In bidirectional translation experiments, we find that models perform better on \textit{a posteriori} conlangs, whose vocabularies are derived from natural languages, reflecting the design characteristics of conlangs. Training on ConlangBench also shows that models can learn all eight conlangs for which sufficient parallel corpora are available, while their learning curves vary depending on how the conlangs were created. Our findings suggest that conlangs provide a unique testbed for investigating how LLMs acquire low-resource languages.
\end{abstract}

\section{Introduction}

\begin{figure}[tb]
    \centering
    \includegraphics[width=\columnwidth]{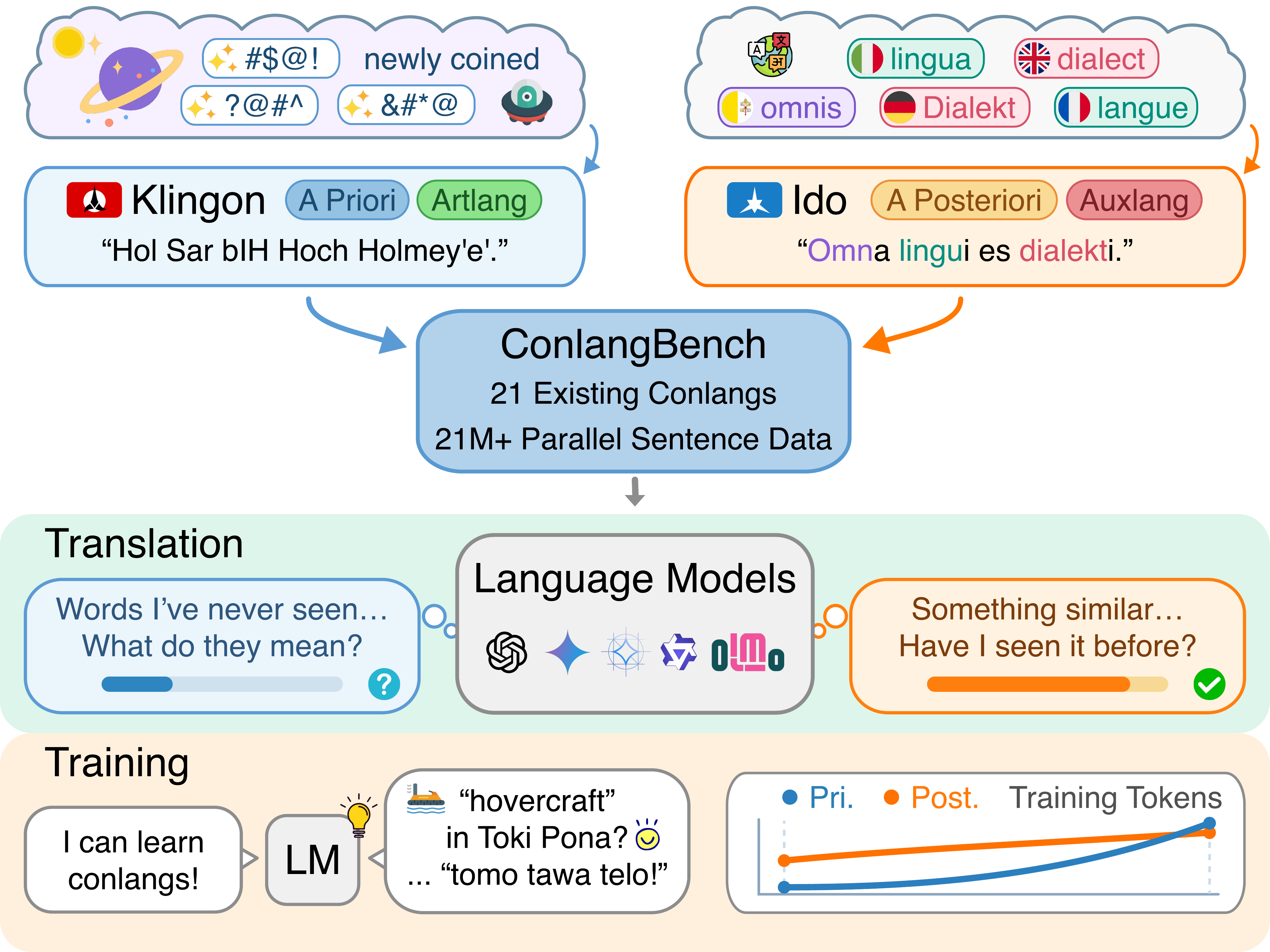}
    \caption{\textbf{ConlangBench} investigates how LLMs translate and learn existing constructed languages (conlangs). We collect parallel corpora and vocabularies for 21 conlangs and analyze model performance on both tasks. Models exhibit distinct performance patterns reflecting the conlangs' categories and design characteristics.}
    \label{fig:figure_1}
\end{figure}
\begin{figure*}[tb]
    \centering
    \includegraphics[width=\textwidth]{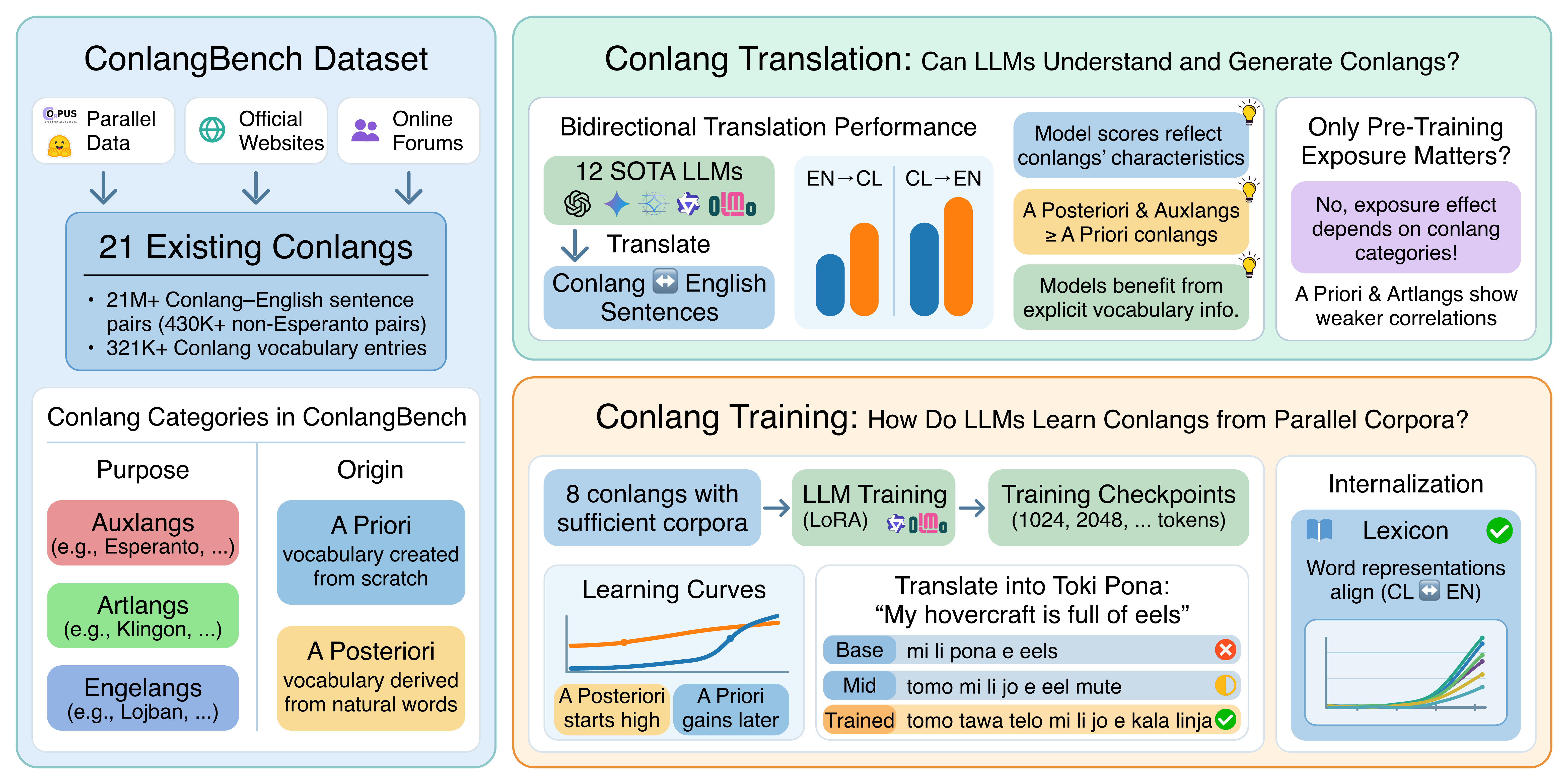}
    \caption{An overview of \textbf{ConlangBench}, the first large-scale benchmark for analyzing LLMs on diverse constructed languages (conlangs). (1) ConlangBench Dataset (\S\ref{sec:the_conlangbench_dataset}): We collect 21M+ conlang--English parallel sentence pairs across 21 existing conlangs (including 430K+ non-Esperanto pairs), along with 321K+ vocabulary entries. (2) Conlang Translation (\S\ref{sec:benchmarking_conlang_translation}): We evaluate 12 LLMs on conlang--English translation, showing that model performance reflects conlangs' characteristics. (3) Conlang Training (\S\ref{sec:training_models_on_conlang_parallel_corpora}): We train LLMs on conlangs with sufficient parallel data, demonstrating distinct learning trajectories across conlang categories.}
    \label{fig:main_figure}
\end{figure*}

Constructed languages, or conlangs, are languages that are deliberately created by individuals or groups for human communication, in contrast to natural languages, whose origins cannot typically be traced to a specific creator~\citep{sanders2020primer, goodall2023constructed, doi:10.1073/pnas.2313473122}. For centuries, conlangs have been designed for a wide range of purposes, including international auxiliary languages such as Esperanto, artistic languages such as Klingon, and engineered languages such as Lojban~\citep{schreyer2021constructed}. Unlike natural languages, whose vocabulary and grammar emerge bottom-up through speaker usage, conlangs are purposefully created in a top-down manner.

This top-down design makes conlangs a unique probe for investigating how LLMs acquire languages, raising two research questions. (1) To what extent can LLMs understand and generate existing conlangs? (2) How do LLMs learn existing conlangs through parallel corpus training? Most conlangs are extremely low-resource, yet they differ markedly across typological categories. For example, can LLMs exploit prior lexical knowledge to translate \textit{a posteriori} conlangs, whose vocabularies are adapted from natural languages, in contrast to \textit{a priori} conlangs, whose vocabularies are constructed from scratch?

Prior studies have mainly focused on creating or analyzing study-specific conlangs designed for experimental purposes~\citep{liu2025goldmedalsroomdiagnosing, alper-etal-2026-conlangcrafter, taguchi-sproat-2026-creating}. However, to the best of our knowledge, none provide a large-scale analysis of diverse existing conlangs using parallel corpora to evaluate and train LLMs, examining how conlang design characteristics influence LLM performance and learning trajectories.

In this paper, we present \textbf{ConlangBench}, an analytical benchmark covering 21 existing conlangs with their corpora and vocabularies. As part of ConlangBench, we construct the first large-scale parallel corpus with over 21 million conlang--English sentence pairs, including more than 430K sentence pairs from non-Esperanto conlangs.\footnote{Since Esperanto is the most widely used conlang as an auxiliary language, orders of magnitude more data are available for it than for most other conlangs~\citep{gonzalez2024networkanalysisapproachconlang}.}

With ConlangBench, we first evaluate the bidirectional translation capabilities of 12 LLMs on conlang--English sentence pairs. Open-weight LLMs generally perform better on \textit{a posteriori} conlangs than on \textit{a priori} conlangs, while pre-training exposure is associated with translation scores for \textit{a posteriori} but not for \textit{a priori} conlangs. This suggests that models translate conlangs by exploiting lexical knowledge of natural languages acquired during pre-training. A few notable exceptions, such as Volapük, are explained by their design characteristics and historical development.

We then investigate how LLMs learn conlangs when trained on parallel corpora. Experiments show that models can learn all eight conlangs for which sufficient training data are collected, but their learning curves differ by conlang category. Our results further reveal that models can acquire lexical knowledge of conlangs from parallel data. Figure~\ref{fig:figure_1} and \ref{fig:main_figure} summarize our main contributions:

\begin{enumerate}
  \item We present \textbf{ConlangBench}, a comprehensive benchmark built on the first large-scale parallel corpus for 21 existing conlangs, with 21M+ conlang--English sentence pairs (including 430K+ pairs from 20 non-Esperanto conlangs) and 321K+ vocabulary entries.
  \item We evaluate various LLMs on bidirectional conlang--English translation, demonstrating that LLM translation performance reflects the design characteristics of conlangs.
  \item We analyze how LLMs learn conlangs through parallel corpus training, revealing distinct learning trajectories across conlang categories and the acquisition of lexical knowledge.
\end{enumerate}

Our study emphasizes the linguistic and cultural significance of various conlangs, opening new opportunities to investigate how LLMs acquire typologically diverse low-resource languages.

\section{Related Work}

\subsection{Constructed Languages}

The origins of constructed languages lie in the European Enlightenment, when scholars seek to develop rationally designed philosophical languages capable of systematically classifying human knowledge, such as Leibniz's ``Lingua Generalis'' and John Wilkins' ``Philosophical Language''~\citep{wilkins1668essay, couturat1961opuscules}. In the late nineteenth century, international auxiliary languages (e.g., Esperanto, Ido, and Volapük) emerge, driven by aspirations for international peace~\citep{schleyer1884volapuk, zamenhof1887, de1919complete}. The tradition of conlang creation further diversifies in the twentieth century with the development of artistic languages for fantasies and science fiction, such as \citeposs{tolkien1955return} Quenya in \textit{The Lord of the Rings} and \citeposs{okrand1992klingon} Klingon in the \textit{Star Trek} franchise. With the spread of the Internet, conlang creators have been able to disseminate their languages more widely through online communities~\citep{schreyer2021constructed}, and these textual resources form the basis of our dataset.

\subsection{Use of Constructed Languages in Language Model Studies}

Researchers have recently used artificially created languages to conduct linguistic experiments with language models. For example, \citet{kallini-etal-2024-mission} and \citet{kuribayashi-etal-2024-emergent} suggest that LLMs struggle to learn cognitively impossible language structures, while \citet{Jeong_Lee_Lee_Han_Yu_2026} show that LLMs can associate sound-symbolic constructed words with certain meanings. Other studies have examined the language learning capabilities of LLMs using newly designed constructed languages~\citep{liu2025goldmedalsroomdiagnosing, ma2025implicit, kouwenhoven-etal-2025-searching, swain2025talkingoompaloompasnovel} or demonstrated LLMs' agentic ability to create new languages~\citep{alper-etal-2026-conlangcrafter, taguchi-sproat-2026-creating}. However, the field of language modeling still lacks analytical studies on evaluation and learning across multiple existing conlangs with rich traditions, although a few studies have individually investigated popular conlangs like Esperanto and Toki Pona~\citep{10.1007/978-3-031-36616-1_52, bick-2025-annotated}.

\section{The ConlangBench Dataset}
\label{sec:the_conlangbench_dataset}

We present the ConlangBench dataset, the first large-scale parallel corpus for existing conlangs, comprising over 21 million conlang--English sentence pairs across 21 conlangs. Excluding Esperanto, which constitutes the majority of the corpus, ConlangBench still contains 430,098 parallel sentence pairs across the remaining 20 conlangs. The dataset also includes vocabulary resources for each conlang.

\subsection{Target Languages}

Conlangs generally have no native speakers and can be created by anyone, making it challenging to determine which existing conlangs should be included in the corpus. We therefore establish a selection criterion to include only sufficiently prominent conlangs that have an ISO 639-3 language code and are registered in the Glottolog 5.3 database~\citep{harald_hammarstrom_2026_18840935} as an ``Artificial Language''. This approach results in 21 target conlangs, excluding conlangs for which detailed corpus data is unavailable.

\subsection{Classification of Conlangs}

This paper classifies conlangs along two widely adopted axes, namely purpose and manner of creation~\citep{peterson2015art, schreyer2021constructed}.

\paragraph{Category by Purpose.}

We group conlangs into three categories based on their purpose: international auxiliary languages (auxlangs), artistic languages (artlangs), and engineered languages (engelangs).

\begin{itemize}[leftmargin=*, labelsep=0.5em]
  \item \textbf{Auxlangs}: languages designed for use as international or regional lingua francas.
  \item \textbf{Artlangs}: languages created for artistic works, such as novels, science fiction, and fantasy.
  \item \textbf{Engelangs}: languages designed for linguistic, philosophical, or logical experimentation.
\end{itemize}

\paragraph{Manner of Creation.}

Conlangs fall into two broad categories in terms of their lexical origin.

\begin{table}[htb]
\centering
\small
\begin{tabular}{p{0.92\columnwidth}}
    \toprule
    \textbf{Kotava (a priori language)} \\
    \midrule
    \textit{Kot ayik sokoblir nuyaf is miltaf gu bagaliuca is rokeem.} \\
    \midrule
    \textbf{Interlingua (a posteriori language)} \\
    \midrule
    \textit{Tote le esseres human nasce libere e equal in dignitate e in derectos.} \\
    \midrule
    \textbf{English Translation} \\
    \midrule
    `All human beings are born free and equal in dignity and rights.' \\
    \bottomrule
\end{tabular}
\caption{Example sentences in two conlangs, Kotava (a priori) and Interlingua (a posteriori). Words in Interlingua closely resemble those of natural languages, particularly English and the Romance languages (e.g., \textit{human} and \textit{dignitate}). In contrast, Kotava words exhibit no surface similarity to natural-language vocabularies.}
\label{tab:conlang_examples}
\end{table}

\begin{itemize}[leftmargin=*, labelsep=0.5em]
  \item \textbf{A priori}: languages whose vocabulary is artificially generated without relying on the lexicons of natural languages.
  \item \textbf{A posteriori}: languages whose vocabulary is derived or adapted from natural languages.
\end{itemize}

Table~\ref{tab:conlang_examples} provides example conlang sentences illustrating the characteristics of each category.

\subsection{Parallel Corpus Collection}

\paragraph{Data Collection.}

To construct the conlang--English parallel corpora, we use data from OPUS~\citep{4992de1b5fb34f3e9691772606b36edf}, Tatoeba~\citep{tiedemann-2020-tatoeba}, Hugging Face datasets, and manually collected texts from more than 30 official conlang websites and fan communities. For detailed source information, refer to Appendix~\ref{sec:datasets}.

\paragraph{Preprocessing.}

After collecting the raw data, we translate all non-English natural language texts into English using Gemma-4-31B-it~\citep{gemmateam2026gemma4}, ensuring that all samples consist of conlang--English pairs. We then split both the conlang and English texts into sentences, retaining samples for which both sides yield the same number of sentences. We also remove noisy, empty, duplicate, excessively long, and misaligned samples. Finally, we split the dataset into training and test sets for each conlang, where the test set size is determined as 10\% of the samples, up to a maximum of 1,000 samples. Table~\ref{tab:conlang_list} provides an overview of the conlangs and their dataset sizes.

\begin{table}[htb]
\centering
\small
\begin{tabular}{llllr}
    \toprule
    \textbf{Code} & \textbf{Conlang} & \textbf{Purp.} & \textbf{Mann.} & \textbf{\# Sample} \\
    \midrule
    ldn & Láadan & Enge. & Pri. & 223 \\
    jbo & Lojban & Enge. & Post. & 44,965 \\
    tok & Toki Pona & Enge. & Post. & 74,679 \\
    \midrule
    qya & Quenya & Art. & Pri. & 29,495 \\
    sjn & Sindarin & Art. & Pri. & 1,081 \\
    tlh & Klingon & Art. & Pri. & 40,167 \\
    bzt & Brithenig & Art. & Post. & 854 \\
    tzl & Talossan & Art. & Post. & 1,370 \\
    \midrule
    avk & Kotava & Aux. & Pri. & 4,827 \\
    afh & Afrihili & Aux. & Post. & 218 \\
    dws & \makecell[l]{Dutton World\\Speedwords} & Aux. & Post. & 79 \\
    epo & Esperanto & Aux. & Post. & 20,954,974 \\
    ido & Ido & Aux. & Post. & 38,614 \\
    igs & Interglossa & Aux. & Post. & 1,121 \\
    ile & Interlingue & Aux. & Post. & 16,112 \\
    ina & Interlingua & Aux. & Post. & 102,248 \\
    isv & Interslavic & Aux. & Post. & 5,564 \\
    lfn & \makecell[l]{Lingua Franca\\Nova} & Aux. & Post. & 53,204 \\
    neu & Neo & Aux. & Post. & 86 \\
    nov & Novial & Aux. & Post. & 948 \\
    vol & Volapük & Aux. & Post. & 14,243 \\
    \midrule
    \textbf{Total} & & & & \textbf{21,385,072} \\
    \bottomrule
\end{tabular}
\caption{Summary of languages in the ConlangBench dataset, our large-scale parallel corpus for 21 conlangs. We collect over 21 million samples of conlang--English sentence pairs. ``Code'' refers to the ISO 639-3 language code assigned to each language. ``Purp.'' and ``Mann.'' denote the conlang category by purpose (i.e., engineered, artistic, and auxiliary languages), and manner of creation (i.e., a priori and a posteriori), respectively.}
\label{tab:conlang_list}
\end{table}

\subsection{Conlang Vocabulary Data}

We also collect vocabulary resources for the 21 conlangs from official websites and fan communities to analyze models' lexical knowledge. The source materials are available in diverse formats, including HTML, PDF, and CSV. We therefore build a collection agent, powered by GPT-5.6 Sol, to retrieve and organize the vocabulary data. The resulting dataset contains 321,858 (\textit{word}, \textit{meaning}) pairs. The \textit{word} field contains the conlang headwords recorded in the sources, while preserving source-specific Romanized orthography and notation. The \textit{meaning} field contains English semantic definitions. Details are provided in Appendix~\ref{sec:vocabulary_collection}.

\section{Benchmarking Conlang Translation}
\label{sec:benchmarking_conlang_translation}

\begin{figure*}[!tb]
    \centering
    \begin{subfigure}[!tb]{\textwidth}
        \centering
        \includegraphics[width=\textwidth]{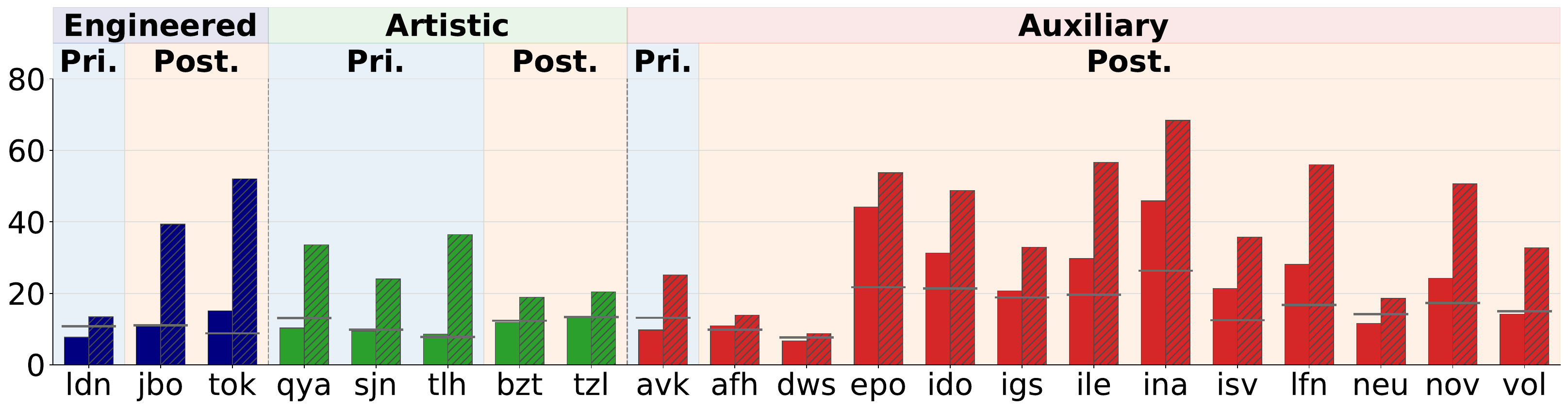}
        \caption{English-to-conlang translation scores (chrF++).}
        \label{fig:translation_scores_base_1}
    \end{subfigure}
    \par\vspace{0.5em}
    \begin{subfigure}[!tb]{\textwidth}
        \centering
        \includegraphics[width=\textwidth]{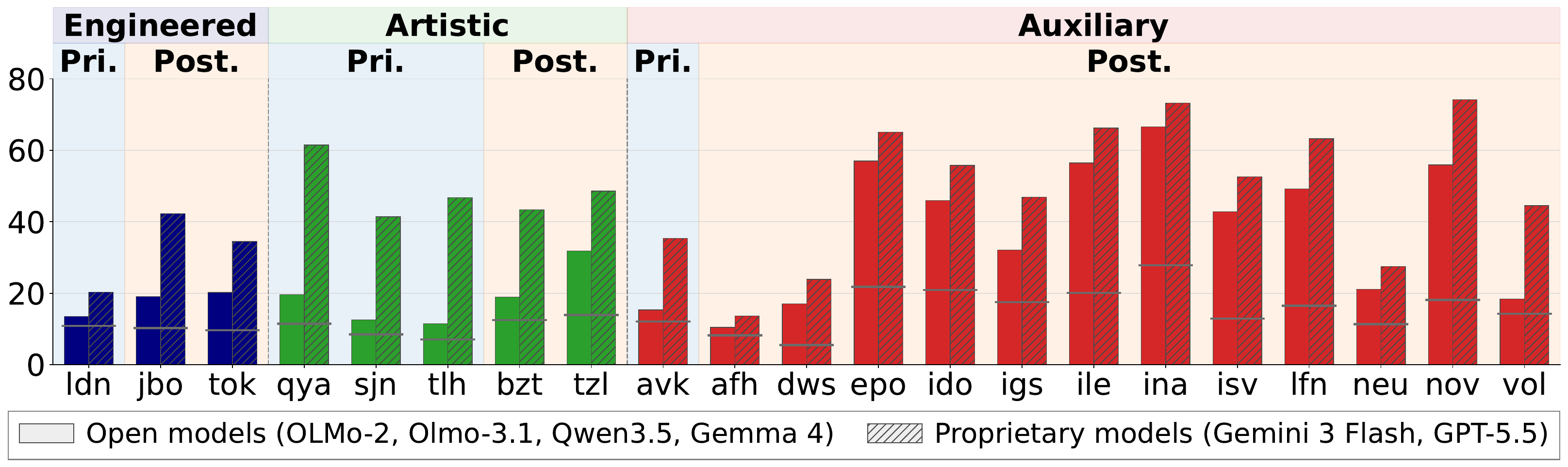}
        \caption{Conlang-to-English translation scores (chrF++).}
        \label{fig:translation_scores_base_2}
    \end{subfigure}
    \caption{chrF++ translation scores (y-axis) of 12 LLMs (10 open-weight and 2 proprietary models) across 21 existing conlangs (x-axis), grouped by conlang category. The gray lines indicate the source-copy baselines. Open-weight models generally achieve higher scores on a posteriori (Post.) conlangs, whose vocabularies are derived from natural languages, than on a priori (Pri.) conlangs. The scores further reflect conlang design characteristics. For example, unlike the pattern observed for most other conlangs, proprietary models have greater difficulty understanding Toki Pona (tok) than generating it.}
    \label{fig:translation_scores_base}
\end{figure*}

To what extent can LLMs understand and generate existing conlangs? We evaluate 12 LLMs (10 open-weight and 2 proprietary) on bidirectional conlang--English translation across 21 conlangs.

\subsection{Methodology}

For each conlang, the translation task consists of two directions: conlang-to-English and English-to-conlang. For each sample in the test split of each conlang in ConlangBench, we prompt 12 LLMs to translate the source language sentence into the target language. A range of proprietary and open-weight LLMs are employed, including gpt-5.5, gemini-3-flash-preview, Gemma 4, Qwen3.5, and OLMo 2/3.1.

For evaluation metrics, we adopt the corpus chrF++ score~\citep{popovic-2017-chrf}, which utilizes both character-level and word-level overlap, since no embedding model is available for reliably evaluating semantic similarity in conlangs. chrF++ calculates an F-score ($\beta=2$) based on hypothesis--reference overlap over character 1- to 6-grams and word 1- to 2-grams.

\subsection{Results}

Language models are capable of translating several existing conlangs, although their overall performance remains limited. The relatively larger and proprietary models generally outperform the smaller and open-weight models. Translation performance also varies across conlang types. Figure~\ref{fig:translation_scores_base} shows chrF++ translation scores in both directions for each conlang and language model. For detailed results, refer to Appendix~\ref{sec:conlang_translation_results}.

\paragraph{Performance Advantage in A Posteriori Auxlangs.}

Open-weight models generally achieve higher translation scores on a posteriori conlangs (e.g., Interlingua and Ido), which derive their vocabulary from natural languages, than on a priori conlangs in both translation directions (macro-average scores of 21.3 vs. 9.2 for English-to-conlang and 35.2 vs. 14.5 for conlang-to-English). This result implies that the strategy of adopting lexicons from natural languages, commonly used in constructing international auxiliary languages, benefits not only human learners but also LLMs. Notably, among a posteriori auxlangs, several conlangs such as Volapük (vol) score much lower than other conlangs in the same category. This pattern is consistent with its history, as Volapük extensively modifies vocabulary derived from natural languages, making the language more difficult to learn and contributing to its replacement by other auxlangs such as Esperanto (epo) in the early twentieth century~\citep{blanke2009causes}.

\paragraph{Reflection of Conlang Design Characteristics.}

In the majority of conlangs and models, conlang-to-English translation (conlang understanding) scores outperform English-to-conlang translation (conlang generation) scores, showing that understanding is easier than generation for language models. Models can exploit similarities to previously learned languages during understanding, whereas generation requires knowledge of the exact surface forms of the conlangs. An interesting exception is Toki Pona (tok), for which conlang generation (52.1) is easier than understanding (34.5) in the large proprietary models (Gemini 3 Flash and GPT-5.5). Since Toki Pona is a ``minimalist'' conlang designed to express maximum meaning with minimal complexity using only around 130 words~\citep{lang2014toki}, understanding Toki Pona requires LLMs to resolve substantial semantic ambiguity.

\subsection{Correlation with Pre-training Exposure Across Conlang Categories}

Model performance can be affected by pre-training exposure, although such exposure is challenging to measure directly. To estimate it, we employ Infini-gram~\citep{liu2024infinigram} to query up to 10,000 8-gram samples drawn from the training split of each conlang against the OLMo-2 pre-training corpus and compute the mean frequency.\footnote{To the best of our knowledge, OLMo-2 13B is the latest open-weight LLM with a queryable pre-training corpus, so we use the model for this analysis. Following prior work, we also set the n-gram length to 8 for querying~\citep{radford2019language}.} We then investigate whether variation in model performance across conlangs is associated with both pre-training exposure and conlang category.

\paragraph{Effect of Conlang Categories.}

\begin{figure}[!tb]
    \centering
    \begin{subfigure}[!tb]{\linewidth}
        \centering
        \includegraphics[width=\linewidth]{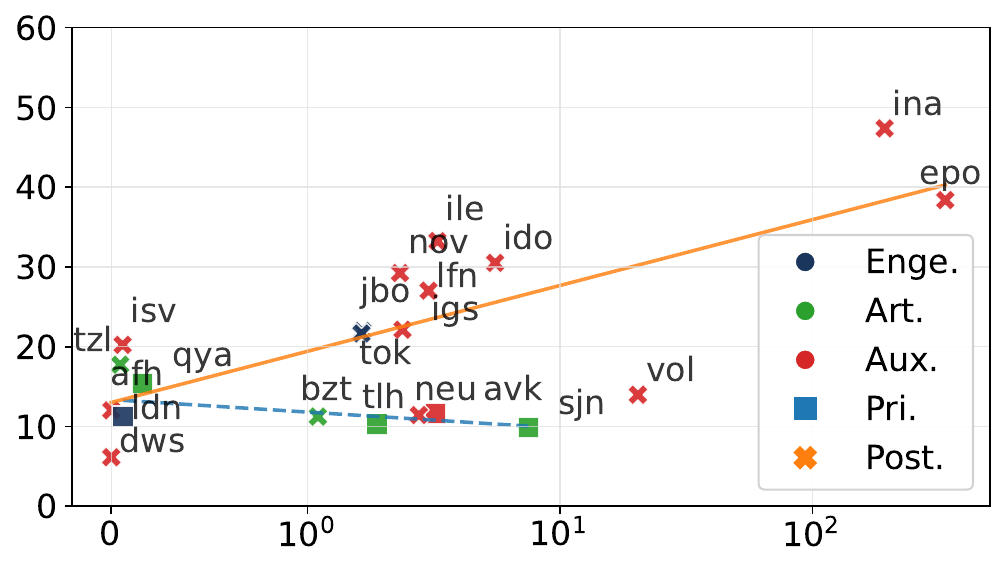}
        \caption{English-to-conlang translations.}
        \label{fig:ngram_translation_correlations_1}
    \end{subfigure}
    \par\vspace{0.5em}
    \begin{subfigure}[!tb]{\linewidth}
        \centering
        \includegraphics[width=\linewidth]{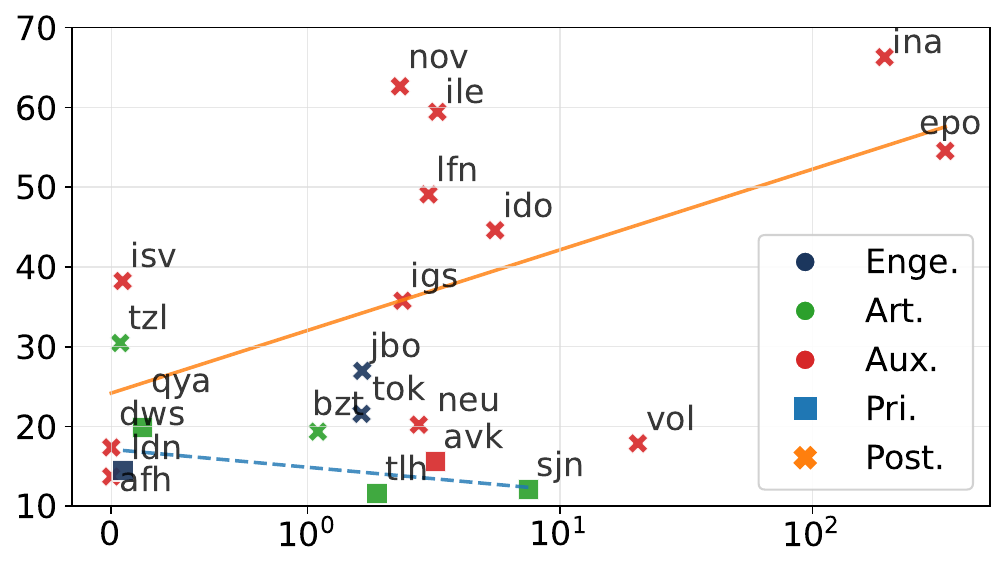}
        \caption{Conlang-to-English translations.}
        \label{fig:ngram_translation_correlations_2}
    \end{subfigure}
    \caption{Correlation between pre-training token exposure (mean 8-gram frequency on the x-axis) and conlang--English translation scores (chrF++ on the y-axis). Unlike a priori (Pri.) conlangs and artlangs (Art.), a posteriori (Post.) conlangs show stronger correlations between pre-training exposure and model performance.}
    \label{fig:ngram_translation_correlations}
\end{figure}

Exposure alone does not account for the performance variation across conlangs. As shown in Figure~\ref{fig:ngram_translation_correlations}, the relationship between pre-training exposure and model performance differs depending on the categories of each conlang. A posteriori conlangs show higher translation scores as exposure increases ($r=0.734$ and $r=0.558$), whereas a priori conlangs like Kotava (avk) and artlangs like Sindarin (sjn) maintain similar scores regardless of their pre-training exposure. This observation suggests that models benefit not only from greater pre-training exposure but also from lexical similarity to natural languages, performing better on conlangs whose vocabularies more closely resemble those of natural languages, such as Interlingua (ina) and Novial (nov).

\subsection{Benefits of Explicit Vocabulary Information}

To investigate whether models benefit from lexical knowledge when translating conlangs into English, we provide LLMs with explicit vocabulary information for each conlang at inference time, similar to \citet{tanzer2024a}. For each test sample, we measure the performance gains obtained by providing vocabulary entries, selecting the conlang--English lexical entry with the shortest character-based edit distance for each word in the sentence.

\paragraph{Different Gains from In-Context Vocabulary.}

The experimental results show that vocabulary information yields larger average chrF++ gains for a priori conlangs than for a posteriori conlangs. For Qwen3.5-9B, the average gains are 9.8 and 5.1, respectively. OLMo-2-13B exhibits a similar pattern, with average gains of 8.2 for a priori conlangs and 6.2 for a posteriori conlangs. These results indicate that a priori conlangs, for which models lack pre-existing lexical knowledge, benefit more from explicit vocabulary information, corroborating that models translate a posteriori conlangs by exploiting lexical knowledge acquired during pre-training.

\section{Training Models on Conlang Parallel Corpora}
\label{sec:training_models_on_conlang_parallel_corpora}

\begin{figure*}[!tb]
    \centering
    \includegraphics[width=\linewidth]{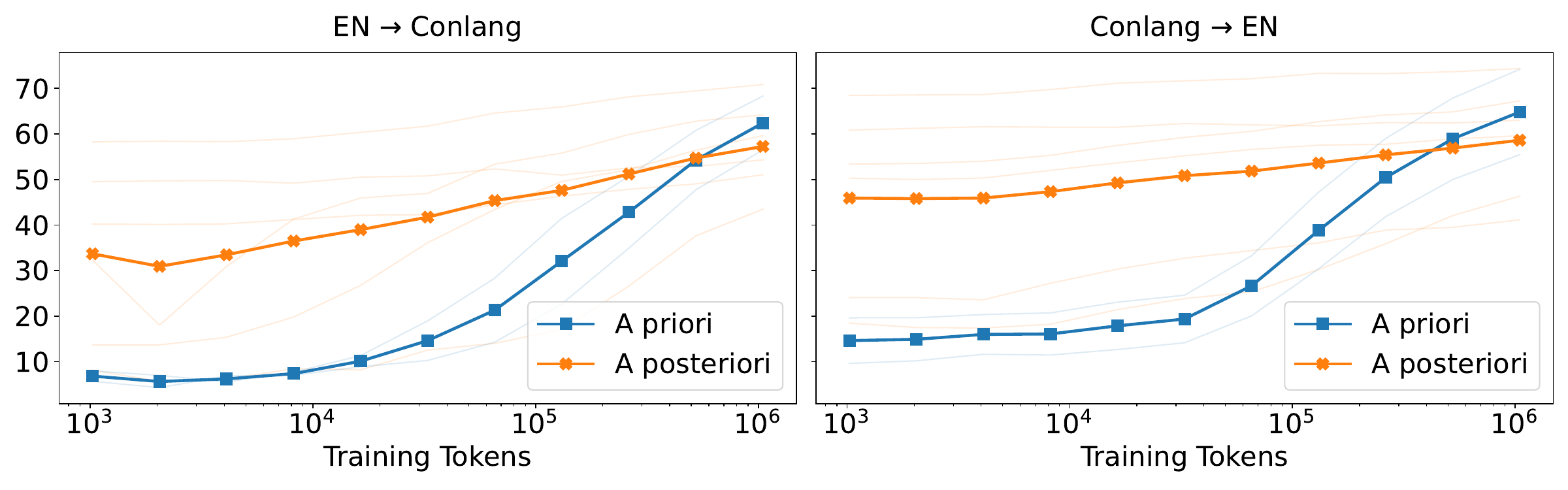}
    \caption{Translation performance (chrF++) of Qwen3.5-9B during training on the ConlangBench parallel corpus as a function of training tokens (log scale). In both translation directions, a posteriori conlangs start with relatively high scores and improve smoothly, whereas a priori conlangs remain at low performance during the first 16K training tokens before increasing steeply, exhibiting distinct learning trajectories.}
    \label{fig:training_results}
\end{figure*}

How do LLMs learn existing conlangs through parallel corpus training? In this section, we examine how LLMs acquire translation capabilities through training on the ConlangBench corpora, analyzing how learning curves vary across conlang categories and the number of training tokens.

\subsection{Methodology}

We train LLMs on parallel corpora for eight conlangs (Lojban, jbo; Toki Pona, tok; Quenya, qya; Klingon, tlh; Esperanto, epo; Ido, ido; Interlingua, ina; and Lingua Franca Nova, lfn), for which we have collected sufficient training samples (over 20,000 sentence pairs each).\footnote{Unless otherwise specified, the results in \S\ref{sec:training_models_on_conlang_parallel_corpora} are based on Qwen3.5-9B. Generalization experiments on other base models are presented in Appendix~\ref{sec:model_generalization}.} We fine-tune models using bidirectional translation samples from the training split of the ConlangBench dataset, using a separate LoRA~\citep{hu2022lora} adapter for each conlang.

During training, each sample is duplicated into conlang-to-English and English-to-conlang translation prompts, and the loss is computed only over the target response tokens. Since the conlang corpus sizes vary, we track performance with respect to the number of training tokens included in the loss. Model checkpoints are saved at geometrically increasing numbers of target-language training tokens, using powers of two (i.e., 1,024, 2,048, ..., final). After training, the model checkpoints are evaluated with the same metric (chrF++) as in \S\ref{sec:benchmarking_conlang_translation}.

\subsection{Results}

Experimental results in Figure~\ref{fig:training_results} and Table~\ref{tab:training_result_table} demonstrate that language models can learn all eight conlangs from the parallel corpora. However, the results also reveal distinct learning curves across conlang categories.

\begin{table}[htb]
\centering
\small
\begin{tabular}{lrrrr}
    \toprule
    \textbf{Conlang} & \multicolumn{2}{c}{\textbf{EN $\rightarrow$ Conlang}} & \multicolumn{2}{c}{\textbf{Conlang $\rightarrow$ EN}} \\
    \cmidrule(lr){2-3} \cmidrule(lr){4-5}
     & \textbf{Base} & \textbf{Trained} & \textbf{Base} & \textbf{Trained} \\
    \midrule
    Lojban (jbo) & 9.0 & \textbf{43.5} & 19.0 & \textbf{46.3} \\
    Toki Pona (tok) & 13.3 & \textbf{59.6} & 23.2 & \textbf{41.1} \\
    \midrule
    Quenya (qya) & 8.6 & \textbf{68.3} & 19.6 & \textbf{74.2} \\
    Klingon (tlh) & 7.9 & \textbf{56.5} & 9.7 & \textbf{55.4} \\
    \midrule
    Esperanto (epo) & 49.5 & \textbf{54.3} & 60.5 & \textbf{63.0} \\
    Ido (ido) & 37.5 & \textbf{51.0} & 50.0 & \textbf{59.6} \\
    Interlingua (ina) & 58.3 & \textbf{70.8} & 68.3 & \textbf{74.3} \\
    \makecell[l]{Lingua Franca\\Nova (lfn)} & 30.3 & \textbf{64.2} & 53.4 & \textbf{67.2} \\
    \bottomrule
\end{tabular}
\caption{chrF++ scores for the base and trained (1,048,576-token) checkpoints. The Qwen3.5-9B model can learn all conlangs from parallel corpora, including Lojban (jbo) and Toki Pona (tok), whose linguistic designs differ fundamentally from those of natural human languages~\citep{lang2014toki, cowan2016complete}.}
\label{tab:training_result_table}
\end{table}

\paragraph{Learning Curve Differences by Conlang Category.}

Figure~\ref{fig:training_results} shows that a priori and a posteriori conlangs in our training set exhibit markedly different learning trajectories. A posteriori conlangs begin with relatively high scores (e.g., 49.5 for English-to-Esperanto) and improve steadily, with mean gains of +24.3 points for English-to-conlang translation and +12.9 points for conlang-to-English translation. In contrast, performance on a priori conlangs remains low during the first 16K training tokens before increasing sharply, with mean gains of +54.1 and +50.2 points. These trends suggest that lexical knowledge transferred from previously learned natural languages benefits initial performance, whereas continued training on parallel corpora gradually diminishes this advantage.

\begin{table}[htb]
\centering
\small
\begin{tabular}{ll}
    \toprule
    \textbf{Checkpoint} & \textbf{EN $\rightarrow$ Toki Pona Translation Output} \\
    \midrule
    \multirow{4}{*}{Reference} & \textit{kala linja (mute) li lon tomo tawa telo mi} \\
    & \textcolor{gray}{`My hovercraft is full of eels'} \\
    & \textcolor{gray}{Lit. `(Many) rope fish are in my}\\
    & \textcolor{gray}{water-moving house.'} \\
    \midrule
    \multirow{2}{*}{0 (Base)} & \textit{mi li pona e eels} \\
    & \textcolor{gray}{`I fix ``eels''.'} \\
    \multirow{2}{*}{4,096} & \textit{mi li wile e tomo pilin.} \\
    & \textcolor{gray}{`I want an emotional house.'} \\
    \multirow{2}{*}{16,384} & \textit{tenpo suno li lon pi telo pi telo pi telo} \\
    & \textcolor{gray}{`A day is in water of water of water.'} \\
    \multirow{2}{*}{65,536} & \textit{tomo mi li jo e eel mute} \\
    & \textcolor{gray}{`My house has many ``eel''s.'} \\
    \multirow{2}{*}{262,144} & \textit{tomo tawa telo mi li jo e kala linja} \\
    & \textcolor{gray}{`My water-moving house has a rope fish.'} \\
    \bottomrule
\end{tabular}
\caption{Example English-to-Toki Pona translations at each training checkpoint. The model gradually learns a unique characteristic of Toki Pona by developing the ability to decompose complex concepts (e.g., ``hovercraft'') into combinations of its limited vocabulary.}
\label{tab:training_checkpoint_examples}
\end{table}

\paragraph{Acquisition of Conlang Characteristics.}

We select a Toki Pona (tok) sentence as a case study to illustrate how LLMs acquire the unique characteristics of individual conlangs. Toki Pona has a deliberately restricted vocabulary of approximately 130 words, which requires speakers to decompose concepts into combinations of basic meanings. Consistent with this, the model trained on Toki Pona achieves the lowest conlang-to-English score (41.1) in Table~\ref{tab:training_result_table}. This design is unique to conlangs, as natural languages are not intentionally created with such restrictions. Table~\ref{tab:training_checkpoint_examples} traces the model's outputs across training checkpoints. Although the base model appears to have some knowledge of Toki Pona, it produces an ungrammatical sentence (e.g., \textit{mi li}) while directly reproducing the English word \textit{eels}. During training, the model gradually learns the circumlocution strategy of Toki Pona, successfully decomposing \textit{hovercraft} into \textit{tomo tawa telo} `water-moving house' and \textit{eel} into \textit{kala linja} `rope fish' after training on 262,144 tokens.\footnote{Toki Pona follows a head-modifier order, with \textit{tomo} meaning `house' and \textit{kala} meaning `fish'.}

\subsection{Internalization of Conlang Lexicons}

To trace models' lexical knowledge acquisition, we examine the alignment between the internal representations of conlang words and those of their English meanings during training.

We analyze the trained model checkpoints for the eight conlangs using a relative representation similarity measure adapted from \citet{wu2025semantic}. For each conlang, we sample $N=100$ (\textit{word}, \textit{meaning}) pairs from the vocabulary data and extract the final-token hidden states of each word and meaning at every layer. Using the normalized representations
$\hat{\mathbf{x}}_i^{(\ell)}$ and
$\hat{\mathbf{y}}_i^{(\ell)}$, we compute the mean similarities of matched (positive) and mismatched (negative) pairs and define the layer-wise lexical alignment score $S^{(\ell)} = S_{\mathrm{pos}}^{(\ell)} - S_{\mathrm{neg}}^{(\ell)}$, where

\[
     S_{\mathrm{pos}}^{(\ell)} = \displaystyle
    \frac{1}{N}\sum_{i=1}^{N}
    \hat{\mathbf{x}}_i^{(\ell)\top}
    \hat{\mathbf{y}}_i^{(\ell)},
\]
\[
    S_{\mathrm{neg}}^{(\ell)} = \displaystyle
    \frac{1}{N(N-1)}
    \sum_{i\ne j}
    \hat{\mathbf{x}}_i^{(\ell)\top}
    \hat{\mathbf{y}}_j^{(\ell)}.
\]

\begin{figure}[!tb]
  \centering
  \includegraphics[width=\columnwidth]{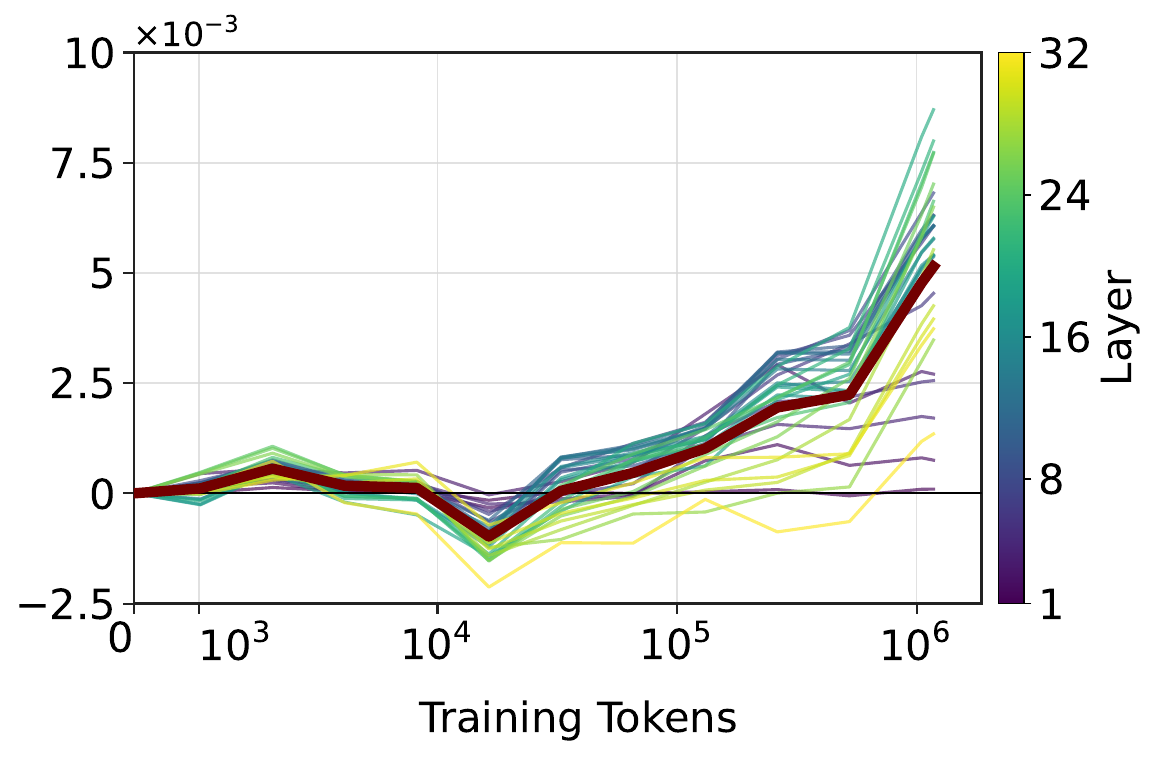}
  \caption{Layer-wise increases in conlang--English lexical alignment score relative to the base model during training. Scores are computed from $N=100$ sampled (\textit{word}, \textit{meaning}) pairs per conlang and averaged across the conlangs. Each of the 32 lines represents one model layer, while the bold line shows the mean across layers.}
  \label{fig:layerwise_sim}
\end{figure}

Figure~\ref{fig:layerwise_sim} reports the increase in lexical alignment relative to the base model, $S^{(\ell)}_t - S^{(\ell)}_{\mathrm{base}}$, as a function of training tokens. Its upward trend suggests that, as training proceeds, the model learns increasingly similar representations for conlang words and their English meanings. Details are provided in Appendix~\ref{sec:conlang_training_results}.

\section{Conclusion}

We introduce ConlangBench, the first large-scale benchmark for evaluating and training LLMs on 21 existing conlangs. ConlangBench comprises 21M+ conlang--English parallel sentence pairs, including 430K+ pairs from 20 non-Esperanto conlangs, and 321K+ vocabulary entries. Experimental results demonstrate that translation performance is associated with both natural-language similarity and conlang characteristics. Trained models also exhibit distinct learning trajectories across conlang categories as they internalize conlang knowledge. Our study suggests that conlangs have the potential to serve as a valuable testbed for investigating low-resource language acquisition in LLMs.

\section*{Limitations}

\paragraph{Resource Constraints and Corpus Imbalance.}

Since modern conlang creators and speakers rely on relatively small-scale online communities, publicly available parallel corpora are very limited. Despite this constraint, we manually collect as much data as possible, resulting in eight of the 21 conlangs with sufficient data for corpus training in \S\ref{sec:training_models_on_conlang_parallel_corpora}. We also observe substantial imbalance in resource availability across conlangs. In particular, Esperanto is far more widely used than other conlangs, providing over 20 million sentence pairs, compared with a combined total of approximately 430K sentence pairs for all 20 other conlangs. Our dataset also includes conlangs with extremely limited available data, such as Dutton World Speedwords and Neo. Therefore, evaluation results for these conlangs should be interpreted with caution.

\paragraph{Limited Semantic and Syntactic Evaluation Metrics for Conlangs.}

To the best of our knowledge, no reliable metrics have been proposed for evaluating semantic or syntactic similarity between generated texts and their corresponding reference texts in various existing conlangs, or between texts in conlangs and those in natural languages such as English. Therefore, we primarily report chrF++ results, a metric based on surface-level character and word overlap. Nevertheless, we demonstrate that chrF++ scores strongly correlate with BERTScore values based on semantic embeddings for round-trip translations in Appendix~\ref{sec:correlation_between_metrics}, supporting the use of chrF++.

\paragraph{Challenges in Conlang Classification.}

Although we adopt widely used classification criteria based on creation purpose and manner following \citet{schreyer2021constructed}, there has been limited discussion of how individual conlangs should be assigned to these categories. For example, we classify several conlangs commonly described as ``philosophical languages'' such as Toki Pona as engelangs, although they may also be classified as artlangs. Furthermore, a posteriori conlangs can be classified based on the degree to which they borrow and modify vocabulary from natural languages. Volapük and Lojban are examples of this because they exhibit low surface similarity to natural languages.

\paragraph{Beyond English-Centric Benchmarking.}

We translate all non-English natural language sentences into English to create conlang--English sentence pairs in ConlangBench. However, given the typological diversity of conlangs, evaluating translation between conlangs and non-English natural languages, such as French or Chinese, may reveal phenomena beyond those reported in this paper. In addition, concerns may arise regarding data quality because portions of the dataset are automatically translated using LLMs. We mitigate this issue through an analysis presented in Appendix~\ref{sec:discussion_translation}.

\section*{Ethical Considerations}

\paragraph{Copyright Considerations for Conlangs.}

Although whether conlangs as linguistic systems are subject to copyright protection remains legally disputed, the moral rights of conlang creators over their original creations should be respected. This is particularly important for artlangs, which often play a significant role as components of creators’ literary works, films, and fictional worlds. Therefore, research on understanding and generating existing conlangs using LLMs should be conducted in ways that respect and promote the rights of their creators and speaker communities.

\paragraph{Language Neutrality.}

No human language is inherently superior or inferior to any other. Similarly, this paper does not seek to rank conlangs according to their artistic merit. We instead evaluate and analyze LLMs across conlangs with different creation purposes and manners. Accordingly, the LLM translation performance reported in this paper should not be interpreted as an assessment of the quality or value of the conlangs themselves.




\bibliography{custom}

\appendix

\section{Experimental Settings}

\subsection{Training Environment}

We use one NVIDIA RTX PRO 6000 Blackwell GPU for evaluating language models and one NVIDIA H200 GPU for training language models. Training Qwen3.5-9B on all eight conlangs to 1,048,576 tokens takes approximately 7 GPU hours on an NVIDIA H200.

\subsection{List of Language Models Used}

The list below provides information on the LLMs used in the evaluation and training experiments.

\begin{itemize}[leftmargin=*, labelsep=0.5em]
    \item OLMo-2-1124-7B~\citep{olmo20252olmo2furious}
    \item OLMo-2-1124-13B~\citep{olmo20252olmo2furious}
    \item Olmo-3.1-32B-Instruct~\citep{olmo2025olmo3}
    \item Qwen3.5-2B~\citep{qwen3.5}
    \item Qwen3.5-4B~\citep{qwen3.5}
    \item Qwen3.5-9B~\citep{qwen3.5}
    \item Qwen3.5-27B~\citep{qwen3.5}
    \item gemma-4-E2B-it~\citep{gemmateam2026gemma4}
    \item gemma-4-E4B-it~\citep{gemmateam2026gemma4}
    \item gemma-4-31B-it~\citep{gemmateam2026gemma4}
    \item gemini-3-flash-preview~\citep{googledeepmind2025gemini3flash}
    \item gpt-5.5~\citep{openai2026gpt55}
\end{itemize}

\subsection{Hyperparameters}

\begin{itemize}[leftmargin=*, labelsep=0.5em]
    \item Evaluation configuration.
    \begin{itemize}
        \item Temperature: 0
        \item Maximum output tokens: $\min(2048,\\ \max(32, 5 \cdot \texttt{reference\_token\_length}))$
    \end{itemize}
    \item Training configuration.
    \begin{itemize}
        \item LoRA settings.
        \begin{itemize}
            \item Rank: 16
            \item Alpha: 32
            \item Dropout: 0.05
            \item Target modules: all-linear
        \end{itemize}
        \item Optimizer: adamw\_torch
        \item Learning rate: 2e-4
        \item Weight decay: 0.01
        \item Scheduler: constant (no warmup)
        \item Gradient clipping: 1.0
        \item Batch size: 64
    \end{itemize}
    \item General configuration.
    \begin{itemize}
        \item Seed: 42
        \item Thinking mode: False
    \end{itemize}
\end{itemize}

\section{Datasets}
\label{sec:datasets}

\subsection{Dataset Source}

The following list provides the official websites, online communities, book references, and Hugging Face datasets used to manually construct the parallel corpora and vocabulary resources for ConlangBench, excluding data obtained from OPUS and Tatoeba. We carefully exclude texts whose creators have explicitly refused to allow their use for AI training.

\begin{itemize}[leftmargin=*, labelsep=0.5em]
    \item Engelangs
    \begin{itemize}
        \item Láadan (ldn)
        \begin{itemize}
            \item \url{http://laadanlanguage.org}
        \end{itemize}
        \item Lojban (jbo)
        \begin{itemize}
            \item \url{https://lensisku.lojban.org}
            \item \url{https://lojban.org}
            \item \url{https://lojban.pw}
            \item \url{https://huggingface.co/datasets/NetherQuartz/minecraft-translations}
            \item \url{https://huggingface.co/datasets/olpa/jbo-corpus}
            \item \url{https://huggingface.co/datasets/smuske/korpora}
        \end{itemize}
        \item Toki Pona (tok)
        \begin{itemize}
            \item \url{https://github.com/lipu-linku/sona}
            \item \url{http://antetokipona.infinityfreeapp.com}
            \item \url{https://huggingface.co/datasets/NetherQuartz/lipu-sewi}
            \item \url{https://huggingface.co/datasets/NetherQuartz/minecraft-translations}
            \item \url{https://omniglot.com/language/phrases/tokipona.htm}
        \end{itemize}
    \end{itemize}
    \item Artlangs
    \begin{itemize}
        \item Quenya (qya)
        \begin{itemize}
            \item \url{https://eldamo.org}
            \item \url{https://ardalambion.net}
            \item \url{https://quettali.org}
            \item \url{https://realelvish.net}
            \item \url{https://huggingface.co/datasets/NetherQuartz/minecraft-translations}
        \end{itemize}
        \item Sindarin (sjn)
        \begin{itemize}
            \item \url{https://eldamo.org}
            \item \url{https://elfdict.com}
            \item \url{https://elvish.org}
            \item \url{https://realelvish.net}
        \end{itemize}
        \item Klingon (tlh)
        \begin{itemize}
            \item The Klingon Dictionary~\citep{okrand1992klingon}
            \item \url{https://tlhingan.org}
            \item \url{https://huggingface.co/datasets/MihaiPopa-1/custom-klingon-33k}
            \item \url{https://huggingface.co/datasets/NetherQuartz/minecraft-translations}
        \end{itemize}
        \item Brithenig (bzt)
        \begin{itemize}
            \item \url{http://steen.free.fr}
        \end{itemize}
        \item Talossan (tzl)
        \begin{itemize}
            \item \url{https://oversteir.talossa.com}
            \item \url{https://wiki.talossa.com}
            \item \url{https://omniglot.com/language/phrases/talossan.htm}
        \end{itemize}
    \end{itemize}
    \item Auxlangs
    \begin{itemize}
        \item Kotava (avk)
        \begin{itemize}
            \item \url{https://kotava.org}
            \item \url{https://europalingua.eu/index.htm}
        \end{itemize}
        \item Afrihili (afh)
        \begin{itemize}
            \item Ni Afrihili Oluga~\citep{attobrah1970ni}
            \item \url{https://fiatlingua.org}
        \end{itemize}
        \item Dutton World Speedwords (dws)
        \begin{itemize}
            \item \url{http://www2.cmp.uea.ac.uk/~jrk/conlang.dir/Speedwords.dict}
        \end{itemize}
        \item Esperanto (epo)
        \begin{itemize}
            \item \url{http://denisowski.org/Esperanto/ESPDIC/espdic.txt}
            \item \url{https://huggingface.co/datasets/NetherQuartz/minecraft-translations}
        \end{itemize}
        \item Ido (ido)
        \begin{itemize}
            \item \url{http://idolinguo.org.uk}
            \item \url{https://huggingface.co/datasets/NetherQuartz/minecraft-translations}
            \item \url{https://ohchr.org}
        \end{itemize}
        \item Interglossa (igs)
        \begin{itemize}
            \item \url{http://glosa.org}
            \item \url{https://glosa.fandom.com}
        \end{itemize}
        \item Interlingue (ile)
        \begin{itemize}
            \item \url{https://occidental-lang.com}
        \end{itemize}
        \item Interlingua (ina)
        \begin{itemize}
            \item \url{http://denisowski.org/Interlingua/IEDICT/iedict.txt}
            \item \url{https://github.com/fuszenecker/InterlinguaEnglishDictionary}
            \item \url{https://interlinguamultilingue.blogspot.com}
            \item \url{https://ohchr.org}
        \end{itemize}
        \item Interslavic (isv)
        \begin{itemize}
            \item \url{https://interslavic-dictionary.com}
            \item \url{https://hackmd.io/@xlO6E2n-S5eX4tROzsbTpg}
            \item \url{http://steen.free.fr}
            \item \url{https://huggingface.co/datasets/NetherQuartz/minecraft-translations}
        \end{itemize}
        \item Lingua Franca Nova (lfn)
        \begin{itemize}
            \item \url{https://github.com/elefen/disionario}
            \item \url{https://elefen.org}
        \end{itemize}
        \item Neo (neu)
        \begin{itemize}
            \item Rapid Method of Neo~\citep{alfandari1966rapid}
            \item \url{https://evertype.com}
            \item \url{https://metrotel.co.uk/mmm/neo.html}
        \end{itemize}
        \item Novial (nov)
        \begin{itemize}
            \item \url{http://blahedo.org/novial}
        \end{itemize}
        \item Volapük (vol)
        \begin{itemize}
            \item \url{https://volap\"uk.com}
            \item \url{https://huggingface.co/datasets/NetherQuartz/minecraft-translations}
        \end{itemize}
    \end{itemize}
\end{itemize}

\subsection{Preprocessing Details}

This section provides additional details on the preprocessing of the ConlangBench parallel corpus.

We first retrieve corpora from the OPUS API for datasets containing the target conlangs. Tatoeba mirrors in OPUS are excluded, and the Tatoeba corpus is instead collected from the official dumps. We then merge the two resources and remove duplicate text pairs with identical (\textit{original\_language}, \textit{target\_language}, \textit{original\_text}, \textit{target\_text}) fields, as well as text pairs in which both sides are conlangs.\footnote{In this section, ``original language'' refers to the natural language for translation, while ``target language'' refers to the conlang.} The preprocessing pipeline consists of four stages.

In the first stage, we compute the frequency of each natural language in the translation corpus. We then remove duplicate text pairs with identical (\textit{target\_language}, \textit{target\_text}) fields, preferentially retaining pairs with English translations. If none of the duplicates has an English translation, we retain the pair whose natural language has the highest frequency in the dataset.

In the second stage, we incorporate datasets from Hugging Face and manually collected data from websites into our corpus. For manually collected data, the authors carefully verify that each source sentence is written in the target conlang and that the paired translation accurately reflects its meaning. We then remove sentence pairs in which the original and target texts are identical.

In the third stage, we translate all non-English natural language texts into English using gemma-4-31B-it, ensuring that the ConlangBench dataset consists exclusively of conlang--English parallel text pairs. Across the 21 conlangs, an average of 23.2\% of the samples in the final dataset contain English texts translated from other natural languages. The translation prompt is provided in Appendix~\ref{sec:prompts}.

In the fourth stage, we split parallel texts into parallel sentences. We use the NLTK \texttt{punkt\_tab} sentence tokenizer~\citep{bird-loper-2004-nltk} for natural language texts and regular expressions for conlang texts. To avoid alignment errors, sentence splitting is applied only to text pairs containing two or more English sentences and having the same number of English and conlang sentences after splitting. Otherwise, the original text pairs are retained.

We then remove near-duplicate sentence pairs and filter out erroneous pairs, including blank texts, pairs with excessively short or long texts on one side (i.e., $\frac{\mathrm{len}(\textit{conlang})}{\mathrm{len}(\textit{English})}\notin[0.2,5.0]$), pairs with nearly identical conlang and English texts, pairs whose conlang text is detected as English, and pairs whose English text is not detected as English. Finally, we truncate texts that contain 2,048 tokens or more.

\subsection{Vocabulary Collection}
\label{sec:vocabulary_collection}

The vocabulary collection agent is implemented using Codex with \texttt{GPT-5.6 Sol}, configured with \textit{high} reasoning effort. It follows a five-stage pipeline: (1) source traversal and acquisition, (2) document refinement, including OCR with MinerU~\citep{wang2024mineru}, (3) vocabulary extraction, (4) translation of glosses into English, and (5) exact duplicate removal and final validation. The complete agent prompt is provided in Table~\ref{tab:prompt_for_vocabulary_collection}. For the resulting vocabulary dataset, we run the collection agent twice, manually review the raw sources collected in each trial, and select the better result for each language. In total, the agent collects 321,858 (conlang \textit{word}, English \textit{meaning}) pairs across 21 conlangs; per-language statistics are reported in Table~\ref{tab:vocabulary_statistics}.

\begin{table}[!tb]
    \centering
    \small
    \begin{tabular}{llr}
        \toprule
        \textbf{Code} & \textbf{Conlang} & \textbf{\# Lemma} \\
        \midrule
        ldn & Láadan & 1,769 \\
        jbo & Lojban & 17,546 \\
        tok & Toki Pona & 178 \\
        \midrule
        qya & Quenya & 6,058 \\
        sjn & Sindarin & 3,207 \\
        tlh & Klingon & 1,427 \\
        bzt & Brithenig & 3,246 \\
        tzl & Talossan & 24,523 \\
        \midrule
        avk & Kotava & 29,519 \\
        afh & Afrihili & 1,221 \\
        dws & Dutton World Speedwords & 3,095 \\
        epo & Esperanto & 63,872 \\
        ido & Ido & 24,270 \\
        igs & Interglossa & 987 \\
        ile & Interlingue & 10,851 \\
        ina & Interlingua & 58,436 \\
        isv & Interslavic & 19,085 \\
        lfn & Lingua Franca Nova & 27,004 \\
        neu & Neo & 6,790 \\
        nov & Novial & 4,065 \\
        vol & Volapük & 14,709 \\
        \midrule
        \textbf{Total} & & \textbf{321,858} \\
        \bottomrule
    \end{tabular}
    \caption{Number of collected (\textit{word}, \textit{meaning}) pairs for each conlang.}
    \label{tab:vocabulary_statistics}
\end{table}

\subsection{Dataset Availability}

We will make the ConlangBench code and dataset publicly available online, to the extent possible.

\section{Correlation between Metrics}
\label{sec:correlation_between_metrics}

\begin{figure}[!tb]
    \centering
    \begin{subfigure}[!tb]{\linewidth}
        \centering
        \includegraphics[width=\linewidth]{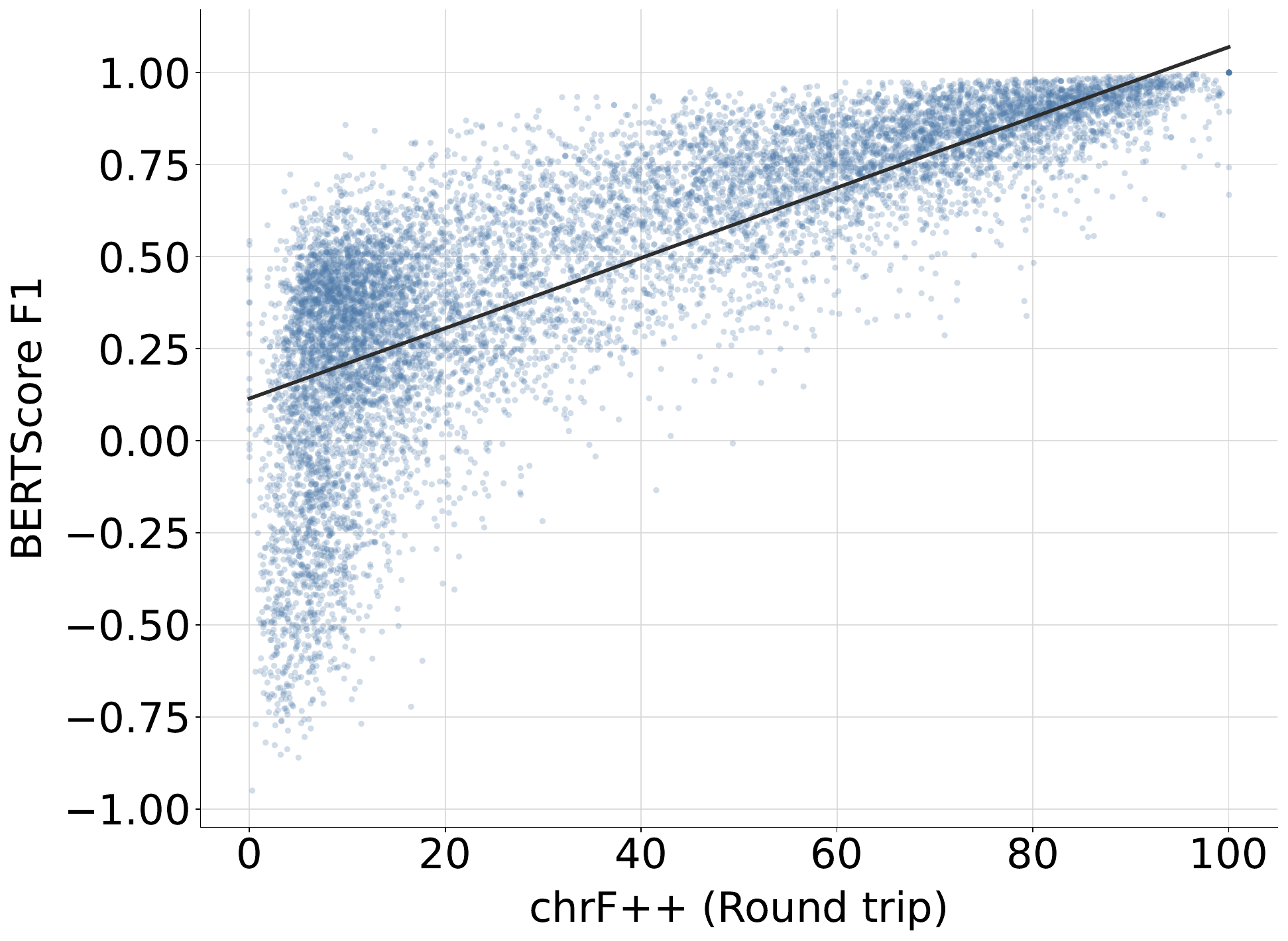}
        \caption{Correlation for the Qwen3.5-9B base model.}
        \label{fig:round_trip_metric_correlation_1}
    \end{subfigure}
    \par\vspace{0.5em}
    \begin{subfigure}[!tb]{\linewidth}
        \centering
        \includegraphics[width=\linewidth]{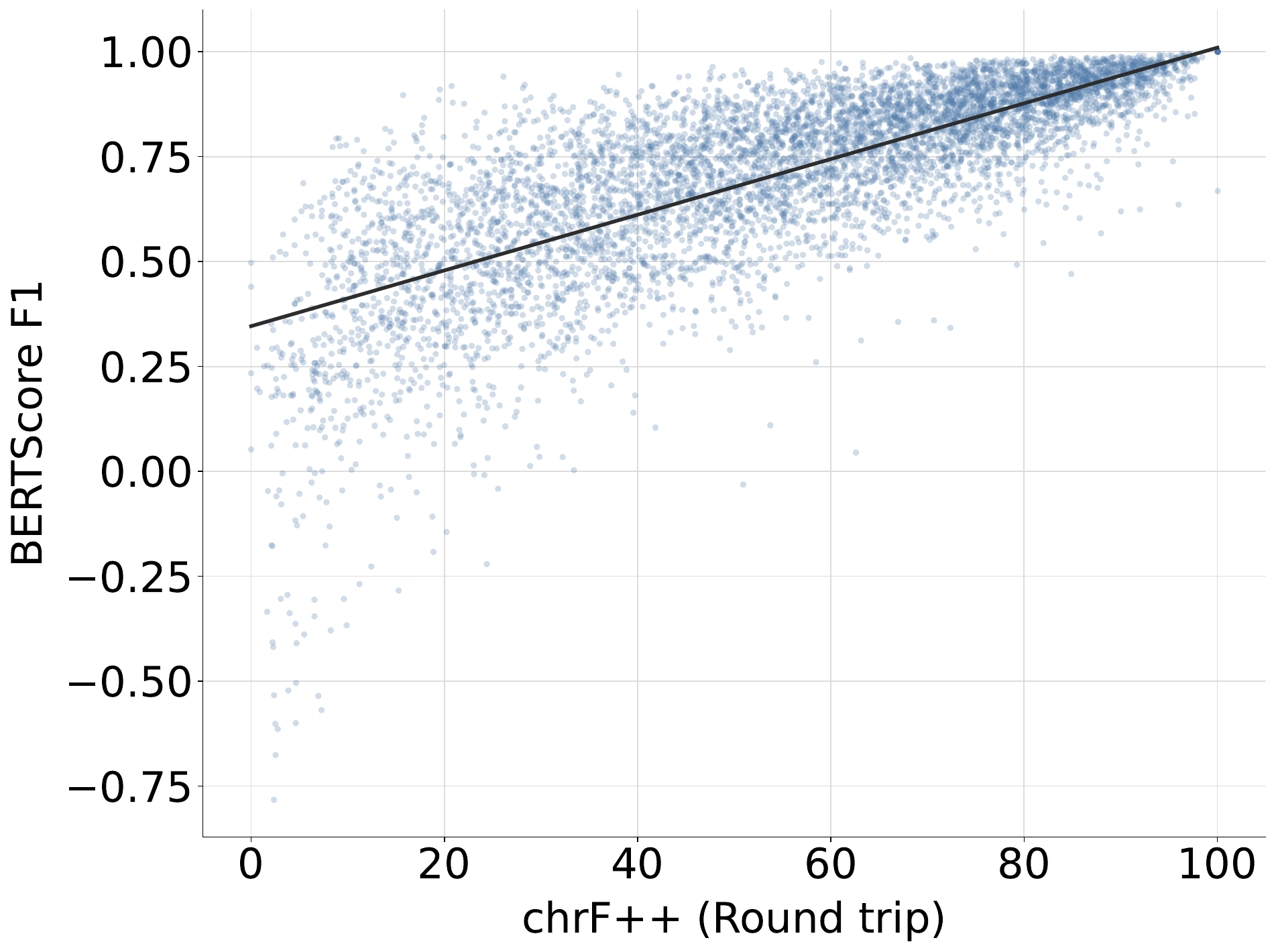}
        \caption{Correlation for the Qwen3.5-9B model trained on ConlangBench.}
        \label{fig:round_trip_metric_correlation_2}
    \end{subfigure}
    \caption{Correlation between round-trip chrF++ scores and round-trip BERTScore on the test split of ConlangBench.}
    \label{fig:round_trip_metric_correlation}
\end{figure}

A potential limitation of our evaluation is that chrF++ measures only surface-level lexical overlap and, unlike embedding-based metrics such as BERTScore, does not directly capture semantic similarity between sentences. Furthermore, to the best of our knowledge, no reliable semantic embedding models are currently available for the conlangs included in our benchmark.

Nevertheless, we provide indirect evidence that chrF++ scores for conlang translation are strongly associated with semantic similarity through a round-trip translation experiment. Experimental models first translate English sentences into a conlang and then translate the resulting conlang sentences back into English. We then compute the BERTScore between the round-trip English outputs and the reference English sentences. Because both BERTScore and chrF++ are evaluated on English text, this procedure avoids the need for conlang embedding models while allowing us to assess whether higher chrF++ scores on conlang translation correspond to better semantic preservation.

To account for the possibility that the base models occasionally copy the source conlang or English text instead of translating it, we report sample correlations for two settings: (1) the Qwen3.5-9B base model across all 21 conlangs and (2) the model trained with 1,048,576 tokens for the eight conlangs used in \S\ref{sec:training_models_on_conlang_parallel_corpora}.

Figure~\ref{fig:round_trip_metric_correlation} shows that round-trip chrF++ scores are strongly correlated with BERTScore for both models. The Qwen3.5-9B base model achieves a Pearson correlation coefficient of $r=0.823$, while the trained models achieve $r=0.831$. These results suggest that, although chrF++ measures only surface-level character and word overlap, it is also meaningful for assessing semantic similarity in conlang translation.

\section{Discussion}

\subsection{Does LLM-based Dataset Translation Affect the Results?}
\label{sec:discussion_translation}

On average, 23.2\% of the non-English natural language translation samples for each conlang are automatically translated into English. Therefore, it is important to verify that this LLM-based translation process does not harm the quality of our experimental results. To assess this, we compute the correlation between conlang--English translation scores obtained from samples originally written in English and those translated from non-English natural languages.

As shown in Table~\ref{tab:non_english_translation_correlation}, all models used in our experiments exhibit remarkably high correlations in conlang-wise translation performance between samples originally written in English and those translated into English from non-English natural languages. These results suggest that the two sets of samples produce highly consistent evaluation outcomes. Therefore, the use of LLM-translated English samples is unlikely to introduce systematic bias into our evaluation results.

\begin{table}[!tb]
\centering
\small
\begin{tabular}{lrr}
    \toprule
    \textbf{Model} & \multicolumn{2}{c}{\textbf{Correlation (Pearson $r$)}} \\
    \cmidrule(lr){2-3}
     & \textbf{EN $\rightarrow$ Conlang} & \textbf{Conlang $\rightarrow$ EN} \\
    \midrule
    OLMo-2-7B & 0.957 & 0.877 \\
    OLMo-2-13B & 0.930 & 0.831 \\
    Olmo-3.1-32B & 0.974 & 0.851 \\
    Qwen3.5-2B & 0.957 & 0.937 \\
    Qwen3.5-4B & 0.978 & 0.958 \\
    Qwen3.5-9B & 0.979 & 0.935 \\
    Qwen3.5-27B & 0.981 & 0.923 \\
    gemma-4-E2B & 0.953 & 0.870 \\
    gemma-4-E4B & 0.965 & 0.870 \\
    gemma-4-31B & 0.978 & 0.891 \\
    gemini-3-flash & 0.958 & 0.833 \\
    gpt-5.5 & 0.947 & 0.821 \\
    \bottomrule
\end{tabular}
\caption{Pearson correlation in conlang-wise translation performance between samples originally written in English and those translated into English from non-English natural languages.}
\label{tab:non_english_translation_correlation}
\end{table}

\subsection{Is chrF++ Biased Toward A Posteriori Conlangs?}

chrF++ measures surface-level character and word overlap. Therefore, a posteriori conlangs may receive inflated scores even when a model simply copies the English source sentence, as their vocabularies are largely derived from natural languages. To account for this effect, we report a source-copy baseline in Figure~\ref{fig:translation_scores_base}. Even after subtracting this baseline, the resulting $\Delta$ chrF++ scores remain consistently higher for a posteriori than for a priori conlangs in both open-weight and proprietary models, as shown in Table~\ref{tab:translation_scores_base_delta}.

\begin{table}[!tb]
\centering
\small
\begin{tabular}{llrr}
    \toprule
    \textbf{Model} & \textbf{Origin} & \multicolumn{2}{c}{\textbf{Macro-Average $\Delta$ chrF++}} \\
    \cmidrule(lr){3-4}
     & & \textbf{EN $\rightarrow$ CL} & \textbf{CL $\rightarrow$ EN} \\
    \midrule
    \multirow{2}{*}{Open} & Pri. & -1.8 & 4.5 \\
    & Post. & \textbf{5.9} & \textbf{20.1} \\
    \multirow{2}{*}{Proprietary} & Pri. & 15.6 & 31.1 \\
    & Post. & \textbf{22.6} & \textbf{33.8} \\
    \bottomrule
\end{tabular}
\caption{$\Delta$ chrF++ translation scores of open-weight and proprietary LLMs for a priori (Pri.) and a posteriori (Post.) conlangs (CL).}
\label{tab:translation_scores_base_delta}
\end{table}

\begin{figure*}[!tb]
    \centering
    \includegraphics[width=\linewidth]{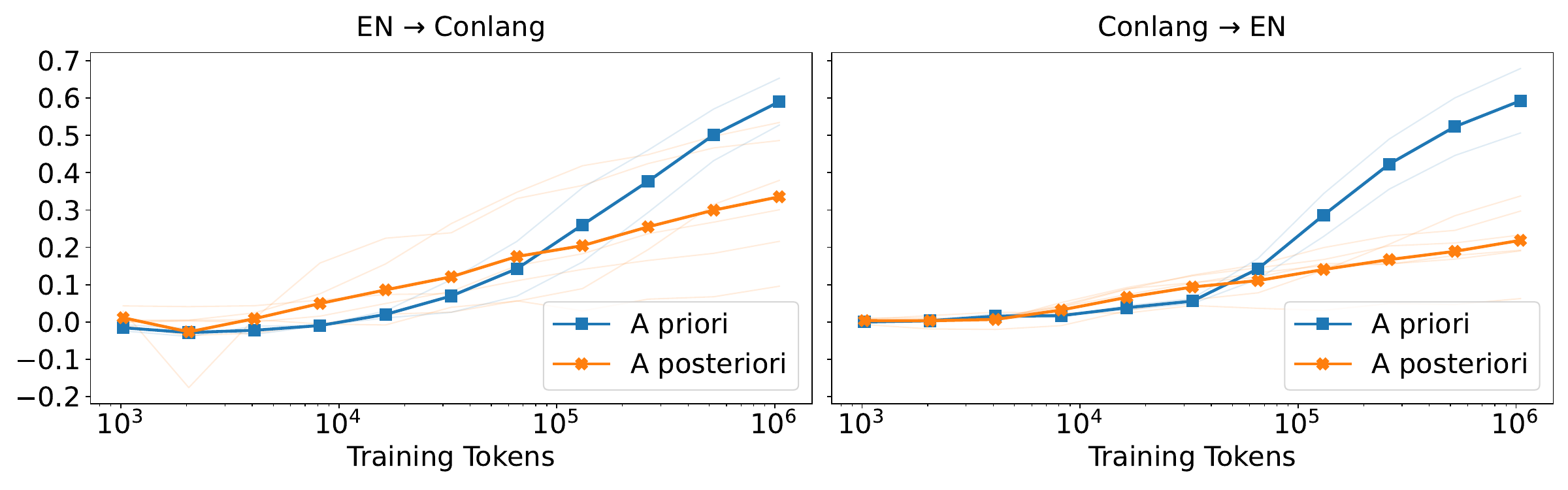}
    \caption{Hake's gain scores (chrF++) of Qwen3.5-9B during training on the ConlangBench parallel corpus as a function of training tokens (log scale).}
    \label{fig:training_results_hakes_gain}
\end{figure*}

\subsection{Are the Differences in Learning Trajectories Due to the Ceiling Effect?}

Since a priori conlangs generally exhibit lower initial translation performance than a posteriori conlangs, the observed differences in learning trajectories across conlang categories may merely reflect a ceiling effect, with lower-performing conlangs having greater room for improvement. To rule out this possibility, we recompute the learning curves using Hake's normalized gain (Equation~\ref{eq:hakes_gain}) for each trained checkpoint, which accounts for the remaining room for improvement. Figure~\ref{fig:training_results_hakes_gain} shows that the differences in learning trajectories remain evident even after controlling for this effect, indicating that our findings cannot be explained solely by a ceiling effect.

\begin{equation}
g =
\frac{S_{\mathrm{trained}} - S_{\mathrm{base}}}
{S_{\max} - S_{\mathrm{base}}}
\label{eq:hakes_gain}
\end{equation}

\subsection{Does the Manner of Creation Explain the Differences in Learning Trajectories Regardless of Creation Purpose?}

\begin{figure}[!tb]
    \centering
    \begin{subfigure}[htb]{\linewidth}
        \centering
        \includegraphics[width=\linewidth]{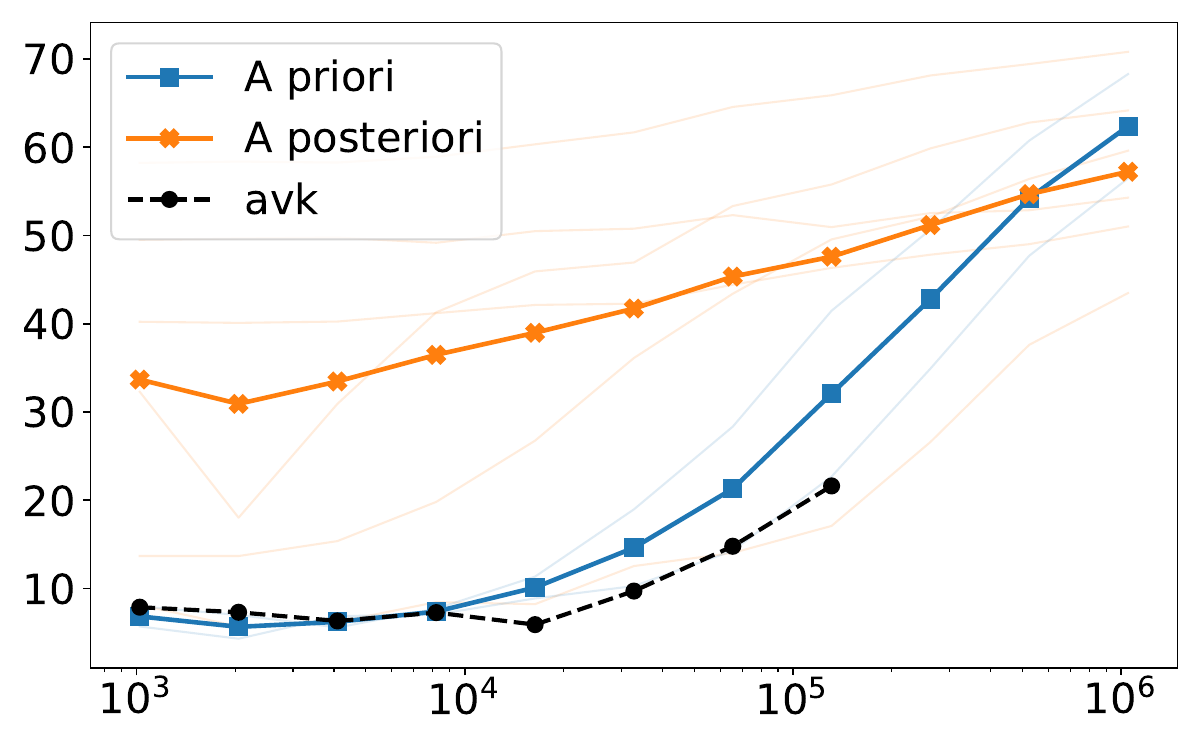}
        \caption{Learning trajectories for English-to-conlang translation.}
        \label{fig:training_highlight_avk_1}
    \end{subfigure}
    \par\vspace{0.5em}
    \begin{subfigure}[htb]{\linewidth}
        \centering
        \includegraphics[width=\linewidth]{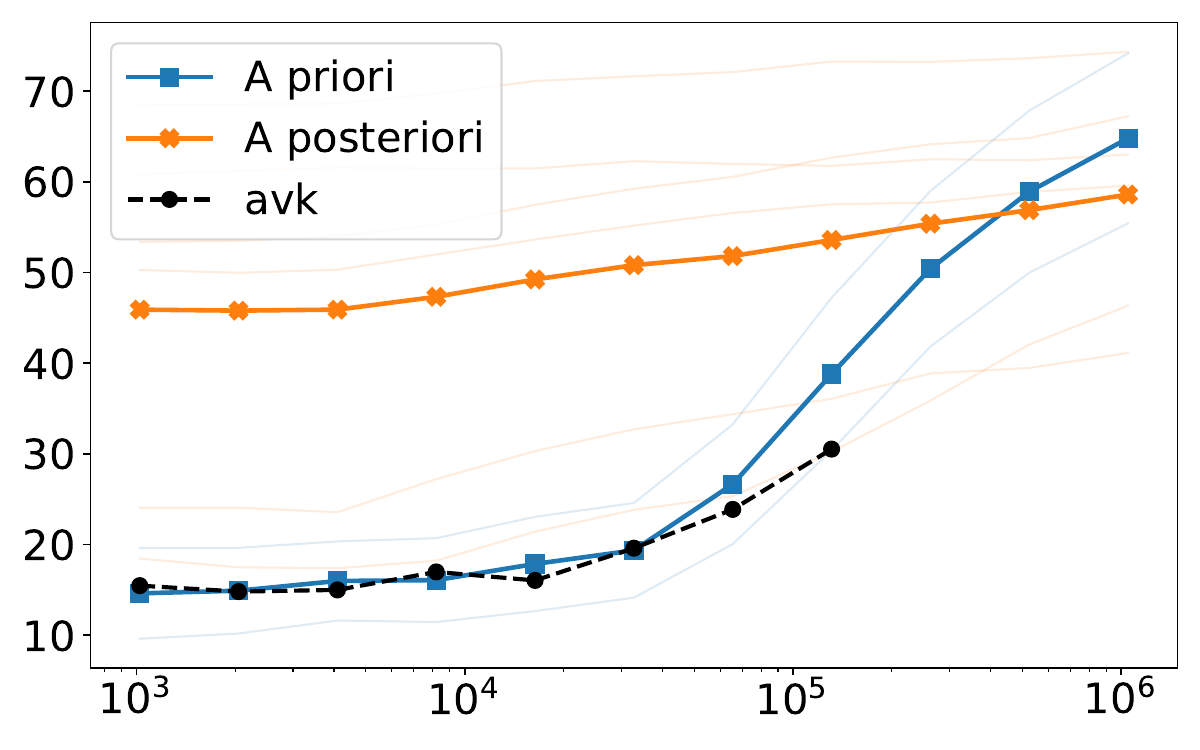}
        \caption{Learning trajectories for conlang-to-English translation.}
        \label{fig:training_highlight_avk_2}
    \end{subfigure}
    \caption{Learning trajectories of Qwen3.5-9B with the curve for an additional a priori auxlang, Kotava (avk).}
    \label{fig:training_highlight_avk}
\end{figure}

Due to data scarcity, only the two a priori artlangs, Quenya and Klingon, are used in the training experiments in \S\ref{sec:training_models_on_conlang_parallel_corpora}. This experimental design ensures a fair comparison by evaluating conlangs with the same amount of training data. To further examine whether the observed learning trajectories depend on the manner of creation rather than the creation purpose, we additionally train Qwen3.5-9B on Kotava (avk), the a priori auxlang with the largest amount of available training data that is not included in the main experiments, and overlay its learning curve on the existing results. As shown in Figure~\ref{fig:training_highlight_avk}, although Kotava's learning curve ends earlier because of its limited training data, it closely follows the average trajectory of the a priori artlangs. This result provides additional evidence that learning trajectories are primarily associated with the manner of creation rather than the creation purpose.

\section{Model Generalization}
\label{sec:model_generalization}

\begin{figure*}[!tb]
    \centering
    \begin{subfigure}[!tb]{\textwidth}
        \centering
        \includegraphics[width=\textwidth]{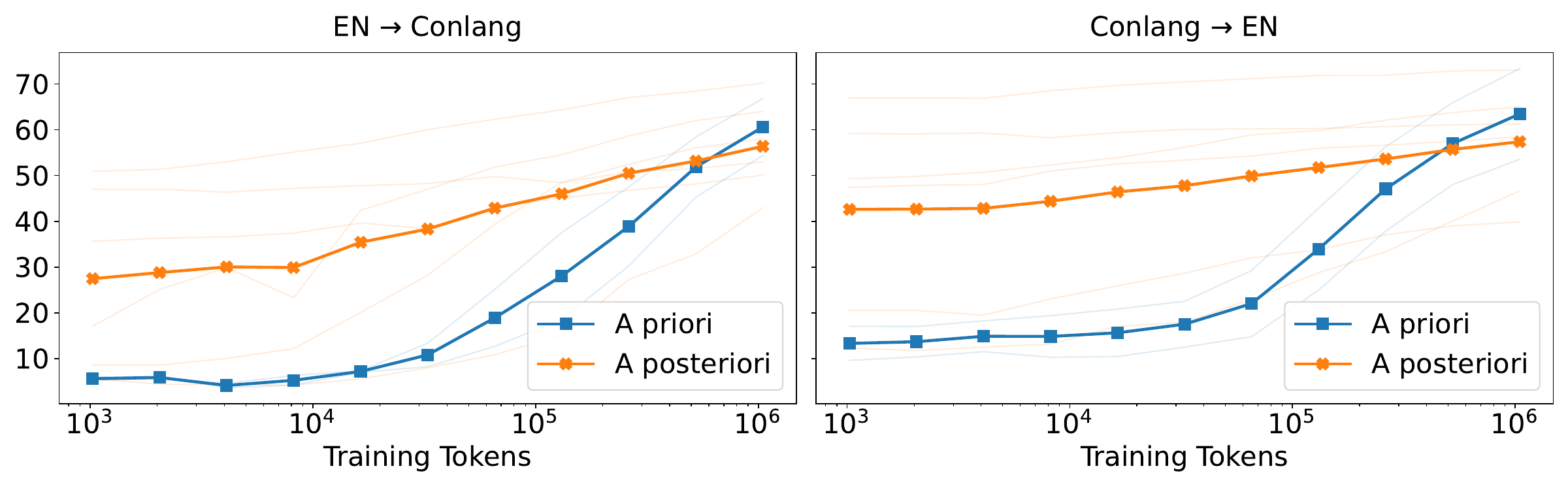}
        \caption{Learning trajectories for Qwen3.5-4B.}
        \label{fig:model_generalization_1}
    \end{subfigure}
    \par\vspace{0.5em}
    \begin{subfigure}[!tb]{\textwidth}
        \centering
        \includegraphics[width=\textwidth]{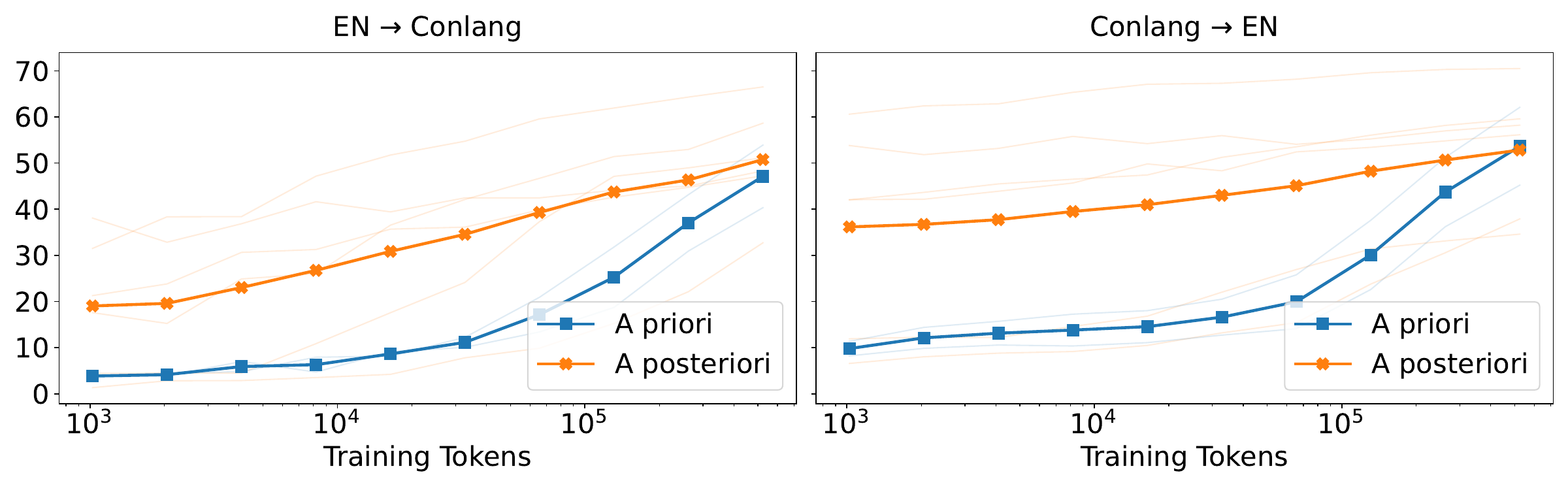}
        \caption{Learning trajectories for Qwen3.5-2B.}
        \label{fig:model_generalization_2}
    \end{subfigure}
    \par\vspace{0.5em}
    \begin{subfigure}[!tb]{\textwidth}
        \centering
        \includegraphics[width=\textwidth]{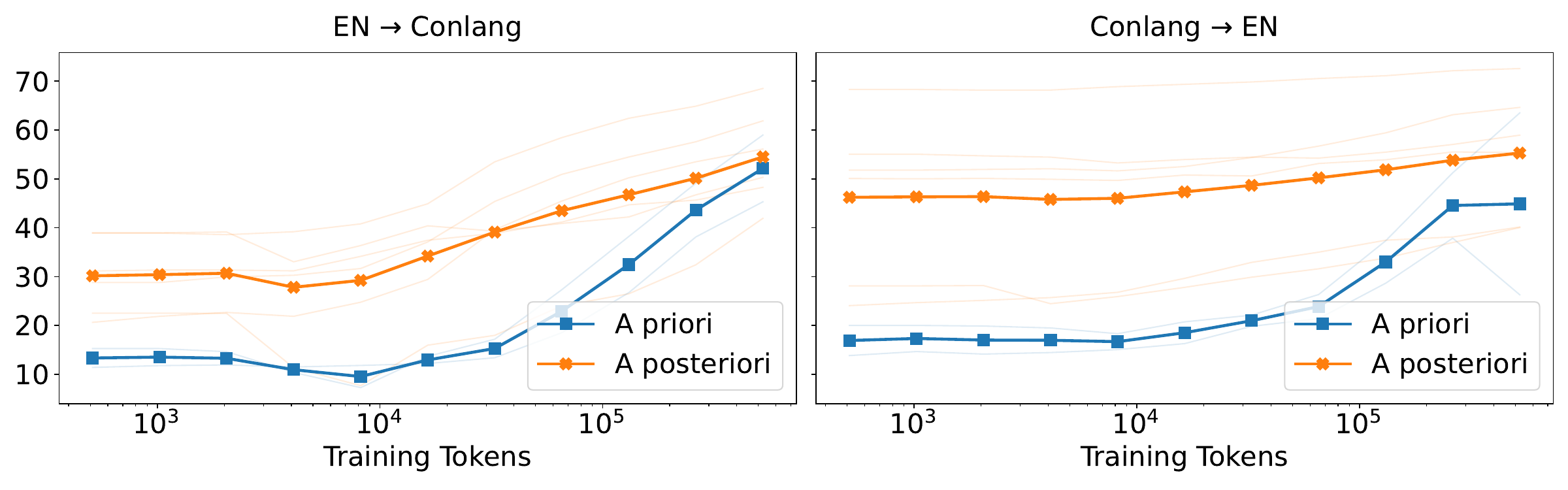}
        \caption{Learning trajectories for OLMo-2-1124-13B-Instruct.}
        \label{fig:model_generalization_3}
    \end{subfigure}
    \caption{Translation performance (chrF++) of various LLMs during training on the ConlangBench parallel corpus as a function of training tokens (log scale).}
    \label{fig:model_generalization}
\end{figure*}

Figure~\ref{fig:model_generalization} presents the results of our generalization experiments, in which different models are trained on conlang--English parallel corpora. As observed for Qwen3.5-9B in \S\ref{sec:training_models_on_conlang_parallel_corpora}, other models also exhibit distinct learning trajectories for a priori and a posteriori conlangs.

\section{Detailed Results}

\subsection{Conlang Translation Results}
\label{sec:conlang_translation_results}

\begin{figure*}[!tb]
    \centering
    \begin{subfigure}[!tb]{\textwidth}
        \centering
        \includegraphics[width=\textwidth]{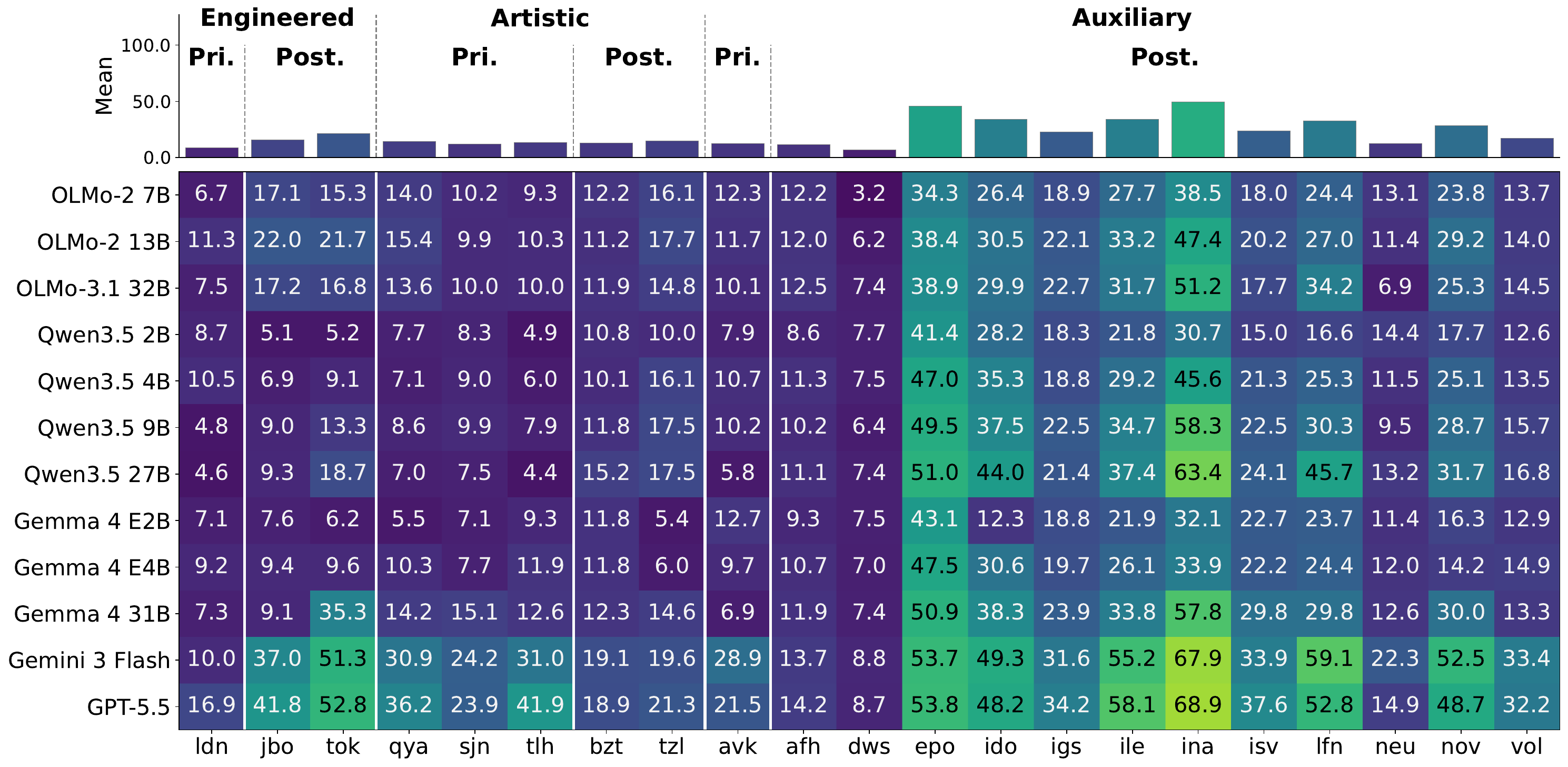}
        \caption{English-to-conlang translation scores (chrF++).}
        \label{fig:conlang_translation_heatmap_1}
    \end{subfigure}
    \par\vspace{0.5em}
    \begin{subfigure}[!tb]{\textwidth}
        \centering
        \includegraphics[width=\textwidth]{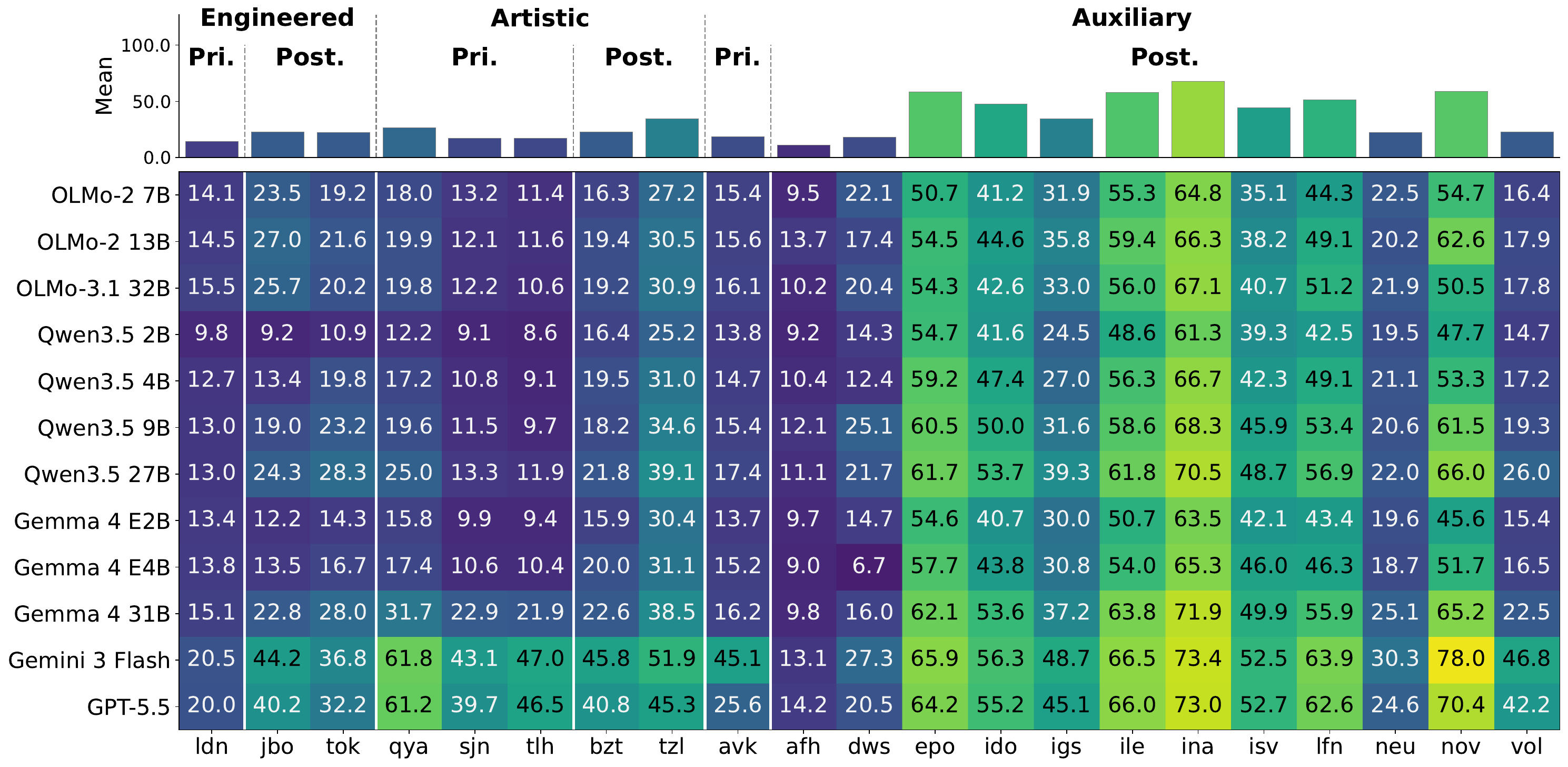}
        \caption{Conlang-to-English translation scores (chrF++).}
        \label{fig:conlang_translation_heatmap_2}
    \end{subfigure}
    \caption{Heatmap of chrF++ translation scores for 12 LLMs (10 open-weight and 2 proprietary models) across 21 existing conlangs.}
    \label{fig:conlang_translation_heatmap}
\end{figure*}

Figure~\ref{fig:conlang_translation_heatmap} illustrates detailed translation scores across conlangs and language models. Table~\ref{tab:example_translations_klingon} and \ref{tab:example_translations_interlingua} show examples of generated conlang translations for each model.

\begin{table*}[tb]
\centering
\small
\begin{tabularx}{\columnwidth}{ll}
    \toprule
    \textbf{Model} & \textbf{EN $\rightarrow$ Klingon (tlh)} \\
    \midrule
    \multicolumn{2}{l}{\emph{Source:} ``A woman is reading.''} \\
    \multicolumn{2}{l}{\emph{Reference:} \textit{laDtaH be'.}} \\
    \midrule
    OLMo-2-7B      & nuqneH \\
    OLMo-2-13B     & nuqneH 'oHbe' \\
    Olmo-3.1-32B   & jIH qIbtaH \\
    \midrule
    gemma-4-E2B    & tlh'a' tlhIngan:pu' tlhInganpu'pu' \\
    gemma-4-E4B    & tlhInganpu' Duj vIghaj. \\
    gemma-4-31B    & be'nI' ghoS. \\
    \midrule
    Qwen3.5-2B     & K'ha'na'na'na'na'na'na'na'na'na'\ldots \\
    Qwen3.5-4B     & QaH qo'pH. \\
    Qwen3.5-9B     & QaD nIj. \\
    Qwen3.5-27B    & qap' qap' qap' qap' qap' qap'\ldots \\
    \midrule
    gemini-3-flash & \textbf{laDtaH be'.} \\
    gpt-5.5        & \textbf{laDtaH be'.} \\
    \bottomrule
\end{tabularx}
\caption{Example results of English-to-Klingon translation for Klingon, an a priori artlang, where open-weight models fail to generate meaningful Klingon sentences. Instead, they produce well-known Klingon fragments such as \textit{nuqneH} `Hello' and \textit{tlhIngan} `Klingon' without compositional control.}
\label{tab:example_translations_klingon}
\end{table*}

\begin{table*}[tb]
\centering
\small
\begin{tabularx}{\columnwidth}{ll}
    \toprule
    \textbf{Model} & \textbf{EN $\rightarrow$ Interlingua (ina)} \\
    \midrule
    \multicolumn{2}{l}{\emph{Source:} ``This country is small.''} \\
    \multicolumn{2}{l}{\emph{Reference:} \textit{Iste pais es parve.}} \\
    \midrule
    OLMo-2-7B      & Esta landa es petite. \\
    OLMo-2-13B     & Hoc pais es petite. \\
    Olmo-3.1-32B   & Iste patria es parva. \\
    \midrule
    gemma-4-E2B    & Questo país è piccolo. \\
    gemma-4-E4B    & Haec patria est parva. \\
    gemma-4-31B    & \textbf{Iste pais es parve.} \\
    \midrule
    Qwen3.5-2B     & Ista landa est parva. \\
    Qwen3.5-4B     & Le pais es petit. \\
    Qwen3.5-9B     & Ille pais es parve. \\
    Qwen3.5-27B    & \textbf{Iste pais es parve.} \\
    \midrule
    gemini-3-flash & \textbf{Iste pais es parve.} \\
    gpt-5.5        & \textbf{Iste pais es parve.} \\
    \bottomrule
\end{tabularx}
\caption{Example results of English-to-Interlingua translation for Interlingua, an a posteriori auxlang, showing progressive improvement with model size. Smaller models tend to generate a sentence in typologically related natural languages such as Latin and French. As model size increases, the outputs converge toward the correct Interlingua translation.}
\label{tab:example_translations_interlingua}
\end{table*}

\subsection{Conlang Training Results}
\label{sec:conlang_training_results}

Table~\ref{tab:training_sweep_table_qwen35_9b}, \ref{tab:training_sweep_table_qwen35_4b}, \ref{tab:training_sweep_table_qwen35_2b}, and \ref{tab:training_sweep_table_olmo2_13b} provide detailed chrF++ scores for the training experiments. We also report the means and standard deviations for each conlang category (a priori and a posteriori). For Klingon (tlh), the total number of training tokens does not reach 1,048,576 under the tokenizers of Qwen3.5-2B and OLMo-2-1124-13B-Instruct, so the corresponding results are omitted from the tables.

Figure~\ref{fig:layerwise_similarities_all} reports changes in lexical alignment scores relative to the base model, \(S^{(\ell)}_t - S^{(\ell)}_{\mathrm{base}}\), for the eight conlangs across 32 layers of Qwen3.5-9B. Each panel corresponds to one layer, and each curve shows one conlang's lexical alignment trajectory as a function of training tokens.

\begin{table*}[t]
\centering
\small
\begin{tabular}{lrrrrrrrr@{\hspace{1.5em}}cc}
    \toprule
    \textbf{Tokens} & \textbf{jbo} & \textbf{tok} & \textbf{qya} & \textbf{tlh} & \textbf{epo} & \textbf{ido} & \textbf{ina} & \textbf{lfn} & \textbf{A priori} ($\pm$SD) & \textbf{A posteriori} ($\pm$SD) \\
    \midrule
    \multicolumn{11}{l}{\textit{EN $\rightarrow$ Conlang (chrF++)}} \\
    \midrule
    Base & 9.0 & 13.3 & 8.6 & 7.9 & 49.5 & 37.5 & 58.3 & 30.3 & 8.3 ($\pm$0.5) & 33.0 ($\pm$19.5) \\
    1{,}024 & 8.1 & 13.7 & 8.0 & 5.7 & 49.5 & 40.2 & 58.2 & 32.3 & 6.8 ($\pm$1.6) & 33.7 ($\pm$19.8) \\
    2{,}048 & 5.7 & 13.7 & 7.0 & 4.3 & 49.7 & 40.1 & 58.4 & 18.0 & 5.7 ($\pm$1.9) & 30.9 ($\pm$21.4) \\
    4{,}096 & 6.2 & 15.4 & 5.6 & 6.9 & 49.8 & 40.3 & 58.3 & 30.9 & 6.2 ($\pm$0.9) & 33.5 ($\pm$20.0) \\
    8{,}192 & 8.5 & 19.8 & 7.7 & 7.1 & 49.2 & 41.2 & 58.9 & 41.3 & 7.4 ($\pm$0.5) & 36.5 ($\pm$18.8) \\
    16{,}384 & 8.2 & 26.8 & 11.3 & 8.9 & 50.5 & 42.1 & 60.3 & 45.9 & 10.1 ($\pm$1.7) & 39.0 ($\pm$18.7) \\
    32{,}768 & 12.5 & 36.1 & 19.0 & 10.3 & 50.8 & 42.3 & 61.7 & 46.9 & 14.6 ($\pm$6.1) & 41.7 ($\pm$16.7) \\
    65{,}536 & 14.0 & 43.4 & 28.3 & 14.3 & 52.3 & 44.4 & 64.6 & 53.3 & 21.3 ($\pm$9.9) & 45.3 ($\pm$17.1) \\
    131{,}072 & 17.1 & 49.6 & 41.5 & 22.7 & 51.0 & 46.3 & 65.9 & 55.8 & 32.1 ($\pm$13.3) & 47.6 ($\pm$16.4) \\
    262{,}144 & 26.6 & 52.1 & 50.7 & 34.9 & 52.5 & 47.8 & 68.1 & 59.9 & 42.8 ($\pm$11.1) & 51.2 ($\pm$14.0) \\
    524{,}288 & 37.6 & 56.4 & 60.8 & 47.7 & 52.9 & 49.0 & 69.4 & 62.8 & 54.2 ($\pm$9.2) & 54.7 ($\pm$11.1) \\
    1{,}048{,}576 & \textbf{43.5} & \textbf{59.6} & \textbf{68.3} & \textbf{56.5} & \textbf{54.3} & \textbf{51.0} & \textbf{70.8} & \textbf{64.2} & \textbf{62.4} ($\pm$8.4) & \textbf{57.2} ($\pm$9.7) \\
    \midrule
    \multicolumn{11}{l}{\textit{Conlang $\rightarrow$ EN (chrF++)}} \\
    \midrule
    Base & 19.0 & 23.2 & 19.6 & 9.7 & 60.5 & 50.0 & 68.3 & 53.4 & 14.6 ($\pm$7.0) & 45.7 ($\pm$20.1) \\
    1{,}024 & 18.5 & 24.1 & 19.6 & 9.6 & 60.8 & 50.3 & 68.4 & 53.4 & 14.6 ($\pm$7.1) & 45.9 ($\pm$20.2) \\
    2{,}048 & 17.5 & 24.1 & 19.6 & 10.2 & 61.2 & 50.0 & 68.6 & 53.5 & 14.9 ($\pm$6.7) & 45.8 ($\pm$20.5) \\
    4{,}096 & 17.4 & 23.6 & 20.4 & 11.6 & 61.6 & 50.3 & 68.6 & 54.0 & 16.0 ($\pm$6.2) & 45.9 ($\pm$20.8) \\
    8{,}192 & 18.2 & 27.2 & 20.7 & 11.5 & 61.5 & 52.0 & 69.7 & 55.3 & 16.1 ($\pm$6.5) & 47.3 ($\pm$20.2) \\
    16{,}384 & 21.4 & 30.3 & 23.1 & 12.7 & 61.5 & 53.6 & 71.1 & 57.5 & 17.9 ($\pm$7.4) & 49.2 ($\pm$19.2) \\
    32{,}768 & 23.8 & 32.7 & 24.6 & 14.1 & 62.3 & 55.2 & 71.6 & 59.2 & 19.4 ($\pm$7.4) & 50.8 ($\pm$18.5) \\
    65{,}536 & 25.3 & 34.4 & 33.3 & 20.1 & 62.0 & 56.6 & 72.1 & 60.6 & 26.7 ($\pm$9.3) & 51.8 ($\pm$18.0) \\
    131{,}072 & 30.2 & 36.1 & 47.2 & 30.4 & 61.8 & 57.5 & 73.3 & 62.7 & 38.8 ($\pm$11.9) & 53.6 ($\pm$16.8) \\
    262{,}144 & 35.9 & 38.9 & 59.0 & 41.8 & 62.5 & 57.7 & 73.2 & 64.1 & 50.4 ($\pm$12.1) & 55.4 ($\pm$14.9) \\
    524{,}288 & 42.1 & 39.5 & 67.9 & 50.0 & 62.4 & 58.9 & 73.6 & 64.8 & 58.9 ($\pm$12.6) & 56.9 ($\pm$13.4) \\
    1{,}048{,}576 & \textbf{46.3} & \textbf{41.1} & \textbf{74.2} & \textbf{55.4} & \textbf{63.0} & \textbf{59.6} & \textbf{74.3} & \textbf{67.2} & \textbf{64.8} ($\pm$13.3) & \textbf{58.6} ($\pm$12.6) \\
    \bottomrule
\end{tabular}
\caption{Translation scores of Qwen3.5-9B during training.}
\label{tab:training_sweep_table_qwen35_9b}
\end{table*}

\begin{table*}[t]
\centering
\small
\begin{tabular}{lrrrrrrrr@{\hspace{1.5em}}cc}
    \toprule
    \textbf{Tokens} & \textbf{jbo} & \textbf{tok} & \textbf{qya} & \textbf{tlh} & \textbf{epo} & \textbf{ido} & \textbf{ina} & \textbf{lfn} & \textbf{A priori} ($\pm$SD) & \textbf{A posteriori} ($\pm$SD) \\
    \midrule
    \multicolumn{11}{l}{\textit{EN $\rightarrow$ Conlang (chrF++)}} \\
    \midrule
    Base & 6.9 & 9.1 & 7.1 & 6.0 & 47.0 & 35.3 & 45.6 & 25.3 & 6.5 ($\pm$0.8) & 28.2 ($\pm$17.5) \\
    1{,}024 & 5.6 & 8.6 & 5.3 & 5.9 & 47.0 & 35.6 & 50.9 & 17.1 & 5.6 ($\pm$0.4) & 27.5 ($\pm$19.7) \\
    2{,}048 & 4.4 & 8.6 & 5.6 & 6.2 & 47.0 & 36.3 & 51.3 & 25.1 & 5.9 ($\pm$0.4) & 28.8 ($\pm$19.6) \\
    4{,}096 & 4.4 & 10.0 & 4.6 & 3.7 & 46.3 & 36.6 & 53.0 & 29.8 & 4.1 ($\pm$0.7) & 30.0 ($\pm$19.5) \\
    8{,}192 & 4.2 & 12.2 & 6.3 & 4.2 & 47.2 & 37.4 & 55.1 & 23.3 & 5.3 ($\pm$1.5) & 29.9 ($\pm$20.0) \\
    16{,}384 & 5.6 & 20.1 & 7.3 & 7.0 & 47.8 & 39.6 & 57.1 & 42.4 & 7.2 ($\pm$0.2) & 35.4 ($\pm$19.1) \\
    32{,}768 & 8.0 & 28.2 & 13.4 & 8.2 & 48.2 & 38.4 & 60.0 & 47.0 & 10.8 ($\pm$3.6) & 38.3 ($\pm$18.2) \\
    65{,}536 & 10.8 & 39.3 & 25.1 & 12.7 & 49.7 & 43.2 & 62.3 & 51.8 & 18.9 ($\pm$8.8) & 42.9 ($\pm$17.6) \\
    131{,}072 & 15.1 & 48.5 & 37.5 & 18.6 & 48.5 & 45.1 & 64.3 & 54.5 & 28.1 ($\pm$13.4) & 46.0 ($\pm$16.6) \\
    262{,}144 & 27.3 & 52.4 & 47.5 & 30.3 & 50.8 & 46.8 & 67.0 & 58.6 & 38.9 ($\pm$12.2) & 50.5 ($\pm$13.4) \\
    524{,}288 & 32.9 & 56.1 & 58.4 & 45.3 & 51.5 & 48.2 & 68.4 & 62.0 & 51.8 ($\pm$9.3) & 53.2 ($\pm$12.3) \\
    1{,}048{,}576 & \textbf{43.0} & \textbf{58.0} & \textbf{66.8} & \textbf{54.4} & \textbf{53.0} & \textbf{50.1} & \textbf{70.2} & \textbf{64.0} & \textbf{60.6} ($\pm$8.8) & \textbf{56.4} ($\pm$9.8) \\
    \midrule
    \multicolumn{11}{l}{\textit{Conlang $\rightarrow$ EN (chrF++)}} \\
    \midrule
    Base & 13.4 & 19.8 & 17.2 & 9.1 & 59.2 & 47.4 & 66.7 & 49.1 & 13.2 ($\pm$5.8) & 42.6 ($\pm$21.4) \\
    1{,}024 & 12.4 & 20.5 & 17.0 & 9.6 & 59.1 & 47.4 & 66.9 & 49.3 & 13.3 ($\pm$5.2) & 42.6 ($\pm$21.6) \\
    2{,}048 & 11.8 & 20.5 & 17.0 & 10.4 & 59.1 & 47.8 & 66.9 & 49.8 & 13.7 ($\pm$4.7) & 42.7 ($\pm$21.8) \\
    4{,}096 & 12.5 & 19.5 & 18.2 & 11.5 & 59.3 & 48.0 & 66.9 & 50.7 & 14.9 ($\pm$4.8) & 42.8 ($\pm$21.9) \\
    8{,}192 & 13.1 & 23.0 & 19.4 & 10.3 & 58.2 & 51.0 & 68.5 & 52.3 & 14.8 ($\pm$6.4) & 44.4 ($\pm$21.5) \\
    16{,}384 & 17.1 & 25.9 & 20.8 & 10.4 & 59.4 & 52.5 & 69.7 & 53.9 & 15.6 ($\pm$7.3) & 46.4 ($\pm$20.4) \\
    32{,}768 & 17.9 & 28.7 & 22.5 & 12.5 & 60.1 & 53.4 & 70.5 & 56.2 & 17.5 ($\pm$7.1) & 47.8 ($\pm$20.1) \\
    65{,}536 & 22.9 & 32.0 & 29.2 & 14.8 & 60.2 & 54.2 & 71.2 & 58.9 & 22.0 ($\pm$10.2) & 49.9 ($\pm$18.5) \\
    131{,}072 & 28.8 & 33.8 & 42.9 & 25.0 & 60.3 & 55.9 & 71.9 & 59.8 & 33.9 ($\pm$12.6) & 51.7 ($\pm$16.8) \\
    262{,}144 & 33.3 & 37.0 & 56.3 & 37.9 & 60.7 & 56.6 & 71.9 & 62.1 & 47.1 ($\pm$13.0) & 53.6 ($\pm$15.2) \\
    524{,}288 & 40.1 & 39.0 & 65.9 & 48.0 & 61.1 & 57.5 & 72.8 & 63.7 & 57.0 ($\pm$12.7) & 55.7 ($\pm$13.5) \\
    1{,}048{,}576 & \textbf{46.7} & \textbf{39.9} & \textbf{73.4} & \textbf{53.5} & \textbf{61.3} & \textbf{58.5} & \textbf{73.0} & \textbf{65.0} & \textbf{63.4} ($\pm$14.0) & \textbf{57.4} ($\pm$12.2) \\
    \bottomrule
\end{tabular}
\caption{Translation scores of Qwen3.5-4B during training.}
\label{tab:training_sweep_table_qwen35_4b}
\end{table*}

\begin{table*}[t]
\centering
\small
\begin{tabular}{lrrrrrrrr@{\hspace{1.5em}}cc}
    \toprule
    \textbf{Tokens} & \textbf{jbo} & \textbf{tok} & \textbf{qya} & \textbf{tlh} & \textbf{epo} & \textbf{ido} & \textbf{ina} & \textbf{lfn} & \textbf{A priori} ($\pm$SD) & \textbf{A posteriori} ($\pm$SD) \\
    \midrule
    \multicolumn{11}{l}{\textit{EN $\rightarrow$ Conlang (chrF++)}} \\
    \midrule
    Base & 5.1 & 5.2 & 7.7 & 4.9 & 41.4 & 28.2 & 30.7 & 16.6 & 6.3 ($\pm$2.0) & 21.2 ($\pm$14.7) \\
    1{,}024 & 1.3 & 4.4 & 4.0 & 3.7 & 38.1 & 21.3 & 31.5 & 17.6 & 3.8 ($\pm$0.2) & 19.0 ($\pm$14.5) \\
    2{,}048 & 2.8 & 4.5 & 4.4 & 3.9 & 32.8 & 23.8 & 38.4 & 15.2 & 4.2 ($\pm$0.4) & 19.6 ($\pm$14.7) \\
    4{,}096 & 2.9 & 4.5 & 4.8 & 7.0 & 36.9 & 30.6 & 38.4 & 24.9 & 5.9 ($\pm$1.6) & 23.0 ($\pm$15.7) \\
    8{,}192 & 3.5 & 10.9 & 7.9 & 4.7 & 41.6 & 31.3 & 47.2 & 26.0 & 6.3 ($\pm$2.3) & 26.7 ($\pm$17.0) \\
    16{,}384 & 4.2 & 17.6 & 8.2 & 9.1 & 39.4 & 35.7 & 51.7 & 36.5 & 8.6 ($\pm$0.7) & 30.9 ($\pm$17.0) \\
    32{,}768 & 7.8 & 24.1 & 12.2 & 10.1 & 42.5 & 36.1 & 54.7 & 42.1 & 11.1 ($\pm$1.5) & 34.6 ($\pm$16.5) \\
    65{,}536 & 9.9 & 37.3 & 20.9 & 13.4 & 42.5 & 39.8 & 59.6 & 46.7 & 17.2 ($\pm$5.3) & 39.3 ($\pm$16.4) \\
    131{,}072 & 15.3 & 47.1 & 31.8 & 18.7 & 44.1 & 42.6 & 61.9 & 51.4 & 25.3 ($\pm$9.3) & 43.7 ($\pm$15.6) \\
    262{,}144 & 22.1 & 49.0 & 43.2 & 30.9 & 45.0 & 44.7 & 64.3 & 53.0 & 37.0 ($\pm$8.6) & 46.4 ($\pm$13.9) \\
    524{,}288 & 32.7 & 51.1 & 53.9 & \textbf{40.3} & 48.3 & \textbf{47.2} & 66.5 & 58.6 & 47.1 ($\pm$9.6) & 50.8 ($\pm$11.4) \\
    1{,}048{,}576 & \textbf{40.7} & \textbf{55.6} & \textbf{63.9} & -- & \textbf{50.3} & 46.9 & \textbf{68.2} & \textbf{60.5} & \textbf{63.9} & \textbf{53.7} ($\pm$9.9) \\
    \midrule
    \multicolumn{11}{l}{\textit{Conlang $\rightarrow$ EN (chrF++)}} \\
    \midrule
    Base & 9.2 & 10.9 & 12.2 & 8.6 & 54.7 & 41.6 & 61.3 & 42.5 & 10.4 ($\pm$2.5) & 36.7 ($\pm$21.9) \\
    1{,}024 & 6.6 & 11.9 & 11.4 & 8.2 & 53.8 & 42.1 & 60.6 & 42.0 & 9.8 ($\pm$2.2) & 36.2 ($\pm$22.1) \\
    2{,}048 & 8.0 & 12.3 & 14.4 & 9.9 & 51.8 & 42.2 & 62.4 & 43.7 & 12.1 ($\pm$3.2) & 36.7 ($\pm$21.9) \\
    4{,}096 & 8.8 & 12.2 & 15.7 & 10.6 & 53.2 & 43.9 & 62.8 & 45.5 & 13.1 ($\pm$3.6) & 37.7 ($\pm$22.2) \\
    8{,}192 & 9.2 & 14.6 & 17.3 & 10.4 & 55.8 & 45.7 & 65.3 & 46.5 & 13.8 ($\pm$4.9) & 39.5 ($\pm$22.6) \\
    16{,}384 & 10.5 & 16.8 & 18.0 & 11.1 & 54.2 & 49.8 & 67.1 & 47.4 & 14.6 ($\pm$4.9) & 41.0 ($\pm$22.3) \\
    32{,}768 & 13.2 & 22.0 & 20.5 & 12.7 & 56.0 & 48.3 & 67.3 & 51.2 & 16.6 ($\pm$5.5) & 43.0 ($\pm$20.9) \\
    65{,}536 & 15.4 & 26.9 & 25.8 & 14.1 & 54.1 & 52.4 & 68.2 & 53.5 & 19.9 ($\pm$8.3) & 45.1 ($\pm$19.7) \\
    131{,}072 & 23.8 & 31.4 & 37.6 & 22.6 & 55.2 & 53.4 & 69.6 & 56.1 & 30.1 ($\pm$10.6) & 48.2 ($\pm$17.2) \\
    262{,}144 & 30.6 & 33.1 & 51.2 & 36.2 & 57.0 & 54.8 & 70.3 & 58.2 & 43.7 ($\pm$10.6) & 50.7 ($\pm$15.6) \\
    524{,}288 & 37.9 & 34.6 & 62.1 & \textbf{45.2} & \textbf{58.2} & \textbf{56.1} & 70.5 & 59.6 & 53.6 ($\pm$11.9) & 52.8 ($\pm$13.8) \\
    1{,}048{,}576 & \textbf{42.2} & \textbf{37.5} & \textbf{70.6} & -- & 57.5 & 55.1 & \textbf{71.0} & \textbf{62.2} & \textbf{70.6} & \textbf{54.2} ($\pm$12.5) \\
    \bottomrule
\end{tabular}
\caption{Translation scores of Qwen3.5-2B during training.}
\label{tab:training_sweep_table_qwen35_2b}
\end{table*}

\begin{table*}[t]
\centering
\small
\begin{tabular}{lrrrrrrrr@{\hspace{1.5em}}cc}
    \toprule
    \textbf{Tokens} & \textbf{jbo} & \textbf{tok} & \textbf{qya} & \textbf{tlh} & \textbf{epo} & \textbf{ido} & \textbf{ina} & \textbf{lfn} & \textbf{A priori} ($\pm$SD) & \textbf{A posteriori} ($\pm$SD) \\
    \midrule
    \multicolumn{11}{l}{\textit{EN $\rightarrow$ Conlang (chrF++)}} \\
    \midrule
    Base & 22.0 & 21.7 & 15.4 & 10.3 & 38.4 & 30.5 & 47.4 & 27.0 & 12.9 ($\pm$3.6) & 31.2 ($\pm$10.1) \\
    1{,}024 & 22.5 & 21.9 & 15.3 & 11.8 & 39.0 & 31.4 & 38.9 & 28.8 & 13.5 ($\pm$2.5) & 30.4 ($\pm$7.5) \\
    2{,}048 & 22.5 & 22.7 & 14.6 & 11.9 & 39.1 & 31.3 & 38.6 & 29.9 & 13.3 ($\pm$1.9) & 30.7 ($\pm$7.3) \\
    4{,}096 & 11.3 & 21.9 & 10.4 & 11.5 & 33.0 & 31.2 & 39.2 & 30.3 & 10.9 ($\pm$0.8) & 27.8 ($\pm$9.8) \\
    8{,}192 & 7.7 & 24.8 & 7.3 & 11.8 & 36.4 & 34.1 & 40.8 & 31.7 & 9.5 ($\pm$3.2) & 29.2 ($\pm$11.8) \\
    16{,}384 & 16.0 & 29.4 & 13.7 & 12.2 & 40.4 & 37.4 & 44.9 & 37.1 & 12.9 ($\pm$1.0) & 34.2 ($\pm$10.3) \\
    32{,}768 & 18.0 & 39.6 & 17.2 & 13.4 & 39.3 & 38.9 & 53.5 & 45.4 & 15.3 ($\pm$2.7) & 39.1 ($\pm$11.8) \\
    65{,}536 & 24.0 & 45.5 & 27.3 & 18.5 & 40.9 & 41.2 & 58.5 & 50.9 & 22.9 ($\pm$6.2) & 43.5 ($\pm$11.6) \\
    131{,}072 & 26.5 & 50.2 & 38.2 & 26.7 & 42.2 & 44.7 & 62.4 & 54.5 & 32.5 ($\pm$8.1) & 46.8 ($\pm$12.3) \\
    262{,}144 & 32.4 & 53.5 & 49.1 & 38.1 & 46.7 & 45.7 & 64.9 & 57.6 & 43.6 ($\pm$7.8) & 50.1 ($\pm$11.2) \\
    524{,}288 & 41.9 & 56.1 & 59.0 & \textbf{45.3} & 50.3 & 48.3 & 68.5 & 61.9 & 52.1 ($\pm$9.7) & 54.5 ($\pm$9.7) \\
    1{,}048{,}576 & \textbf{48.0} & \textbf{57.1} & \textbf{67.4} & -- & \textbf{51.5} & \textbf{50.1} & \textbf{69.8} & \textbf{64.8} & \textbf{67.4} & \textbf{56.9} ($\pm$8.8) \\
    \midrule
    \multicolumn{11}{l}{\textit{Conlang $\rightarrow$ EN (chrF++)}} \\
    \midrule
    Base & 27.0 & 21.6 & 19.9 & 11.6 & 54.5 & 44.6 & 66.3 & 49.1 & 15.7 ($\pm$5.8) & 43.9 ($\pm$16.9) \\
    1{,}024 & 28.1 & 24.7 & 20.0 & 14.7 & 55.1 & 50.0 & 68.3 & 51.8 & 17.3 ($\pm$3.8) & 46.3 ($\pm$16.8) \\
    2{,}048 & 28.2 & 25.1 & 19.9 & 14.1 & 54.7 & 50.1 & 68.2 & 51.9 & 17.0 ($\pm$4.1) & 46.4 ($\pm$16.6) \\
    4{,}096 & 24.5 & 25.7 & 19.5 & 14.4 & 54.5 & 49.9 & 68.2 & 52.1 & 17.0 ($\pm$3.6) & 45.8 ($\pm$17.3) \\
    8{,}192 & 25.9 & 26.8 & 18.3 & 15.0 & 53.3 & 49.7 & 68.9 & 51.6 & 16.7 ($\pm$2.3) & 46.0 ($\pm$16.7) \\
    16{,}384 & 27.8 & 29.6 & 20.7 & 16.3 & 54.0 & 50.8 & 69.4 & 52.5 & 18.5 ($\pm$3.2) & 47.3 ($\pm$15.9) \\
    32{,}768 & 29.9 & 32.9 & 22.1 & 19.8 & 54.5 & 50.6 & 69.8 & 54.4 & 21.0 ($\pm$1.6) & 48.7 ($\pm$15.0) \\
    65{,}536 & 31.6 & 35.0 & 26.3 & 21.4 & 54.3 & 53.2 & 70.6 & 56.7 & 23.9 ($\pm$3.5) & 50.2 ($\pm$14.6) \\
    131{,}072 & 33.8 & 37.4 & 37.3 & 28.6 & 55.5 & 53.9 & 71.1 & 59.4 & 33.0 ($\pm$6.1) & 51.9 ($\pm$14.0) \\
    262{,}144 & 37.0 & 38.1 & 51.2 & \textbf{37.9} & 57.1 & 55.5 & 72.2 & 63.1 & 44.6 ($\pm$9.5) & 53.8 ($\pm$13.9) \\
    524{,}288 & 40.0 & 40.2 & 63.5 & 26.3 & 59.0 & 55.4 & \textbf{72.6} & 64.6 & 44.9 ($\pm$26.3) & 55.3 ($\pm$13.1) \\
    1{,}048{,}576 & \textbf{44.4} & \textbf{42.3} & \textbf{71.6} & -- & \textbf{60.0} & \textbf{56.7} & 72.3 & \textbf{66.5} & \textbf{71.6} & \textbf{57.0} ($\pm$11.9) \\
    \bottomrule
\end{tabular}
\caption{Translation scores of OLMo-2-1124-13B-Instruct during training.}
\label{tab:training_sweep_table_olmo2_13b}
\end{table*}

\section{Prompts}
\label{sec:prompts}

Table~\ref{tab:prompt_for_dataset_translation}, \ref{tab:prompt_for_model_translation}, \ref{tab:prompt_for_model_translation_vocabulary}, and \ref{tab:prompt_for_vocabulary_collection} report the prompts used in this paper.

\section{Use of AI Assistants}

We use AI assistants for coding support, grammar correction, and polishing the wording of text written by the authors.

\clearpage
\begingroup
\newcommand{\layersimilaritypanel}[1]{%
    \begin{minipage}[t]{0.48\textwidth}
        \centering
        \includegraphics[width=\linewidth]{figures/files/layerwise-similarities/similarity-layer-#1.pdf}
    \end{minipage}%
}

\begin{figure*}[p]
    \centering
    \layersimilaritypanel{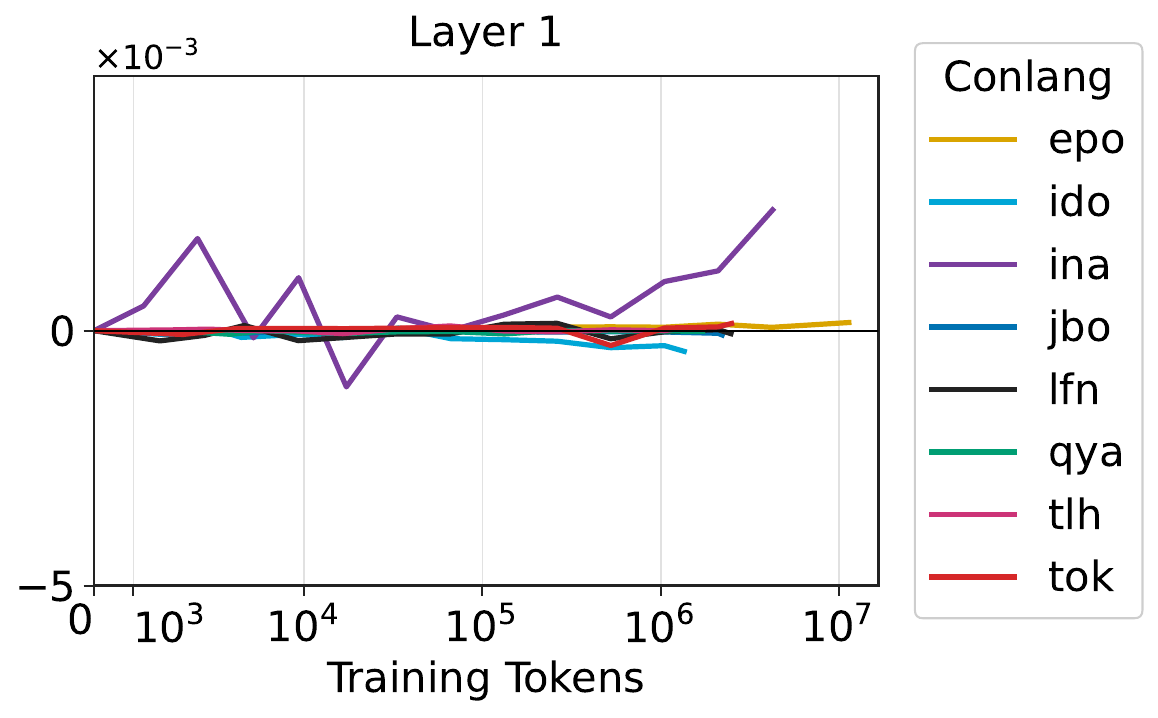}\hfill\layersimilaritypanel{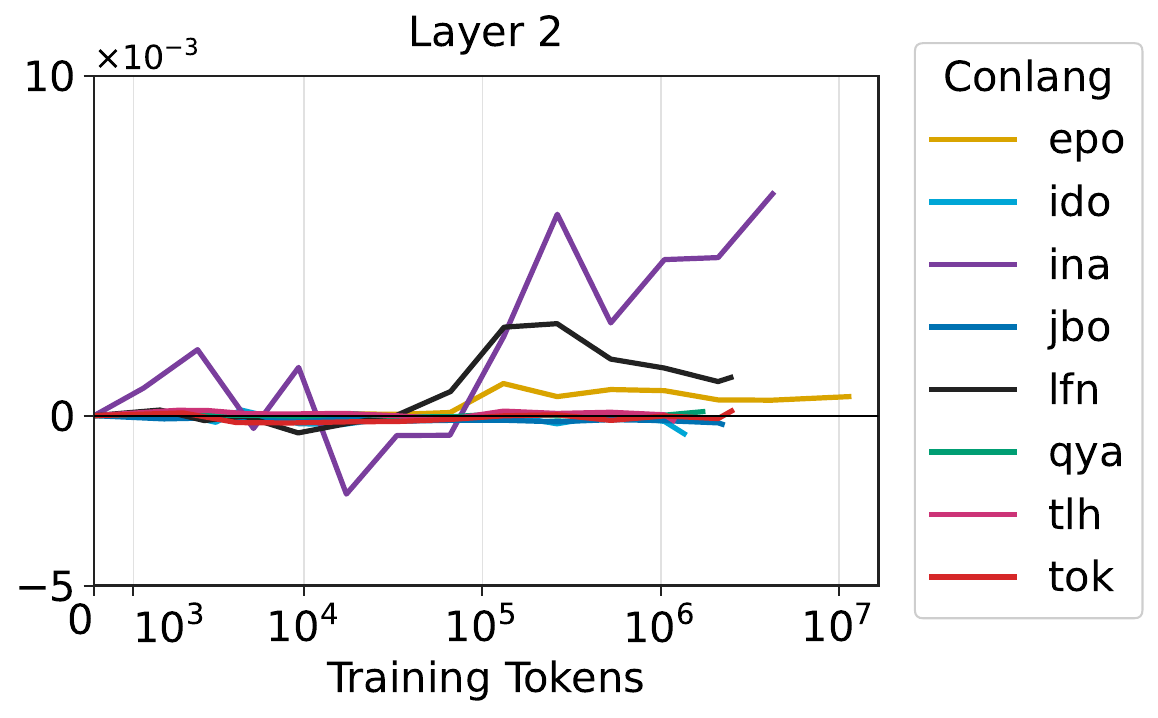}\par\medskip
    \layersimilaritypanel{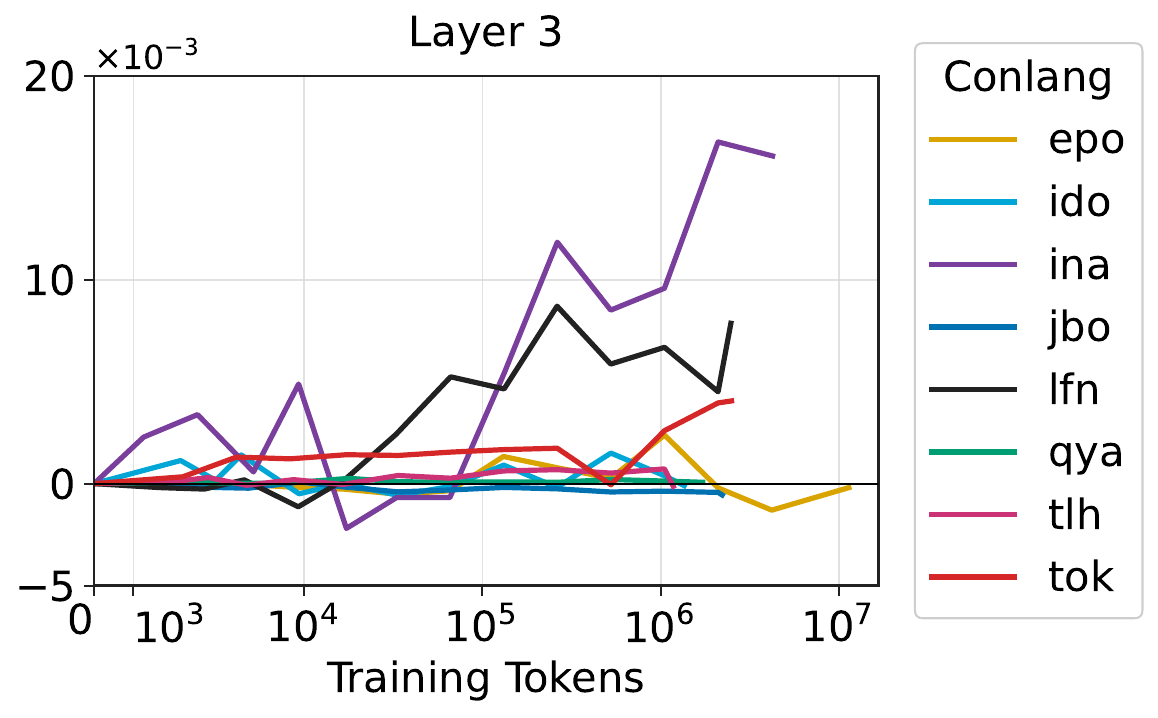}\hfill\layersimilaritypanel{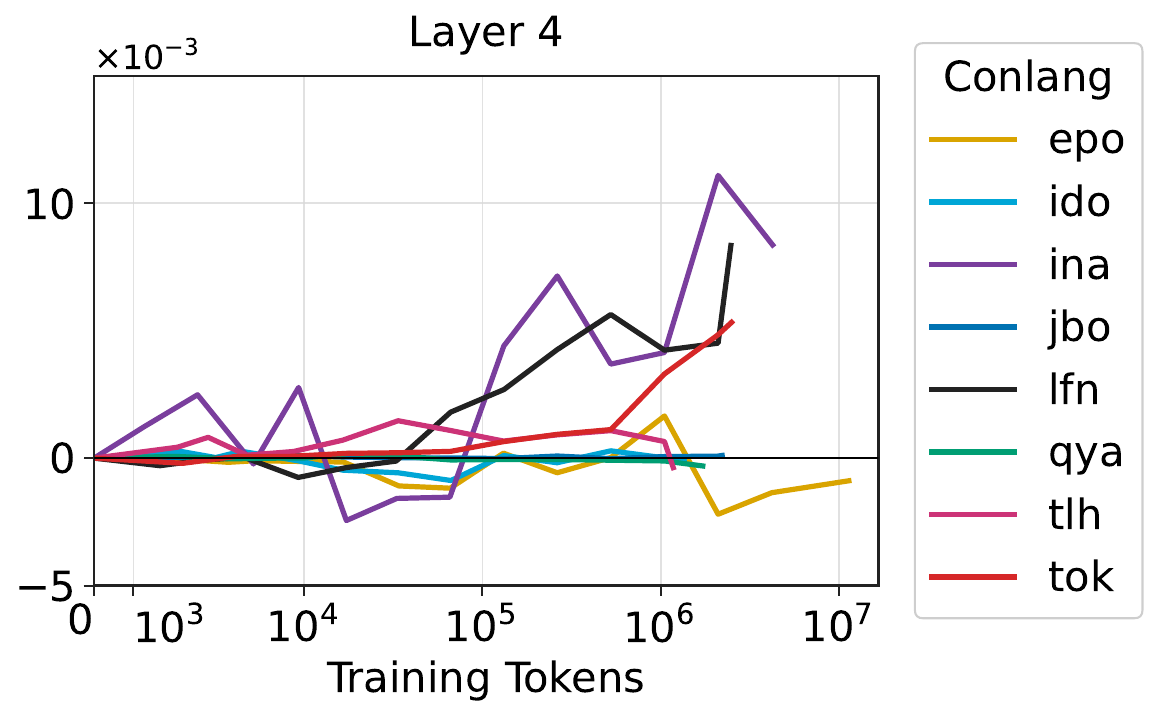}\par\medskip
    \layersimilaritypanel{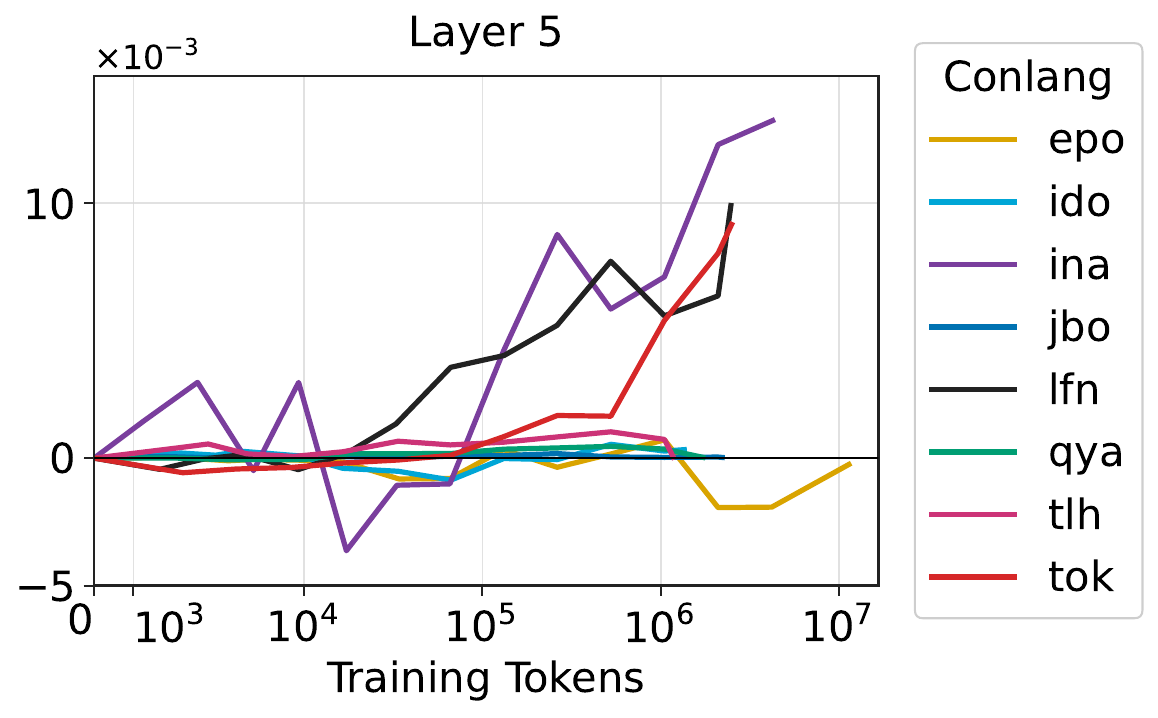}\hfill\layersimilaritypanel{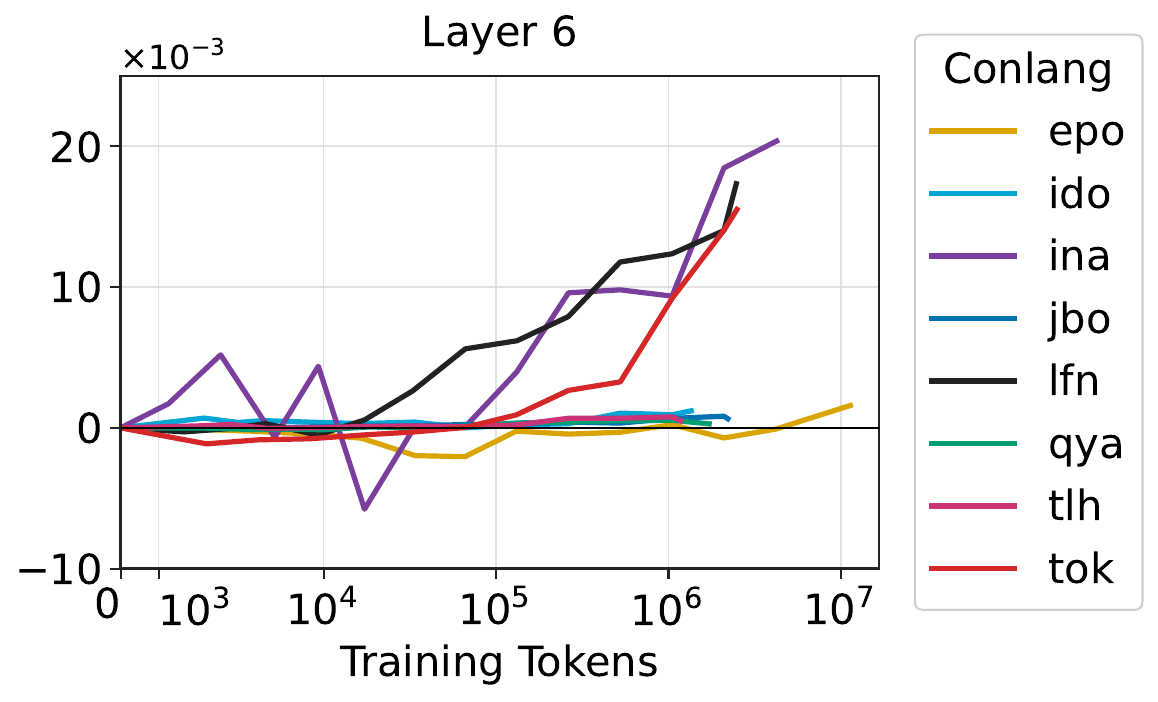}\par\medskip
    \layersimilaritypanel{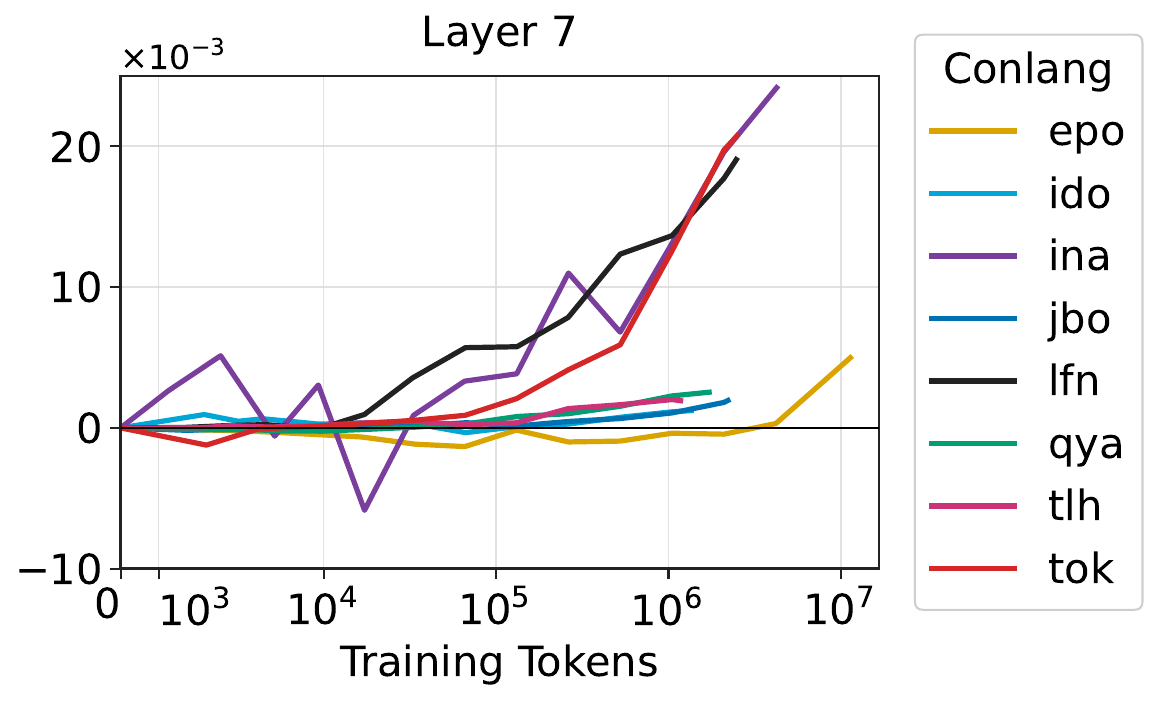}\hfill\layersimilaritypanel{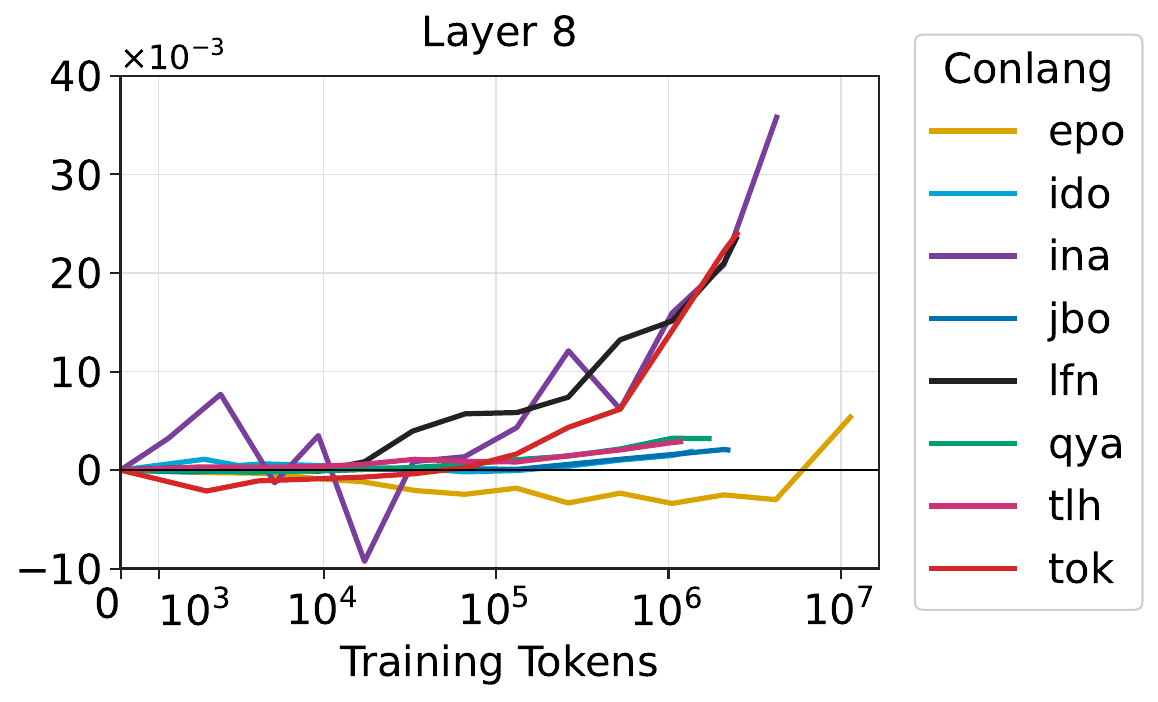}
    \caption{Layer-wise conlang--English lexical alignment scores for all 32 model layers. Each panel shows the score for the eight trained conlangs as a function of training tokens.}
    \label{fig:layerwise_similarities_all}
\end{figure*}

\begin{figure*}[p]
    \ContinuedFloat
    \centering
    \layersimilaritypanel{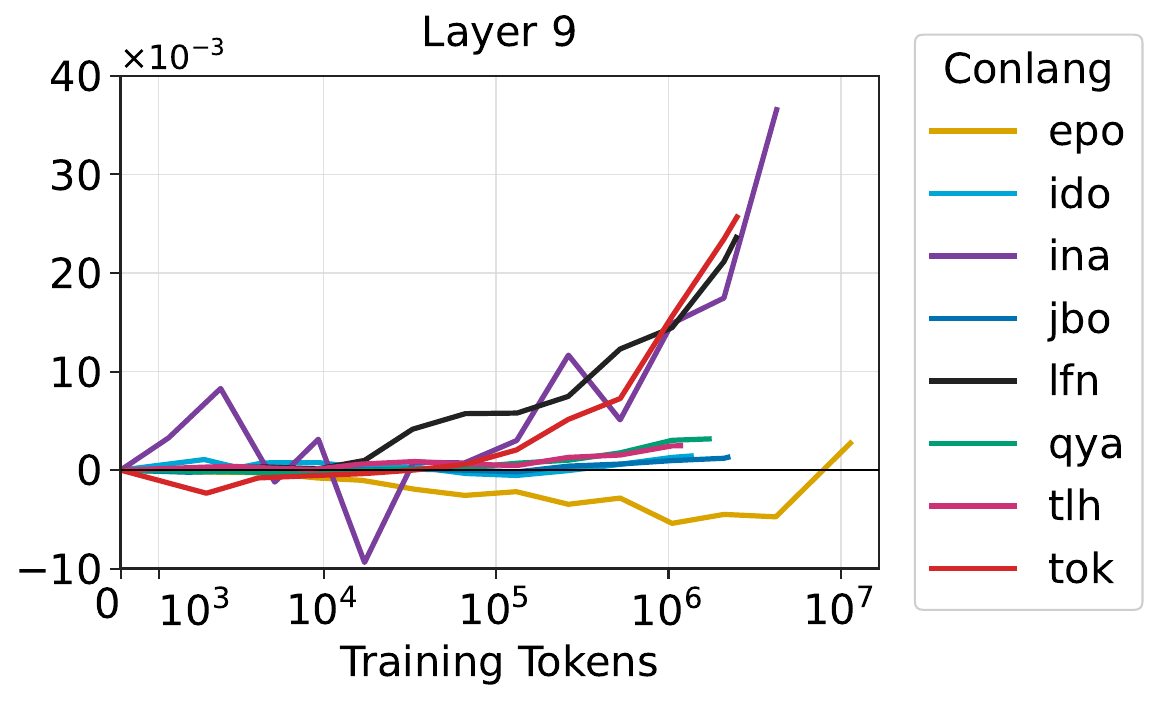}\hfill\layersimilaritypanel{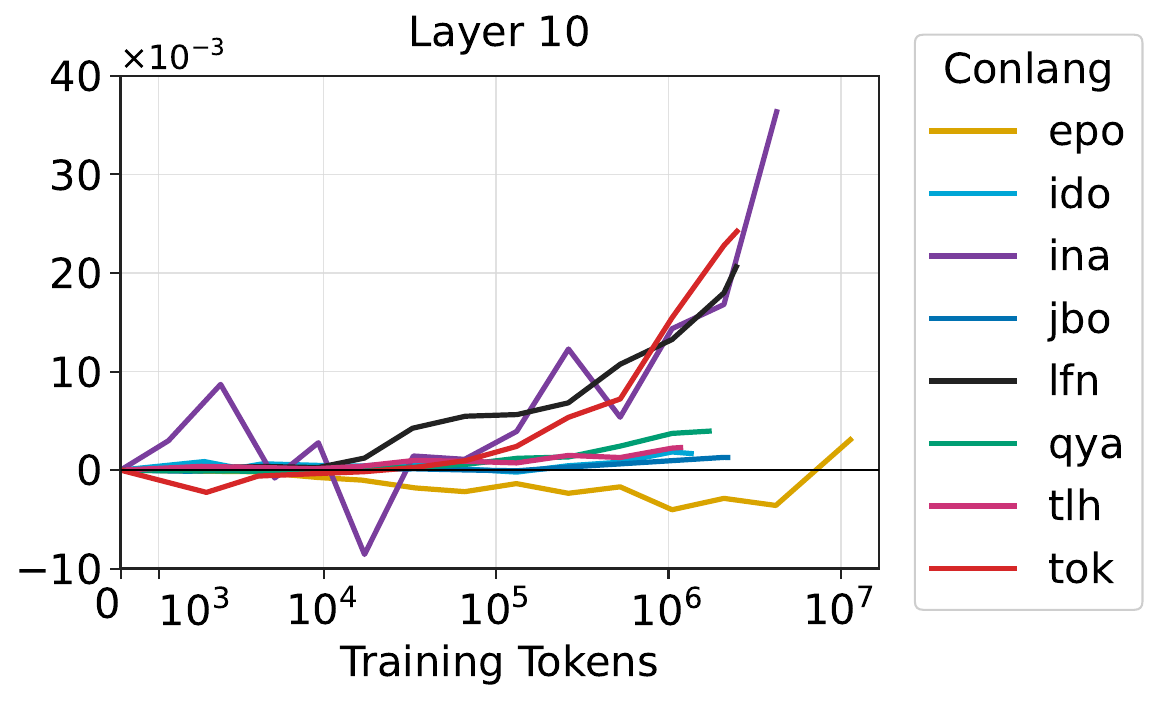}\par\medskip
    \layersimilaritypanel{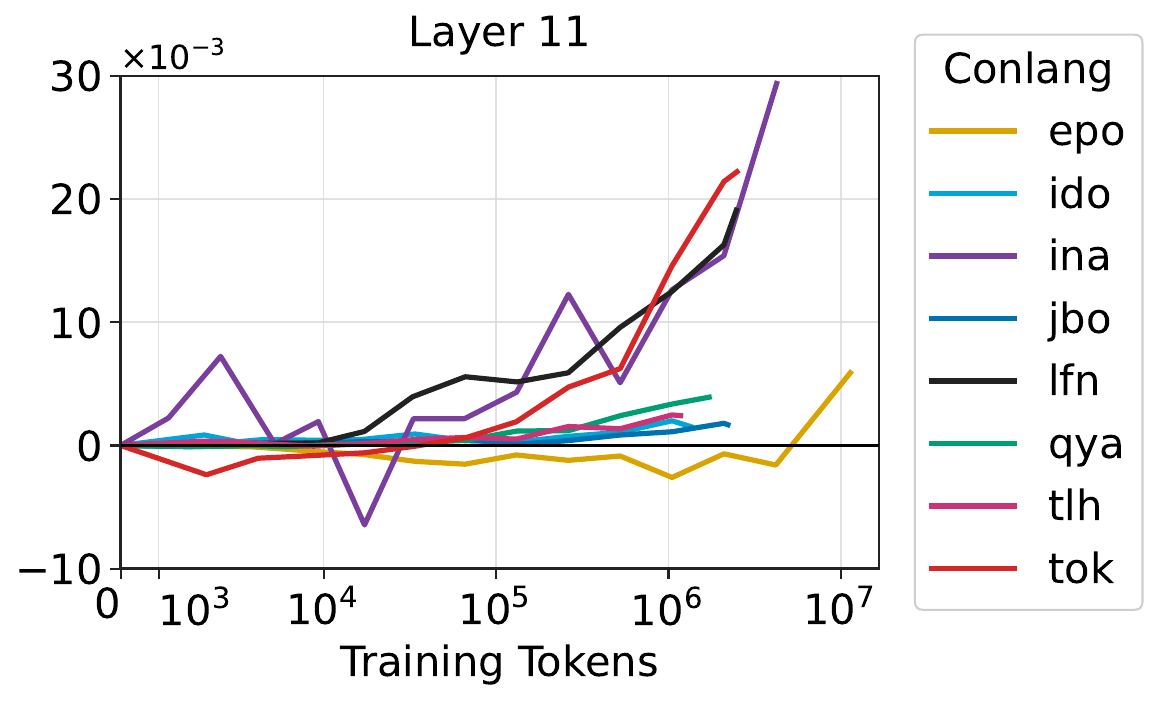}\hfill\layersimilaritypanel{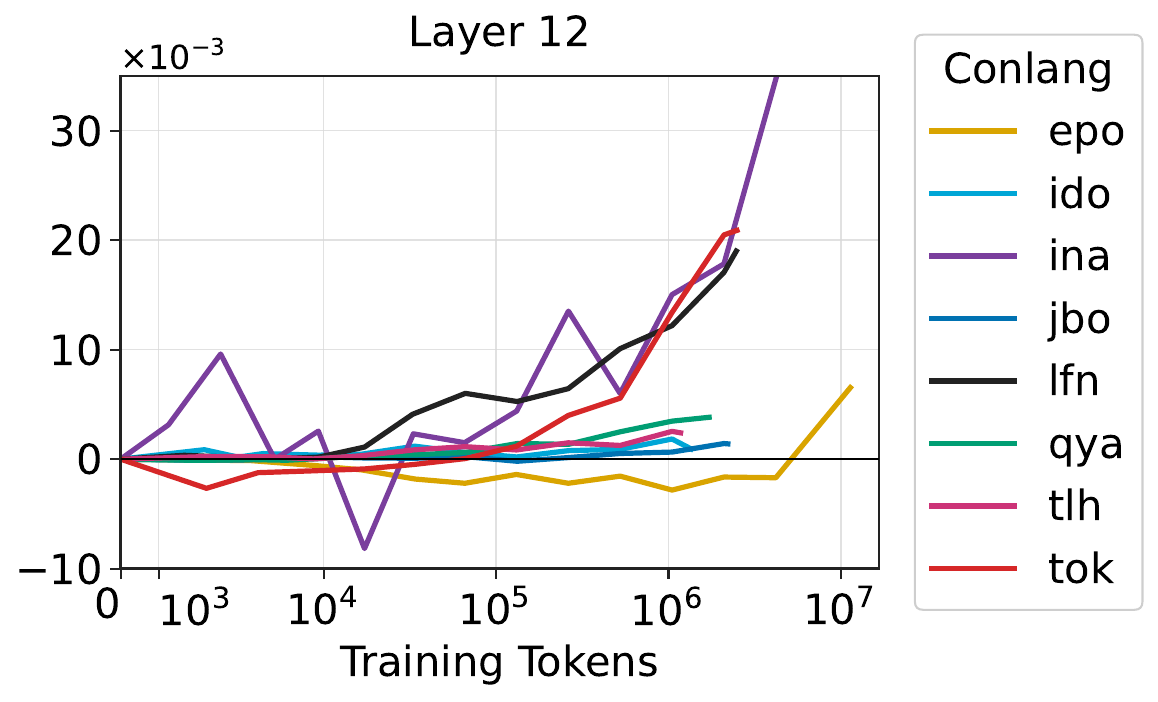}\par\medskip
    \layersimilaritypanel{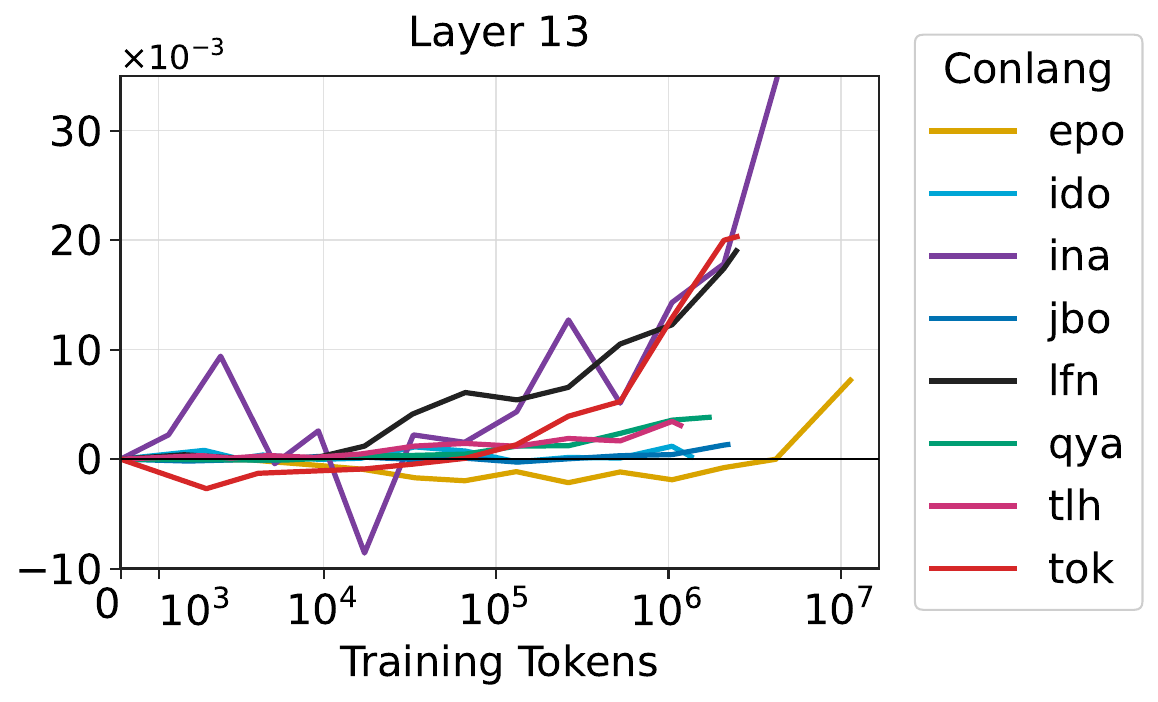}\hfill\layersimilaritypanel{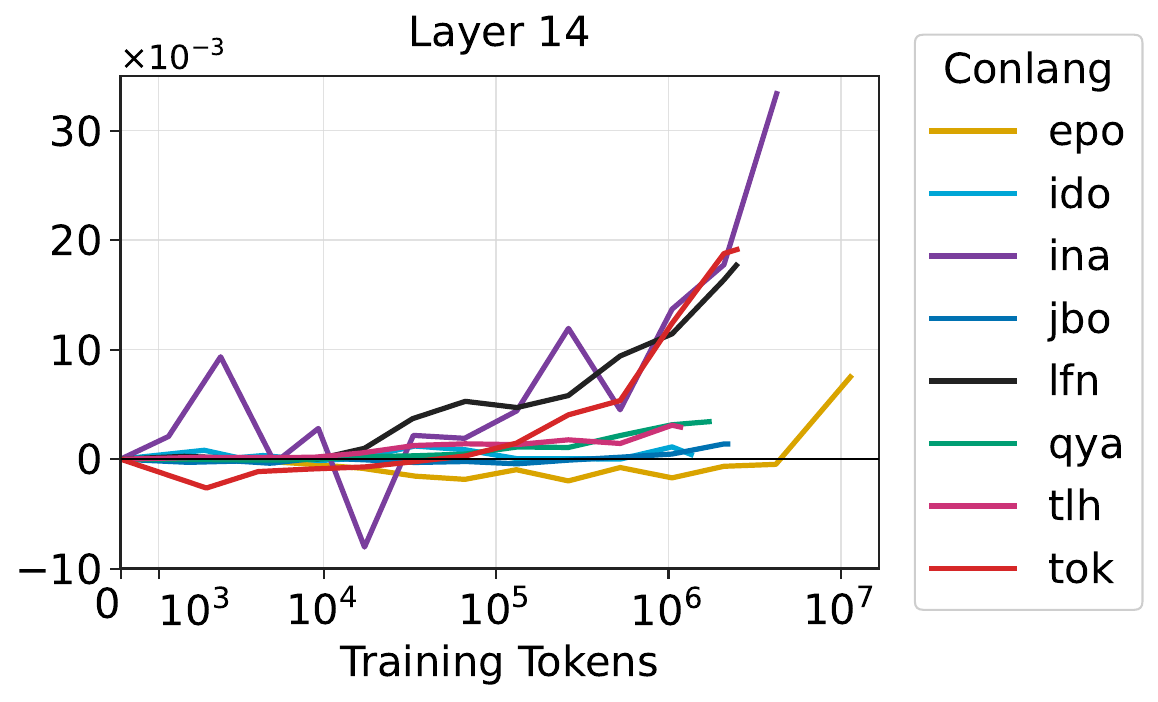}\par\medskip
    \layersimilaritypanel{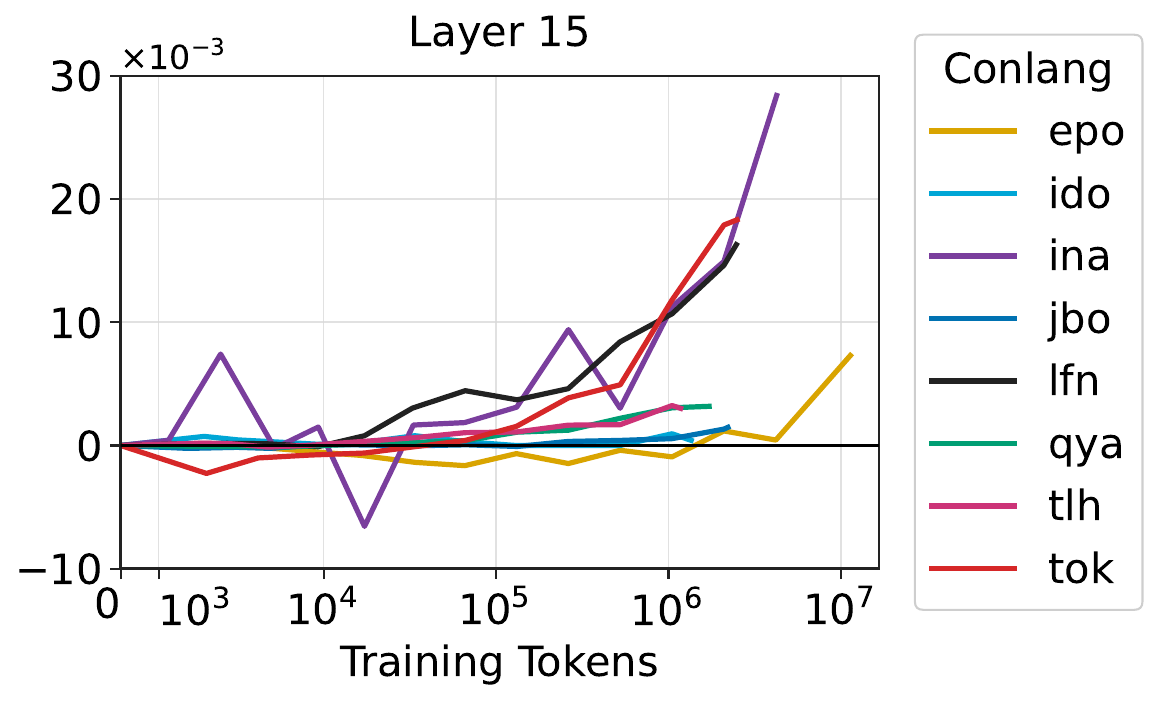}\hfill\layersimilaritypanel{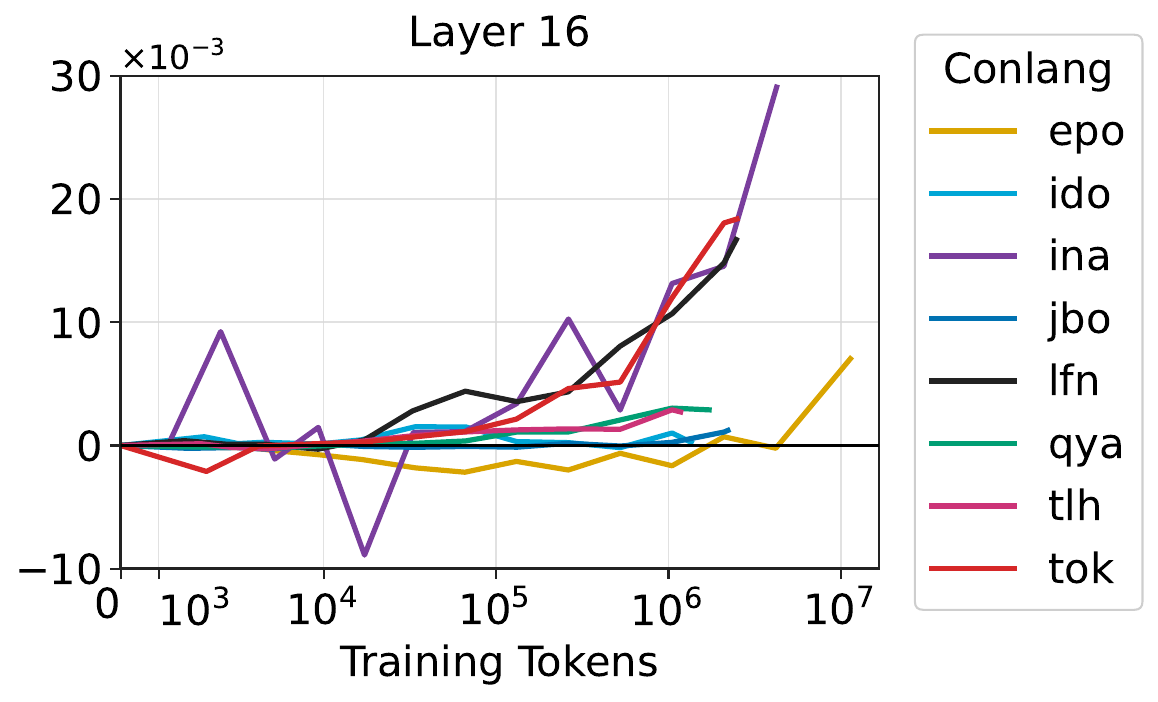}
    \caption[]{Layer-wise conlang--English lexical alignment scores (continued).}
\end{figure*}

\begin{figure*}[p]
    \ContinuedFloat
    \centering
    \layersimilaritypanel{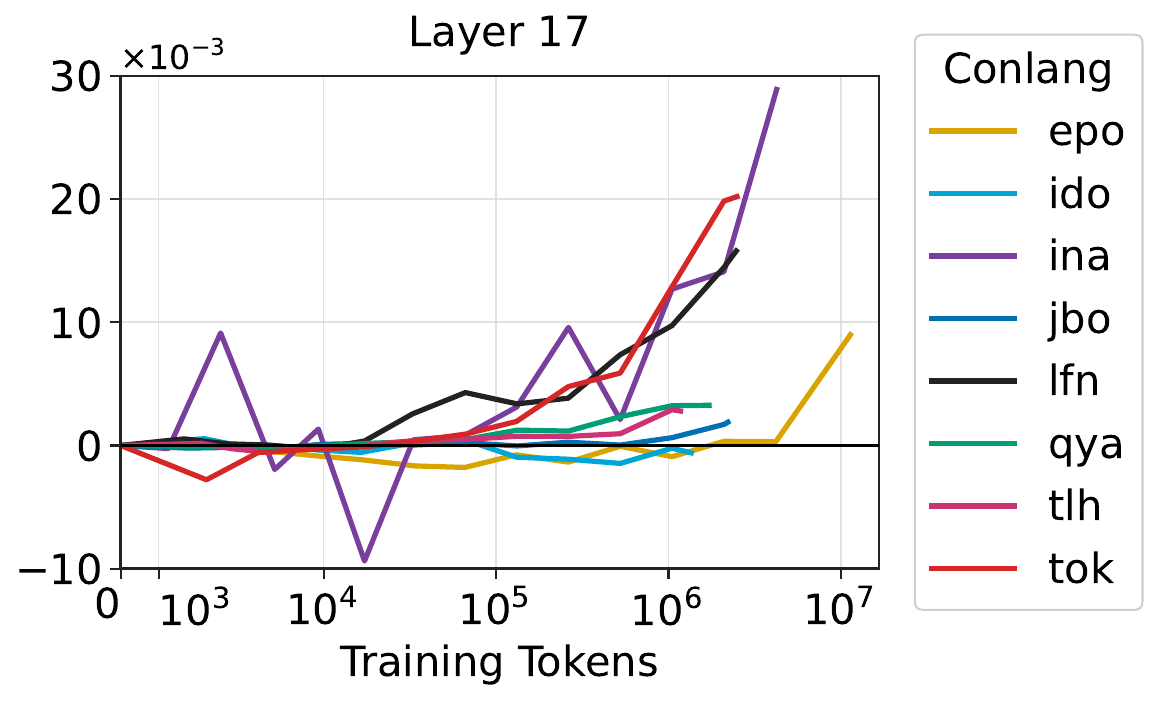}\hfill\layersimilaritypanel{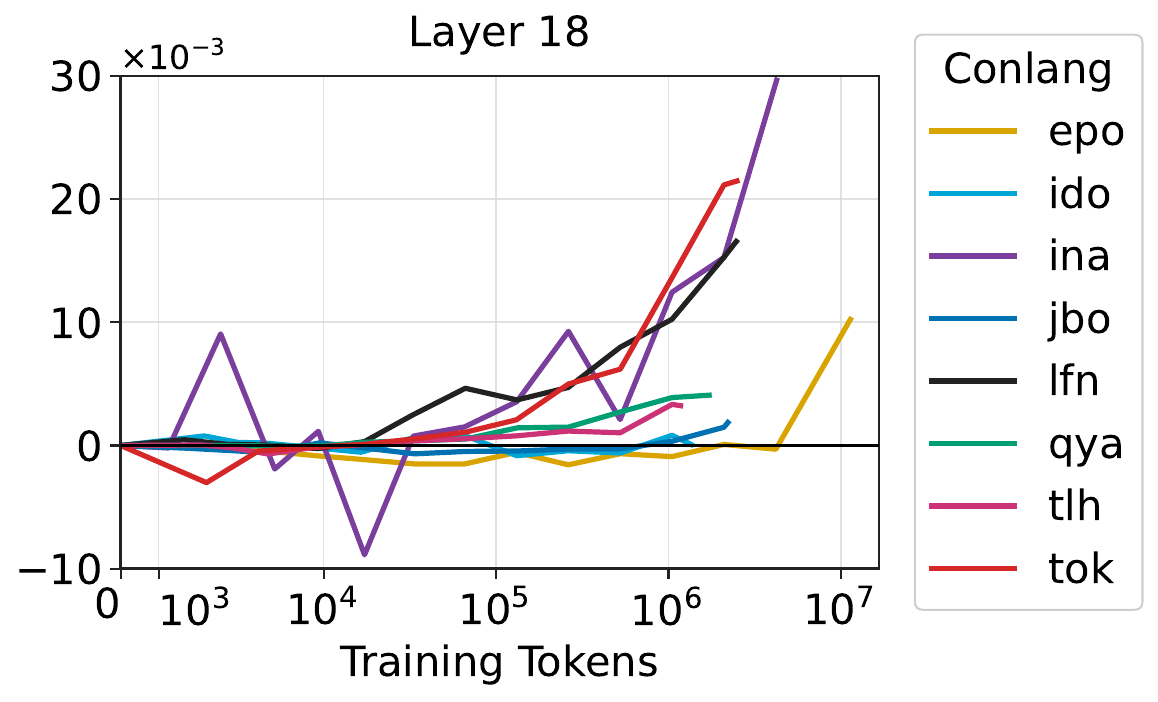}\par\medskip
    \layersimilaritypanel{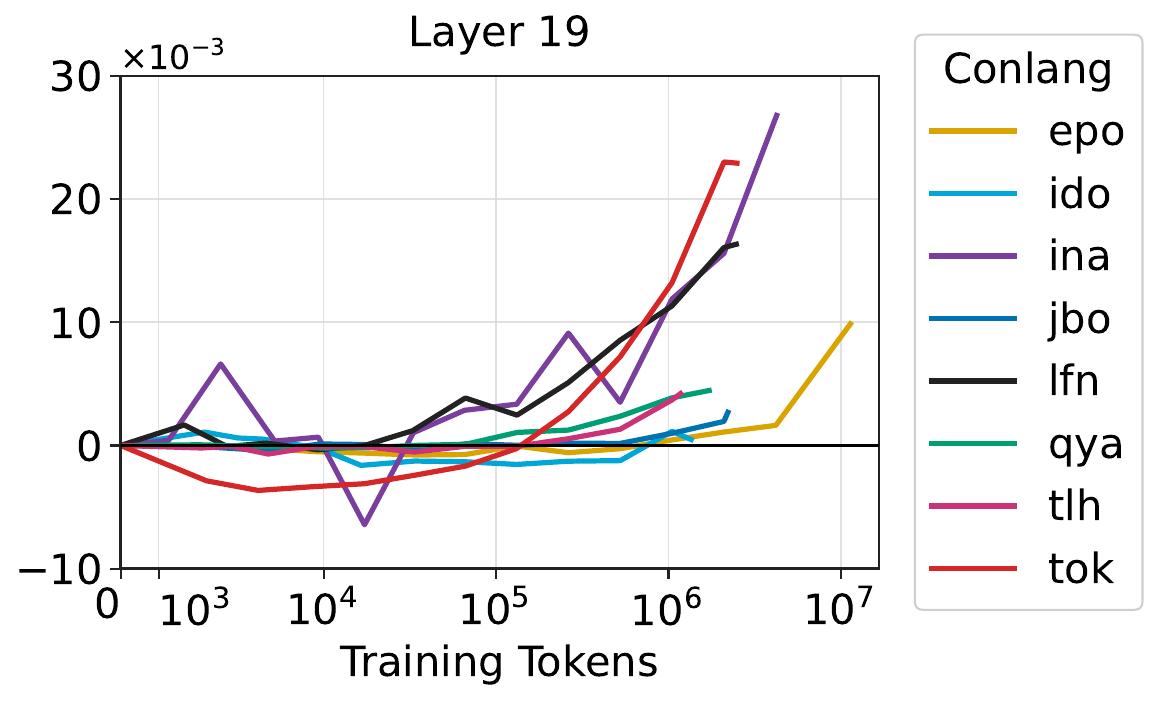}\hfill\layersimilaritypanel{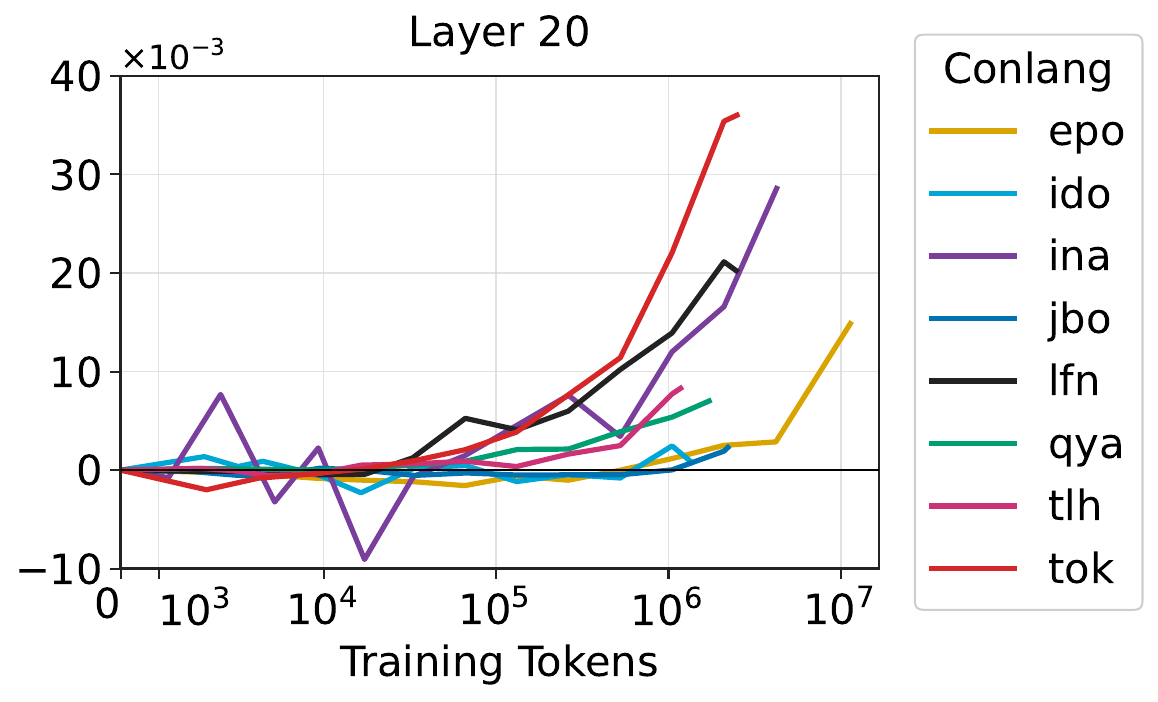}\par\medskip
    \layersimilaritypanel{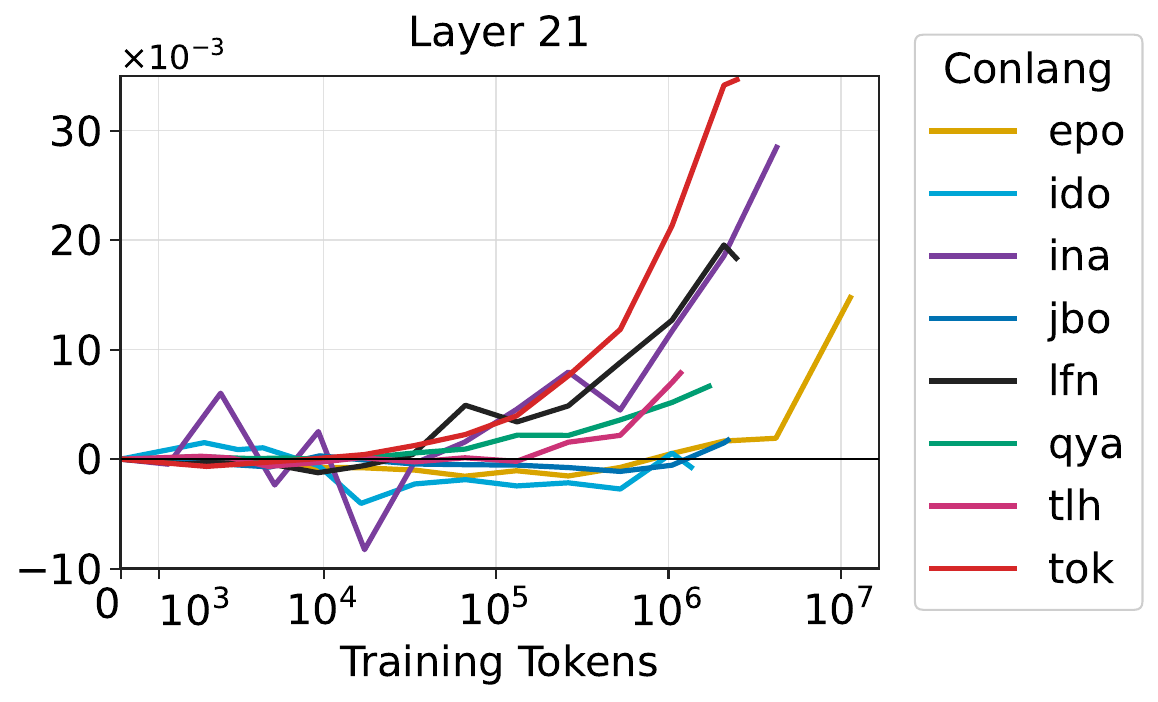}\hfill\layersimilaritypanel{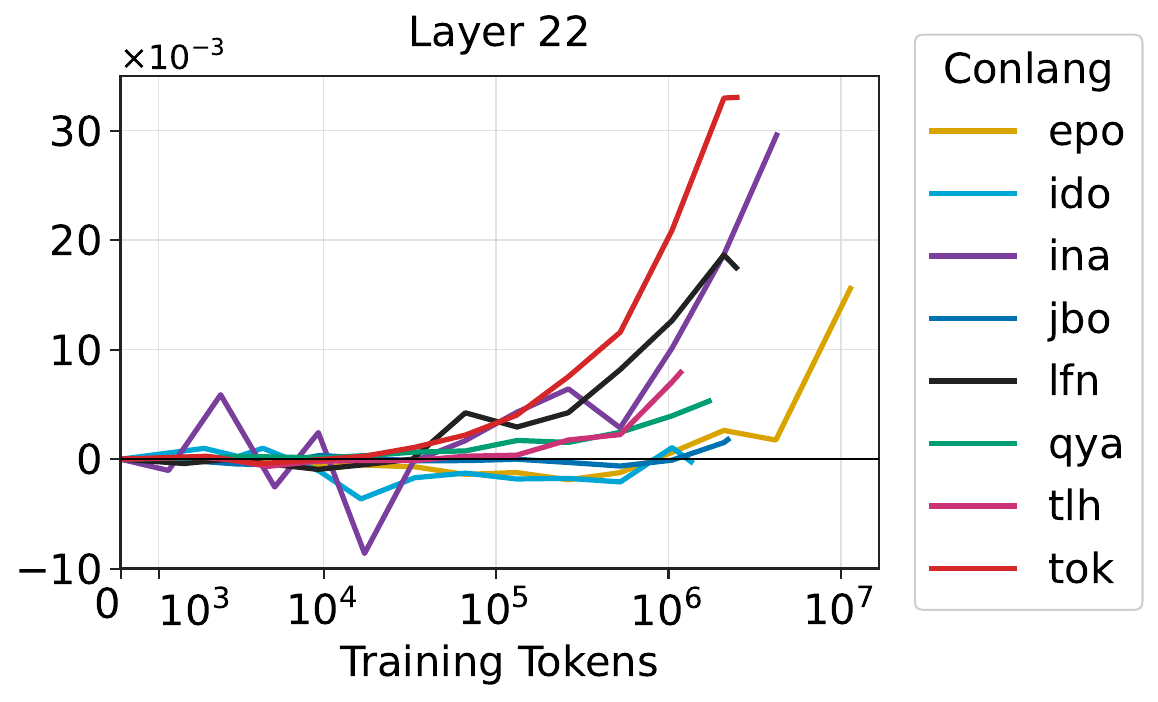}\par\medskip
    \layersimilaritypanel{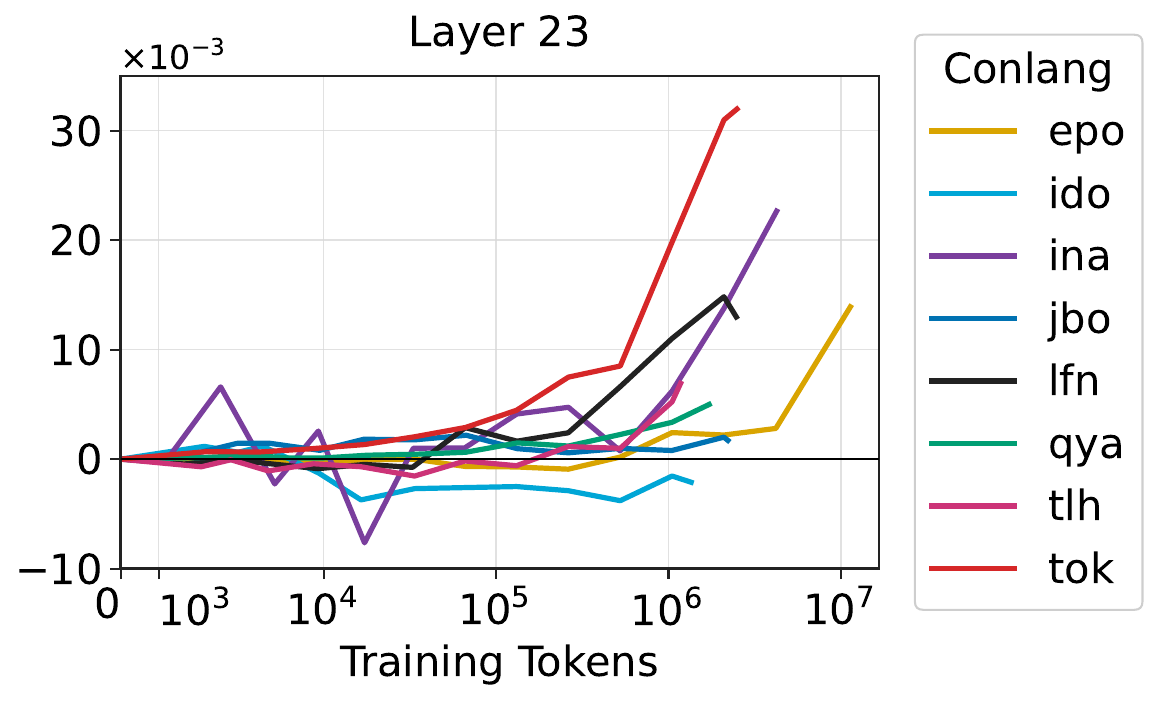}\hfill\layersimilaritypanel{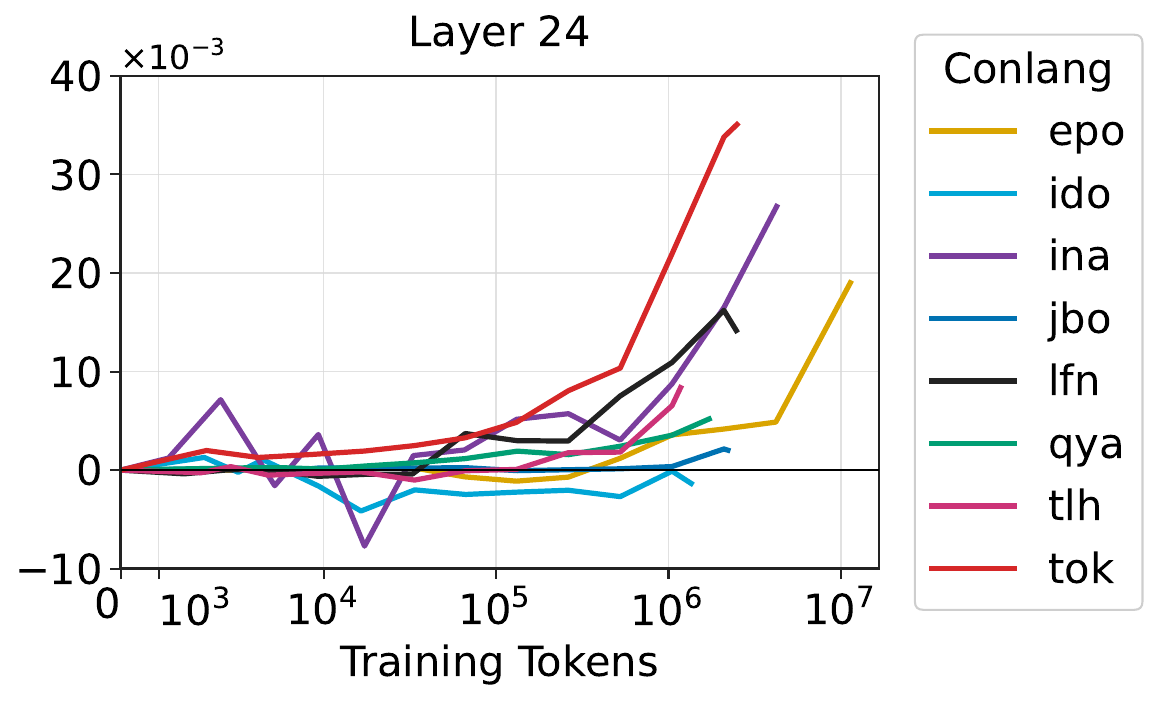}
    \caption[]{Layer-wise conlang--English lexical alignment scores (continued).}
\end{figure*}

\begin{figure*}[p]
    \ContinuedFloat
    \centering
    \layersimilaritypanel{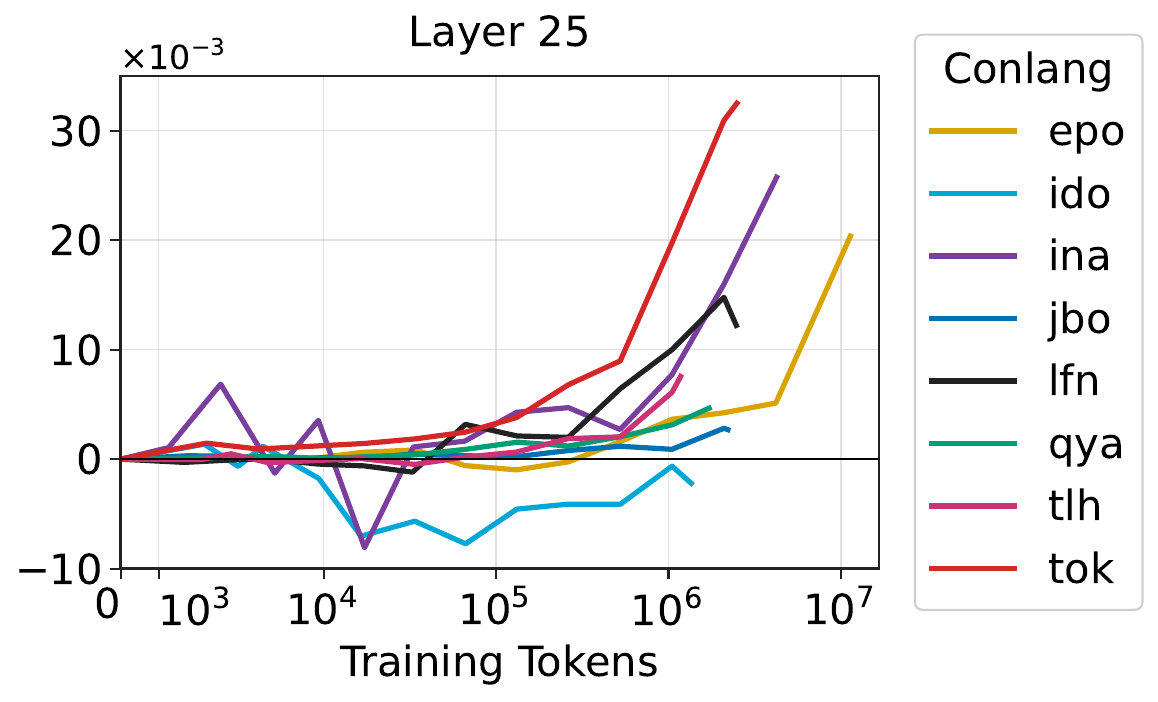}\hfill\layersimilaritypanel{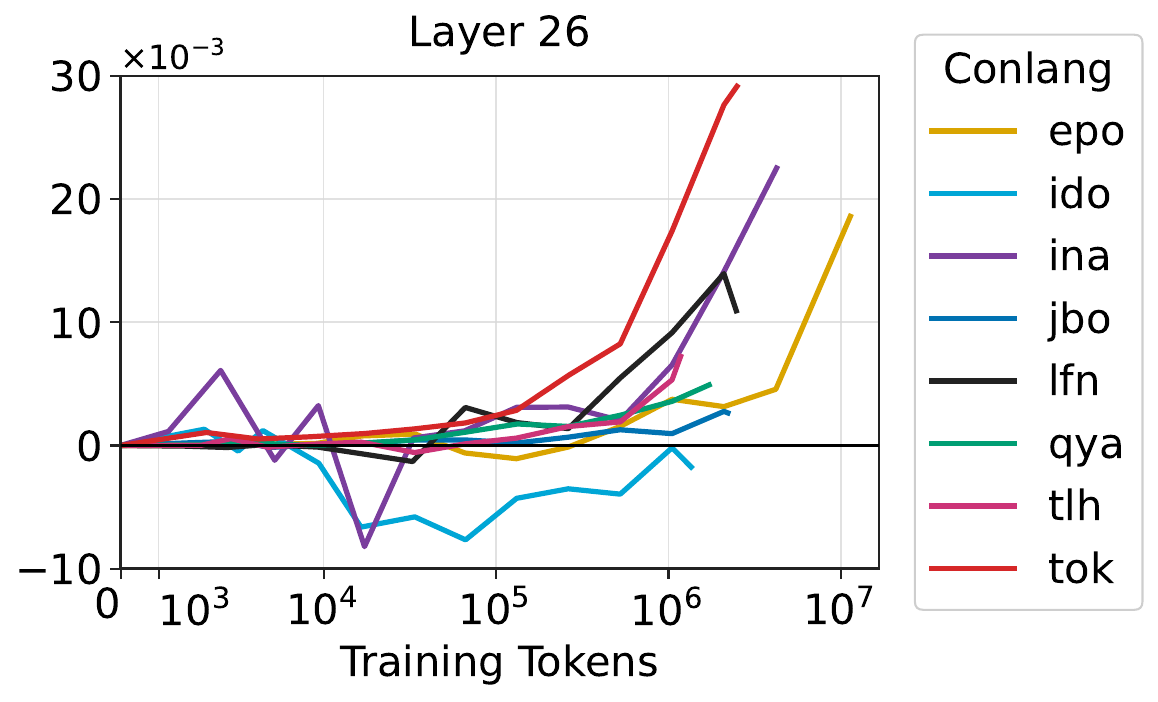}\par\medskip
    \layersimilaritypanel{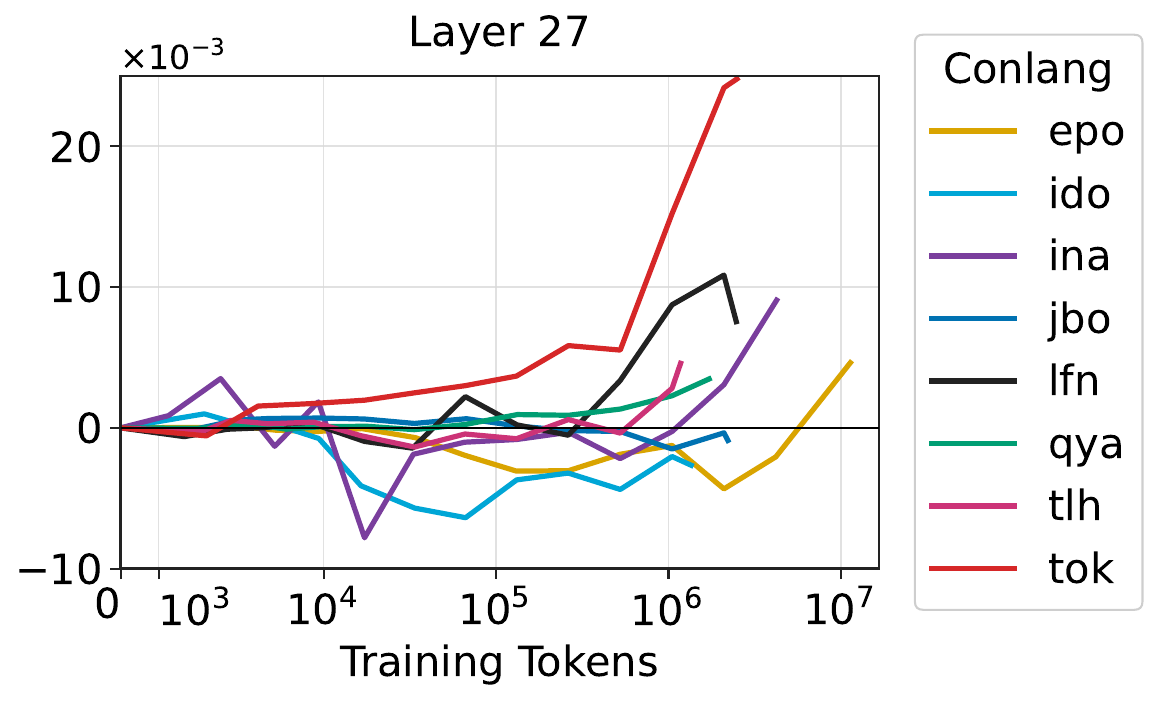}\hfill\layersimilaritypanel{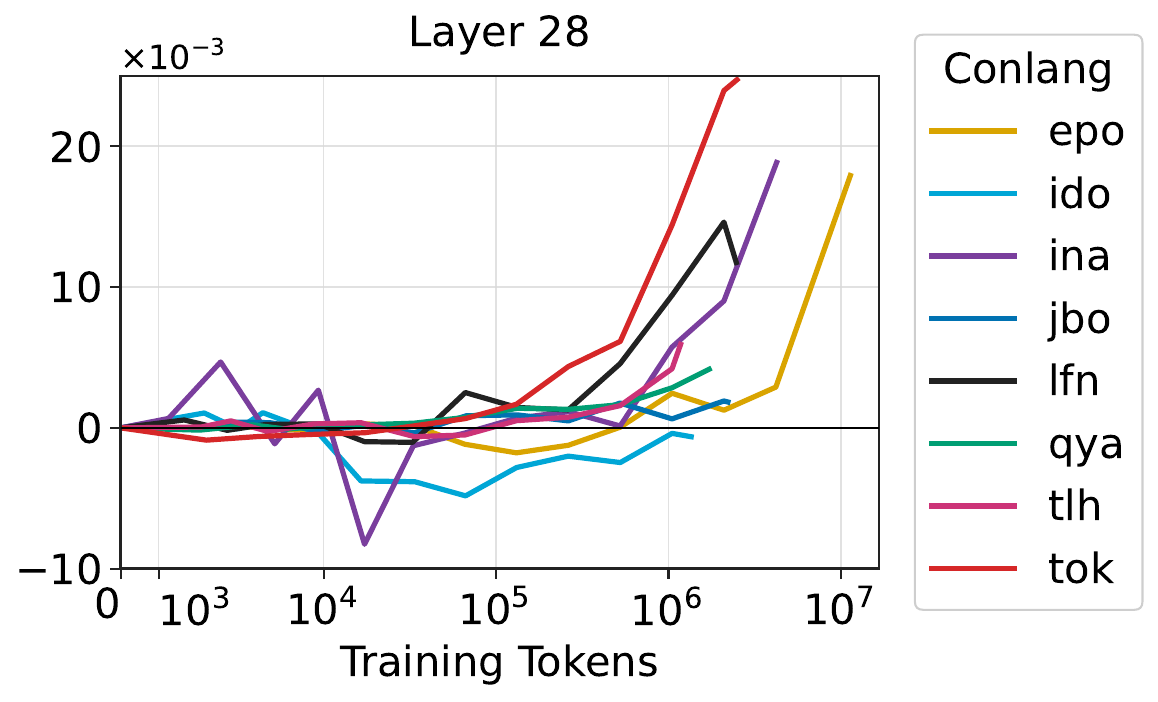}\par\medskip
    \layersimilaritypanel{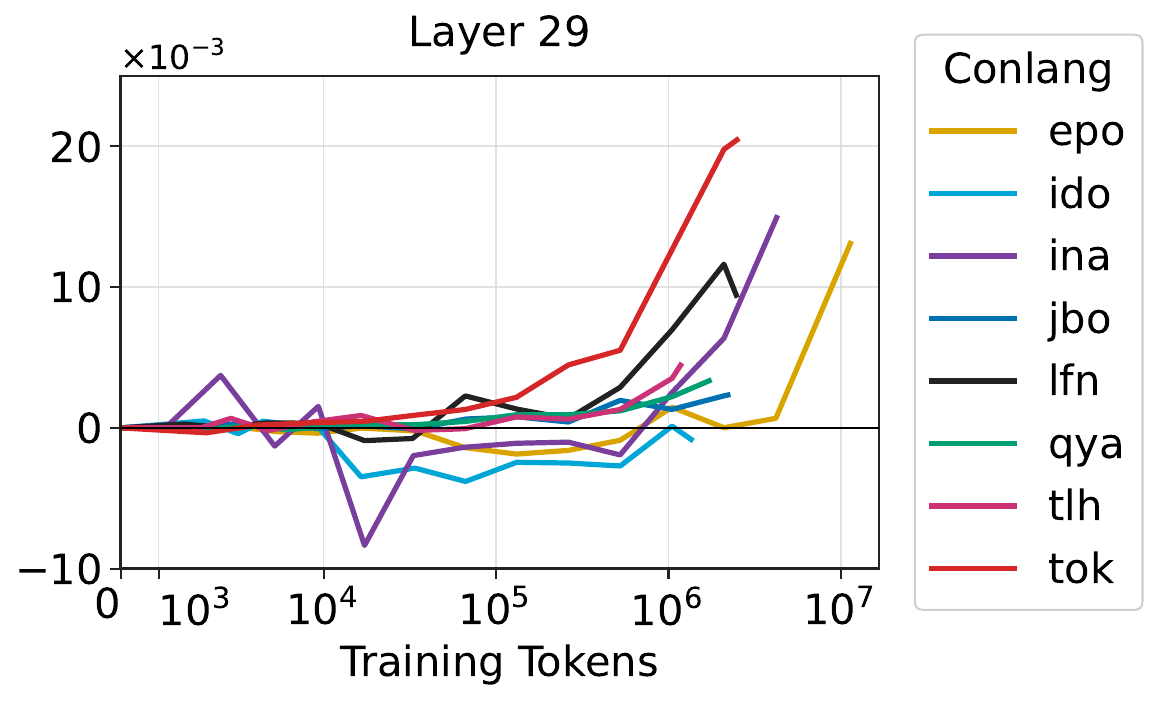}\hfill\layersimilaritypanel{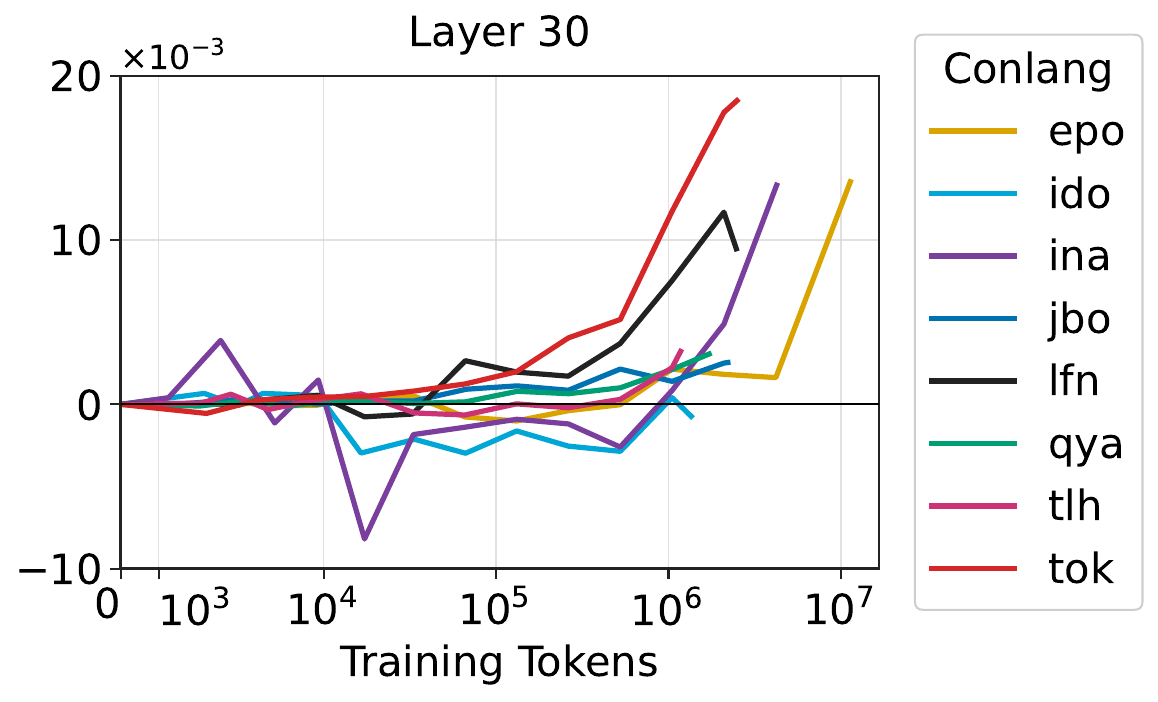}\par\medskip
    \layersimilaritypanel{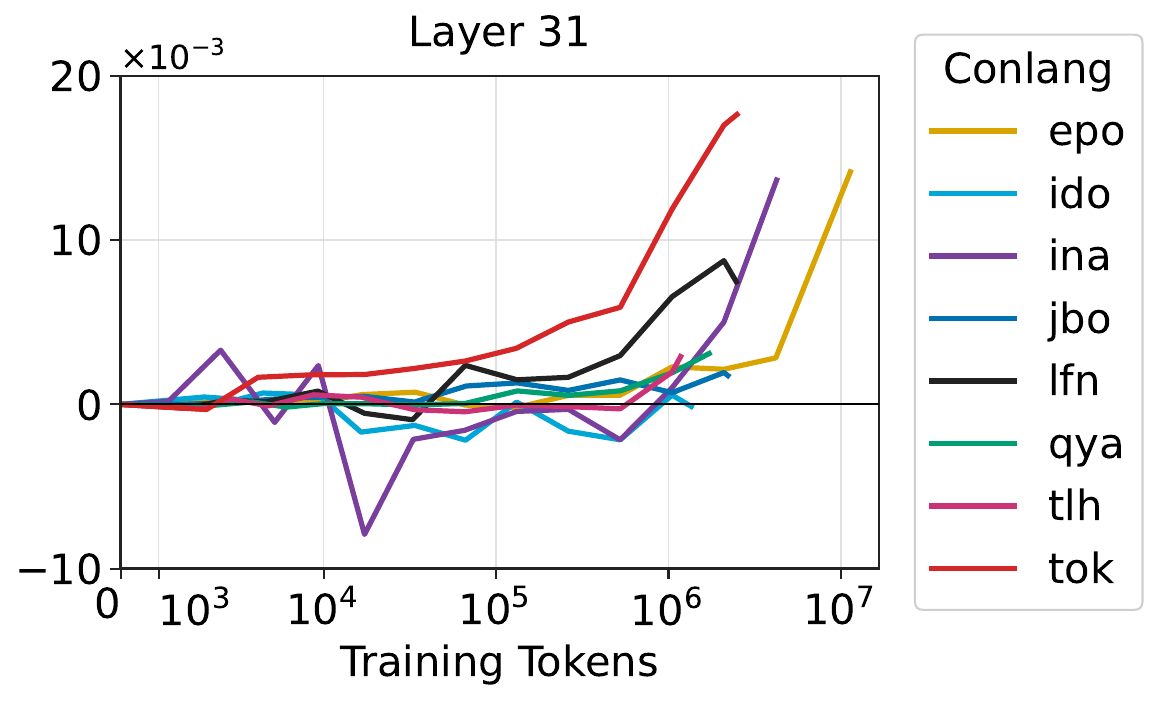}\hfill\layersimilaritypanel{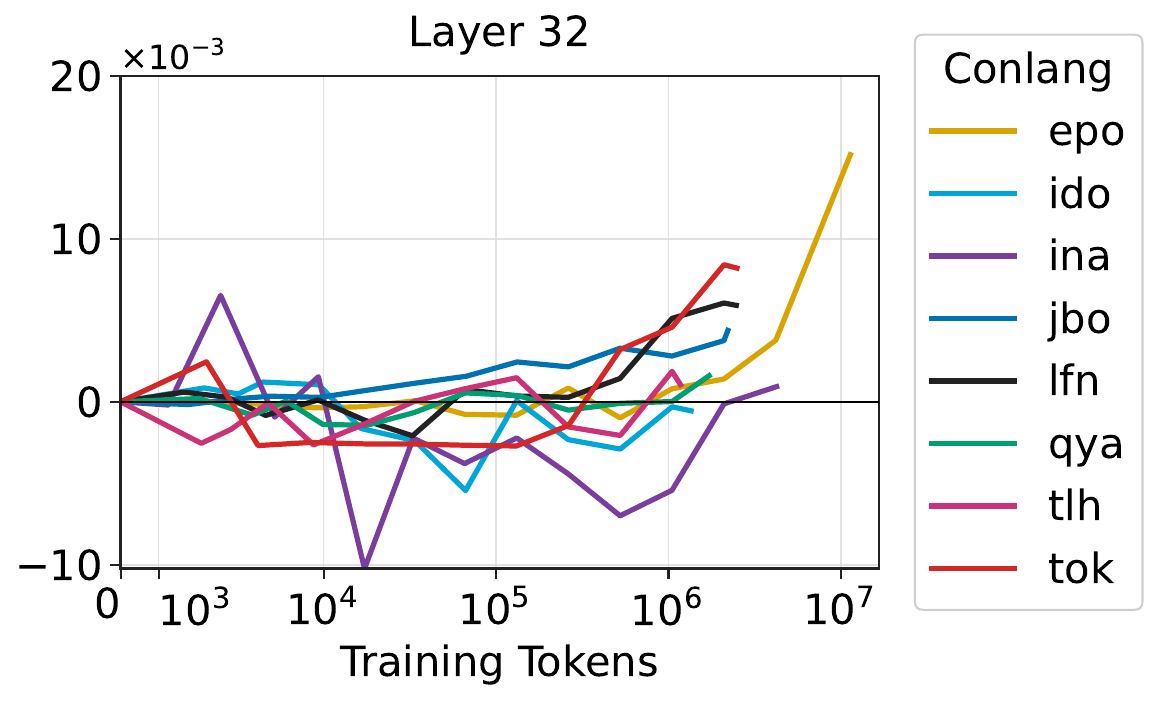}
    \caption[]{Layer-wise conlang--English lexical alignment scores (continued).}
\end{figure*}
\endgroup

\clearpage

\begin{table*}[htb]
\centering
\small
    \begin{tabular}{p{0.9\textwidth}}
        \toprule
Translate the following [\verb|{lang_name}|] text into English.\\
Output the translation and nothing else. If you cannot translate the text, output an empty string.\\\\
\textnormal{[}\verb|{lang_name}|]\\
\verb|{text}|\\
        \bottomrule
    \end{tabular}
    \caption{Prompt for translating non-English natural language data.}
\label{tab:prompt_for_dataset_translation}
\end{table*}

\begin{table*}[htb]
\centering
\small
    \begin{tabular}{p{0.9\textwidth}}
        \toprule
Translate the following text from \verb|{src_lang_name}| (\verb|{src_lang_code}|) to \verb|{tgt_lang_name}| (\verb|{tgt_lang_code}|). Only output the translated text and nothing else.\\\\
Text to translate:\\
\verb|{src_text}|\\
        \bottomrule
    \end{tabular}
    \caption{Prompt for model translation.}
\label{tab:prompt_for_model_translation}
\end{table*}

\begin{table*}[htb]
\centering
\small
    \begin{tabular}{p{0.9\textwidth}}
        \toprule
You are translating from \verb|{src_lang_name}| (\verb|{src_lang_code}|) to \verb|{tgt_lang_name}| (\verb|{tgt_lang_code}|). Use the vocabulary reference below as background knowledge. Do not translate, summarize, or mention the vocabulary reference.\\\\
\textnormal{[}VOCABULARY REFERENCE]\\
\verb|{vocabulary_doc_text}|\\
\textnormal{[}/VOCABULARY REFERENCE]\\\\
Translate the following text from \verb|{src_lang_name}| (\verb|{src_lang_code}|) to \verb|{tgt_lang_name}| (\verb|{tgt_lang_code}|). Only output the translated text and nothing else.\\\\
Text to translate:\\
\verb|{src_text}|\\
        \bottomrule
    \end{tabular}
    \caption{Prompt for model translation with explicit vocabulary information.}
\label{tab:prompt_for_model_translation_vocabulary}
\end{table*}

\clearpage
\onecolumn
\begin{center}
\rule{\textwidth}{0.08em}
\end{center}
\vspace{-0.8\baselineskip}
\begin{Verbatim}[
    fontsize=\small,
    breaklines=true,
    breakanywhere=true,
    breaksymbolleft={},
    xleftmargin=0.02\textwidth,
    xrightmargin=0.02\textwidth,
]
---
name: build-vocabulary-data
description: Build one auditable conlang vocabulary-data JSONL file from supplied URL sources. Use for `/build-vocabulary-data` requests with one `code:` and a non-empty `sources:` list, including web traversal, PDF or DjVu OCR with MinerU, conservative lexical extraction, source auditing, and final validation.
---

# Build Vocabulary Data

## Input

- Process exactly one lowercase `code`, use every supplied `sources` URL as a traversal seed, and
  keep the supplied `output-dir` unchanged.
- Do not read `vocabulary-data-targets.jsonl` unless the user explicitly requests batch processing.
- Treat all paths below as relative to `vocabulary_collection/`.

## Output contract

Persist only these locations:

```text
{output_dir}/source/raw/{code}/          unmodified accepted source files
{output_dir}/source/refined/{code}/      clean UTF-8 Markdown source files
{output_dir}/source/history/{code}/{run_id}/
  01-traverse.jsonl                     traversal decisions
  02-refine.jsonl                       refinement decisions
  03-extract.jsonl                      extraction decisions and rejected records
  summary.json                          run status, counts, checksums, limitations
{output_dir}/{code}.jsonl                final vocabulary data
{output_dir}/.tmp/build-vocabulary-data/{code}/{run_id}/
                                        disposable downloads, OCR trees, and scripts
```

Put unlisted artifacts in the disposable directory. Keep accepted raw and refined files across
runs, adding a short checksum rather than overwriting different bytes. Treat completed history
as immutable and remove the disposable directory after success.

## Start and history

Start every invocation before fetching:

```bash
.venv/bin/python .codex/skills/build-vocabulary-data/scripts/pipeline.py \
  --output-dir "{output_dir}" start \
  --code "{code}" --source "{source_url}"
```

Repeat `--source` for every seed and capture the printed `run_id`.

Use these schemas:

- `01-traverse.jsonl`: one record per fetched candidate with `status`, exact `url`, and `reason`.
  Accepted records add repository-relative `raw_file`, `content_type`, and `sha256`.
- `02-refine.jsonl`: one record per accepted raw source with `status`, `raw_file`, `method`, and
  `reason`. Accepted records add repository-relative `refined_file` and `sha256`.
- `03-extract.jsonl`: one `record_type: "source"` per accepted refined source with `status`,
  `refined_file`, `method`, and `reason`. Accepted records add non-negative `source_records`,
  `accepted_records`, `rejected_records`, `emitted_rows`, and one-based `output_line_start` and
  `output_line_end`. For each rejected lexical record, add `record_type: "rejected_record"` with
  `refined_file`, stable `locator`, and concrete `reason`; add `headword` and `original_gloss`
  when safe.
- `summary.json`: let the helper manage it; do not edit it.

Create stage records in the disposable directory and append them after each stage:

```bash
.venv/bin/python .codex/skills/build-vocabulary-data/scripts/pipeline.py \
  --output-dir "{output_dir}" append \
  --code "{code}" --run-id "{run_id}" --stage traverse \
  --records-file "{traverse_jsonl}"
```

Use `--stage refine` and `--stage extract` for the later stages.

If the build cannot finish, run:

```bash
.venv/bin/python .codex/skills/build-vocabulary-data/scripts/pipeline.py \
  --output-dir "{output_dir}" fail \
  --code "{code}" --run-id "{run_id}" --reason "{concrete_reason}"
```

## Stage 1: traverse and save raw

1. Fetch every supplied seed into the disposable work directory.
2. For HTML, stay on the same scheme, host, and port; remove fragments and tracking parameters,
   deduplicate equivalent URLs, and inspect at most 100 pages across all seeds.
3. Prioritize vocabulary resources and pages with explicit lexical pairs. Skip navigation,
   media, archives, authentication, search, feeds, comments, and unrelated pages. Never reject a
   candidate merely because its page, lexical records, or glosses are wholly or partly
   non-English.
4. Record every fetched candidate, including rejections, with a precise reason.
5. Save accepted bodies byte-for-byte under `source/raw/{code}/`, without headers or manifests.
   Use deterministic safe names and add a short checksum on collision.

## Stage 2: refine

Process only accepted files in `source/raw/{code}/`.

1. Produce one clean Markdown file per usable raw file. Preserve order, structure, examples, and
   stable row or page locators; remove only boilerplate without lexical evidence.
2. Convert non-PDF sources to readable Markdown without flattening record boundaries.
3. Process every PDF through MinerU without separately testing or extracting embedded text.
   Convert DjVu to a temporary PDF first. Verify `nvidia-smi` and CUDA, use `pipeline` with
   `auto`, and keep all output disposable:

   ```bash
   MINERU_DEVICE_MODE=cuda CUDA_VISIBLE_DEVICES="${CUDA_VISIBLE_DEVICES:-0}" \
     .venv/bin/mineru -p "{input_pdf}" -o "{auto_work_dir}" -b pipeline -m auto -l en
   ```

4. Inspect the beginning, middle, and end. If garbled or structurally unusable, rerun with OCR:

   ```bash
   MINERU_DEVICE_MODE=cuda CUDA_VISIBLE_DEVICES="${CUDA_VISIBLE_DEVICES:-0}" \
     .venv/bin/mineru -p "{input_pdf}" -o "{ocr_work_dir}" -b pipeline -m ocr -l en
   ```

   Use OCR only when cleaner, and record the chosen method.
5. Run `fail` if MinerU cannot use CUDA. Keep MinerU trees, converters, and scripts disposable.
6. Accept only refinements with intact headword/gloss boundaries and locators; otherwise record
   a concrete rejection reason. Save accepted `.md` files under `source/refined/{code}/`.

## Stage 3: extract

Extract only from accepted refined Markdown; consult raw files only to verify field boundaries or
recover the exact source text.

1. Test each record layout and parser branch. Prefer explicit structure over broad regexes.
2. Reject sources whose lexical records cannot be distinguished conservatively. Emit an entry
   only when one record explicitly supports both headword and explanation; do not infer fields
   or treat examples, prose, navigation, or paradigms as entries. Never reject a source or record
   merely because its explanation is not in English.
3. Log each malformed, ambiguous, blank, or uncertain record as `rejected_record` with reviewable
   text and a locator, then omit it.
4. Copy each accepted record's explanation into `original_gloss` exactly as it appears in the
   source. Preserve its original language, spelling, case, punctuation, diacritics, sense
   divisions, part-of-speech labels, and other labels. Do not translate, normalize, paraphrase,
   trim, or silently repair `original_gloss`; JSON escaping does not count as changing its value.
5. Preserve source order, assign each accepted extraction a contiguous output range, and
   reconcile:

   ```text
   source_records = accepted_records + rejected_records
   emitted_rows = output_line_end - output_line_start + 1
   ```

   Explain any accepted record that expands into multiple rows.
6. Write the extracted rows to a temporary file for Stage 4.

## Final row format

Write UTF-8 JSONL with one compact object per non-empty line and exactly these non-empty string
fields in this order:

```jsonl
{"word":"conlang word","meaning":"English explanation","original_gloss":"source explanation","source_url":"https://example.com/exact-entry-page"}
```

- In `word`, preserve spelling, meaningful case, diacritics, apostrophes, and hyphenation. Include
  listed roots, affixes, particles, variants, and phrases; do not invent inflections.
- Keep `original_gloss` as an exact copy of the source explanation, including its language,
  spelling, case, punctuation, diacritics, senses, part-of-speech, and labels. Never translate or
  edit it.
- Write a faithful English explanation in `meaning` during Stage 4.
- Use the exact HTTP(S) entry page or standalone-file URL in `source_url`, without fragments or
  tracking parameters.
- Keep repeated headwords when their `meaning` differs.

## Stage 4: translate meanings into English

Process every extracted row, including rows whose `original_gloss` is wholly or partly
non-English. Non-English glosses are valid source data and are not a rejection reason.

1. If `original_gloss` is already English, copy it exactly into `meaning`.
2. Otherwise, translate the non-English content into faithful, natural English. Preserve every
   supported sense, part-of-speech distinction, usage label, and explanatory note without adding
   unsupported information.
3. Change only `meaning`; when decoded from JSON, `original_gloss` must exactly equal the source
   explanation. JSON escaping is the only permitted representational difference. If a translation
   is uncertain, use the most literal supported English rendering and record the uncertainty as a
   material limitation instead of rejecting the source merely for being non-English.
4. Validate the complete translated result and atomically replace
   `{output_dir}/{code}.jsonl`.

## Stage 5: remove exact duplicates

Let the `finish` command programmatically remove duplicate rows before final validation. It keeps
the first row for each exact `(word, meaning)` pair and removes every later match, even when
`original_gloss` or `source_url` differs. Matching is literal and case-sensitive; do not trim,
case-fold, normalize Unicode, or otherwise merge near-matches. The helper records the number of
removed rows as `counts.duplicate_rows_removed` in `summary.json`.

## Finish

After appending all stage records, run:

```bash
.venv/bin/python .codex/skills/build-vocabulary-data/scripts/pipeline.py \
  --output-dir "{output_dir}" finish \
  --code "{code}" --run-id "{run_id}" \
  --limitation "{material_limitation_if_any}"
```

Omit `--limitation` when none exists and repeat it when needed. This command performs Stage 5
before it completes the summary. Report success only after it passes. Report only the final path,
row count, exact duplicate rows removed, accepted/rejected source and record counts, and material
limitations.
\end{Verbatim}
\vspace{-0.8\baselineskip}
\begin{center}
\rule{\textwidth}{0.08em}
\captionof{table}{Prompt for the vocabulary collection agent.}
\label{tab:prompt_for_vocabulary_collection}
\end{center}
\twocolumn

\end{document}